\documentclass[Afour,sageh,times]{sagej}
\usepackage{amsmath,amssymb,amsfonts}
\usepackage{graphicx}
\usepackage{booktabs} % for professional tables
\graphicspath{{./figures/}}
\DeclareGraphicsExtensions{.pdf,.jpeg,.png}
\usepackage{textcomp}
\usepackage{xcolor}
\usepackage{units}
\usepackage[caption=false, font=footnotesize]{subfig}
\usepackage{multirow}
\usepackage{gensymb}

\usepackage[inline]{enumitem}   

\usepackage{moreverb}
\usepackage[hyphens]{url}
\usepackage[colorlinks,bookmarksopen,bookmarksnumbered,citecolor=red,urlcolor=red]{hyperref}
\usepackage{cleveref}

\usepackage{array}
\usepackage{tablefootnote}

\usepackage[linesnumbered,ruled,vlined]{algorithm2e}
\SetKwInput{KwInput}{Input}                % Set the Input
\SetKwInput{KwOutput}{Output}              % set the Output

\DeclareUnicodeCharacter{03B2}{\ensuremath{\beta}}
\DeclareUnicodeCharacter{03B8}{\ensuremath{\theta}}
\DeclareUnicodeCharacter{2212}{\ensuremath{-}}

\DeclareMathOperator*{\argmax}{arg\,max}

\def\BibTeX{{\rm B\kern-.05em{\sc i\kern-.025em b}\kern-.08em
    T\kern-.1667em\lower.7ex\hbox{E}\kern-.125emX}}

\newcommand{\ie}{\textit{i.e.,}\ }
\newcommand{\eg}{\textit{e.g.,}\ } % the \ ensures latex does not treat command as end of sentence

\newcommand{\figref}[1]{Fig.~\ref{#1}}
\newcommand{\tabref}[1]{Table~\ref{#1}}
\newcommand{\figrefs}[2]{Figs.~\ref{#1} and \ref{#2}}

\newcount\Comments  % 1 suppresses notes to selves in text
\usepackage{color}
\definecolor{darkgreen}{rgb}{0,0.5,0}
\definecolor{purple}{rgb}{1,0,1}
\newcommand{\kibitz}[2]{\ifnum\Comments=0\textcolor{#1}{#2}\fi}

\begin{document}
%%%%%%%%%%%%%%%%%%%%%%%%%%%%%%%%%%%%%%%%%%%%%%%%%%%%%%%%%%%%%%%%%%%%%%%%%%%%%%%%
\runninghead{Xie et al.}

\title{\LARGE \bf Semantic2D: Enabling Semantic Scene Understanding with 2D Lidar Alone}

\author{
Zhanteng Xie\affilnum{1},
Yipeng Pan\affilnum{1},
Yinqiang Zhang\affilnum{1},
Jia Pan\affilnum{1},
Philip Dames\affilnum{2}
}

\affiliation{
\affilnum{1}School of Computing and Data Science, The University of Hong Kong, Hong Kong SAR, China \\
\affilnum{2}Department of Mechanical Engineering, Temple University, Philadelphia, PA 19122, USA
}

\corrauth{Jia Pan, Philip Dames}
\email{jpan@cs.hku.hk, pdames@temple.edu}

%%%%%%%%%%%%%%%%%%%%%%%%%%%%%%%%%%%%%%%%%%%%%%%%%%%%%%%%%%%%%%%%%%%%%%%%%%%%%%%%
\begin{abstract}
This article presents a complete semantic scene understanding workflow using only a single 2D lidar. This fills the gap in 2D lidar semantic segmentation, thereby enabling the rethinking and enhancement of existing 2D lidar-based algorithms for application in various mobile robot tasks. It introduces the first publicly available 2D lidar semantic segmentation dataset and the first fine-grained semantic segmentation algorithm specifically designed for 2D lidar sensors on autonomous mobile robots. To annotate this dataset, we propose a novel semi-automatic semantic labeling framework that requires minimal human effort and provides point-level semantic annotations. The data was collected by three different types of 2D lidar sensors across twelve indoor environments, featuring a range of common indoor objects. Furthermore, the proposed semantic segmentation algorithm fully exploits raw lidar information -- position, range, intensity, and incident angle -- to deliver stochastic, point-wise semantic segmentation. We present a series of semantic occupancy grid mapping experiments and demonstrate two semantically-aware navigation control policies based on 2D lidar. These results demonstrate that the proposed semantic 2D lidar dataset, semi-automatic labeling framework, and segmentation algorithm are effective and can enhance different components of the robotic navigation pipeline. Multimedia resources are available at: \url{https://youtu.be/P1Hsvj6WUSY}.

\end{abstract}

%%%%%%%%%%%%%%%%%%%%%%%%%%%%%%%%%%%%%%%%%%%%%%%%%%%%%%%%%%%%%%%%%%%%%%%%%%%%%%%%
\keywords{Semantic scene understanding, Semantic segmentation, 2D Lidar, Dataset, Mobile robotics}
\maketitle

%%%%%%%%%%%%%%%%%%%%%%%%%%%%%%%%%%%%%%%%%%%%%%%%%%%%%%%%%%%%%%%%%%%%%%%%%%%%%%%%
\section{Introduction}
\label{sec:introduction}
\begin{figure}[t]
    \centering
    \includegraphics[width=0.47\textwidth]{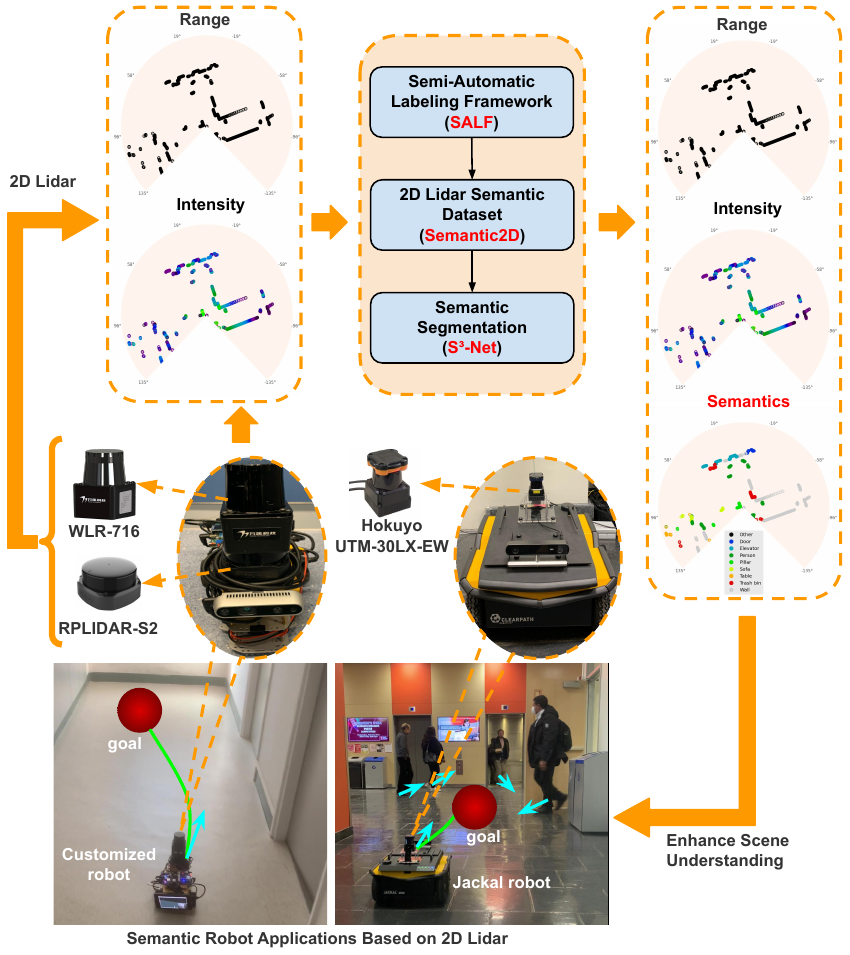}
    \caption{Workflow for 2D lidar semantic segmentation, enabling enhanced scene understanding for mobile robotics.
            }
    \label{fig:semantic2d_overview}
\end{figure}
Semantic scene understanding plays a crucial role in autonomous mobile robots and human-robot interaction systems, as it enables mobile robots to navigate by semantically interpreting the environment in a human-like manner.
It is a prerequisite for various robotic tasks including multi-object detection and tracking~\citep{wen2023semantically}, semantic mapping~\citep{kostavelis2015semantic}, and autonomous navigation~\citep{xie2023drlvo}.
Since cameras and lidar are the most common sensors for mobile robots to perceive their surroundings, semantic segmentation of each image pixel or each lidar point provides a solution for scene understanding. 
While cameras can provide richer human-level semantic information than lidar sensors through various computer vision algorithms (\eg object recognition~\citep{zhao2019object} and scene segmentation~\citep{minaee2021image}), they generate higher-dimensional data and raise more significant privacy concerns.
Lidar sensors, especially 2D lidar, offer a viable alternative for mobile robot applications that require privacy protection, lightweight processing, and lower costs.
Furthermore, compared with camera systems, lidar typically provides more accurate distance measurements and is more robust to poor or changing lighting situations.
However, extracting higher-level information (\eg semantic scene understanding) from 2D lidar data is more challenging due to the lack of publicly available datasets, annotation tools, and limited segmentation algorithms.

To fill these gaps, this article proposes a comprehensive semantic scene understanding workflow for 2D lidar by creating a high-quality 2D lidar semantic segmentation dataset (\ie Semantic2D), designing an efficient Semi-Automatic Labeling Framework for Semantic Annotation (\ie SALSA) with minimal human effort, developing a Stochastic Semantic Segmentation Network (\ie S$^3$-Net) to deliver fine-grained 2D lidar semantic segmentation, and applying the 2D lidar semantic information to enhance various mobile robotics applications (\eg semantic occupancy grid mapping and semantic robot navigation) that require semantic scene understanding, as shown in ~\figref{fig:semantic2d_overview}.
This 2D lidar workflow allows us to re-investigate and improve existing robotics algorithms that use 2D lidar sensors, such as object tracking, mapping, localization, and navigation, by facilitating semantic scene understanding without additional camera sensors.
Specifically, this article presents six primary contributions:

\begin{enumerate}[noitemsep]%[label={(\protect\arabic\perhapsasterisk*)}]
    \item We present the first publicly available  2D lidar semantic segmentation dataset suitable for indoor environments, Semantic2D~\citep{semantic2d_dataset}, which comprises data collected from twelve distinct indoor environments across seven buildings using three types of lidar sensors. 
    
    \item We propose the first 2D semi-automatic labeling framework for semantic annotation, SALSA, to provide fine-grained 2D point-level semantic annotations. It leverages a manually labeled environment map and the Iterative Closest Point (ICP) algorithm to label the raw 2D lidar data with minimal human effort. 
    Researchers can readily use this framework to create and label 2D lidar datasets collected with their own robots and 2D lidar sensors.

    \item We develop a hardware-friendly 2D lidar stochastic semantic segmentation algorithm, S$^3$-Net, that is based on a variational autoencoder (VAE) and can be deployed on resource-constrained robots.
    The algorithm converts raw 2D lidar data (\ie point position, range, and intensity) into a set of input features, with ablation studies used to select features that maximize classification accuracy.
    The output is a fine-grained segmentation for each 2D lidar point, including a stochastic distribution of each point’s segmentation via variational inference techniques.

    \item We validate the ability of our S$^3$-Net to deliver 2D lidar point-level semantic segmentation using the Semantic2D dataset and provide a comprehensive benchmark of segmentation performance against two state-of-the-art geometry-based segmentation algorithms.
    Our results show that S$^3$-Net achieves higher classification accuracy, higher intersection over union, and faster inference speed compared to other coarse-grained algorithms (\eg line extraction~\citep{pfister2003weighted} and leg detection~\citep{bellotto2008multisensor}). 

    \item We explore how our Semantic2D dataset can enhance various mobile robot applications (\eg object tracking, environment mapping, robot localization, and navigation) that require semantic scene understanding.
    Specifically, we first demonstrate its utility in semantic occupancy grid mapping, showing that the dataset provides accurate 2D semantic lidar measurements for building 2D semantic maps.
    We then propose two semantic-aware navigation control policies, called Semantic Pfeiffer and Semantic CNN, based on existing learning-based control policies that use 2D lidar raw range data (\ie Peiffer's policy~\citep{pfeiffer2017perception}) and preprocessed range data (\ie Xie's CNN policy~\citep{xie2021towards}), to improve autonomous navigation in dynamic environments. Through simulated and real-world experiments, we show that our semantically-aware control policies achieve better navigation performance than the original end-to-end approaches without semantic information~\citep{pfeiffer2017perception, xie2021towards}.
    
    \item We open-source our proposed Semantic2D dataset and SALSA labeling framework (\url{https://github.com/TempleRAIL/semantic2d}), the semantic segmentation algorithm S$^3$-Net (\url{https://github.com/TempleRAIL/s3_net}), and the semantic-aware control policy Semantic CNN (\url{https://github.com/TempleRAIL/semantic_cnn_nav}).
    By making these resources available to the robotics community, we aim to advance semantic scene understanding using 2D lidar and inspire improvements in 2D lidar-based tracking, mapping, localization, and navigation algorithms.
    
\end{enumerate}

\section{Related Work}
\label{sec:related_work}
In this section, we provide a detailed description of prior work on semantic datasets, semantic labeling, semantic segmentation, and semantic applications. 

\subsection{Semantic Dataset}
\label{subsec:semantic_dataset}
As summarized by \citet{gao2021we}, numerous high-quality lidar semantic datasets have been released in recent years, including Semantic3D~\citep{hackel2017semantic3d}, KITTI~\citep{geiger2013vision}, SemanticKITTI~\citep{behley2019semantickitti}, Paris-Lille-3D~\citep{roynard2018paris}, and SemanticPOSS~\citep{pan2020semanticposs}.
However, these datasets exclusively focus on semantic segmentation of 3D lidar point cloud data and target outdoor autonomous driving scenarios. 
Recently, \citet{guo2024lidar} introduced LiDAR-Net, a 3D lidar semantic dataset for everyday indoor scenes.
While 2D lidar semantic datasets for indoor mobile robotics could theoretically be extracted from existing 3D datasets, this approach is computationally prohibitive and limits both dataset customization and adaptation to specific 2D lidar sensor characteristics.

\textbf{Contributions}: \: To the best of our knowledge, no publicly available semantic dataset exists for 2D lidar in mobile robotics applications. 
Compared to 3D lidar sensors, 2D lidar offers significant advantages—including lower cost, smaller size, and reduced computational requirements—making them highly suitable for mobile robots operating in 2.5D environments. 
Bridging the gap between 3D and 2D lidar semantic segmentation is therefore of considerable importance. 
To address this, we present the Semantic2D dataset: the first publicly available 2D lidar semantic dataset for mobile robotics, featuring nine categories of typical indoor objects across twelve distinct environments.

\subsection{Semantic Labeling}
\label{subsec:semantic_labeling}
A major challenge in creating semantic lidar datasets is efficiently annotating each point with a class label. 
A direct approach, used for datasets like SemanticKITTI~\citep{behley2019semantickitti}, is to manually label all lidar points using a visual labeling tool. 
However, this process is time-consuming and labor-intensive. To reduce manual effort, some studies employ multimodal sensor setups (\eg adding cameras) and leverage 2D image semantic segmentation to generate 3D lidar labels~\citep{varga2017super, piewak2018boosting}. 
Nevertheless, these approaches require additional sensors and complex calibration procedures. 
To improve labeling efficiency without extra sensors, weakly supervised methods have been proposed. 
For instance, \citet{wei2020multi} introduced a weakly supervised learning technique for 3D point cloud segmentation using only scene- and subcloud-level labels, while \citet{ren20213d} developed WyPR, a framework that generates weak labels to minimize human input. 
Furthermore, \citet{liu2022less} co-designed an efficient 3D lidar annotation pipeline that combines heuristic pre-segmentation with semi-/weakly-supervised learning to significantly reduce manual annotation. 
Despite these advances, all these methods target 3D lidar data and still necessitate substantial manual intervention via visual labeling tools.

While 3D lidar presents relatively clear object shapes that facilitate manual annotation from visualizations, 2D lidar offers less distinguishable features (\eg doors, elevators, and walls all appear as straight lines), making direct human labeling challenging. 
Although geometry-based extraction algorithms can provide annotations for specific objects, such as walls via line extraction~\citep{pfister2003weighted}, people via leg detection~\citep{bellotto2008multisensor}, or vehicles via nearly equidistant beam extraction~\citep{thuy2009non}, they yield only coarse-grained labels for certain object types, rather than fine-grained, point-level annotations.

\textbf{Contributions}: \: Creating a 2D lidar semantic dataset requires an effective and efficient fine-grained labeling framework. 
To address this need, we introduce SALSA, a semi-automatic semantic labeling framework that combines a manually labeled environment map with the Iterative Closest Point (ICP) algorithm. 
This approach minimizes human effort by automatically annotating raw 2D lidar data, providing the fine-grained semantic labels used in the Semantic2D dataset.

\subsection{Semantic Segmentation}
\label{subsec:semantic_segmentation}
Numerous 3D lidar semantic segmentation methods have been developed for autonomous driving scenarios~\citep{yan2024benchmarking}. 
These approaches can be broadly categorized into three groups: point-based segmentation~\citep{qi2017pointnet, qi2017pointnet++, wu2019pointconv}, projection-based segmentation~\citep{wu2019squeezesegv2, xu2020squeezesegv3, milioto2019rangenet++}, and voxel-based segmentation~\citep{graham20183d, han2020occuseg, zhu2021cylindrical, zhang2020polarnet}. 
In contrast, only a limited number of geometry-based algorithms have been proposed for 2D lidar segmentation of specific objects. 
For instance, \citet{pfister2003weighted} developed a weighted line-fitting algorithm to extract linear features from 2D lidar scans, while \citet{bellotto2008multisensor} introduced a laser-based leg detection method that identifies human patterns. 
Similarly, \citet{thuy2009non} presented a vehicle detection algorithm based on distance similarity of reflected beams. Building on this work, \citet{rubio2013connected} proposed a 2D lidar segmentation approach using a Connected Components algorithm to provide coarse-grained segmentation. 
However, these geometry-based methods are limited to coarse-grained segmentation of specific object types (e.g., lines, people, vehicles) and cannot provide fine-grained, point-level semantic segmentation.

\textbf{Contributions}: \: To address the limitations of existing 2D lidar segmentation algorithms, we propose S$^3$-Net, a hardware-friendly stochastic semantic segmentation network based on a Variational Autoencoder (VAE) designed for resource-constrained robots.
Our approach provides fine-grained segmentation for each 2D lidar point, enabling enhanced semantic scene understanding without requiring camera sensors.

\begin{table*}[t]
  \small\sf\centering
  \caption{The detailed configuration of the environments, robots, and sensors}
  \scalebox{0.75}{
  \begin{tabular}{l | l | l | c c c c}
    \toprule
    \textbf{Environment} & \textbf{Robot} & \textbf{Sensor} & 
      \textbf{Range (m)} & \textbf{Horizontal  FOV (\degree)} & \textbf{Angular Resolution (\degree)} & \textbf{\# Points}\\ 
    \midrule
    
    % Temple University, Jackal robot, Hokuyo
    \multirow{2}{*}{Temple University} 
      & \multirow{2}{*}{Jackal robot} 
        & Hokuyo UTM-30LX-EW lidar 
        & [0.1, 60] & 270 & 0.25 & 1,081\\ 
      & & ZED 2 stereo camera 
        & [0.3, 20] & 110 & —  & — \\ 
    \midrule

    % HKU, Customized robot, WLR-716 
    \multirow{3}{*}{The University of Hong Kong} 
      & \multirow{3}{*}{Customized robot} 
        & WLR-716 lidar 
        & [0.15, 25] & 270 & 0.33 & 811 \\ 
      & & RPLIDAR-S2 lidar 
        & [0.2, 30] & 360 & 0.18 & 1,972 \\ 
      & & Intel RealSense D435 camera 
        & [0.3, 3] & 85.2 & — & — \\

    \bottomrule
  \end{tabular}
  }
  \label{tab:sensor_config}
\end{table*}

\subsection{Semantic Application}
\label{subsec:semantic_application}
Extracting semantic information from 2D lidar data makes it possible to use that information in downstream applications, such as multi-object tracking, semantic mapping, semantic localization, and semantic navigation.
Previously, 2D lidar-based object tracking works could only detect and track one type of specific objects based on their specific geometry shapes, such as pedestrians~\citep{bellotto2008multisensor, chen2019pedestrian} or vehicles~\citep{thuy2009non}.
Using our proposed 2D lidar semantic segmentation algorithm (\ie S$^3$-Net), 2D lidar-based object tracking algorithms can detect and track different types of objects.
Similarly, while existing semantic mapping works~\citep{ma2017multi, zhang2018semantic, chaplot2020semantic} require the use of additional RGB-D/depth cameras or 3D lidar to provide semantic information, there is still a gap in traditional 2D lidar semantic mapping.

\textbf{Contributions}: \: Our work aims to bridge this gap and show how the proposed 2D lidar semantic segmentation work can semantically label occupancy grid maps, generate semantic occupancy grid maps, and perform semantic localization. 
In addition, since 2D lidar is the key perception sensor for mobile robot navigation, many mature navigation control policies~\citep{pfeiffer2017perception, fan2020distributed, guldenring2020learning, xie2021towards, xie2023drlvo} use 2D lidar data as input.
However, due to the lack of 2D lidar semantic segmentation algorithms, they could not previously utilize the benefits that semantic information provides.
To bridge this gap, we propose two improved semantic-aware navigation control policies (\ie Semantic Pfeiffer and Semantic CNN) based on pre-existing 2D lidar-based navigation policies~\citep{pfeiffer2017perception, xie2021towards}, respectively.

\begin{figure*}[t]
    \centering
    \subfloat[Engineering lobby, Temple]{
            \centering
            \includegraphics[width=0.23\textwidth]{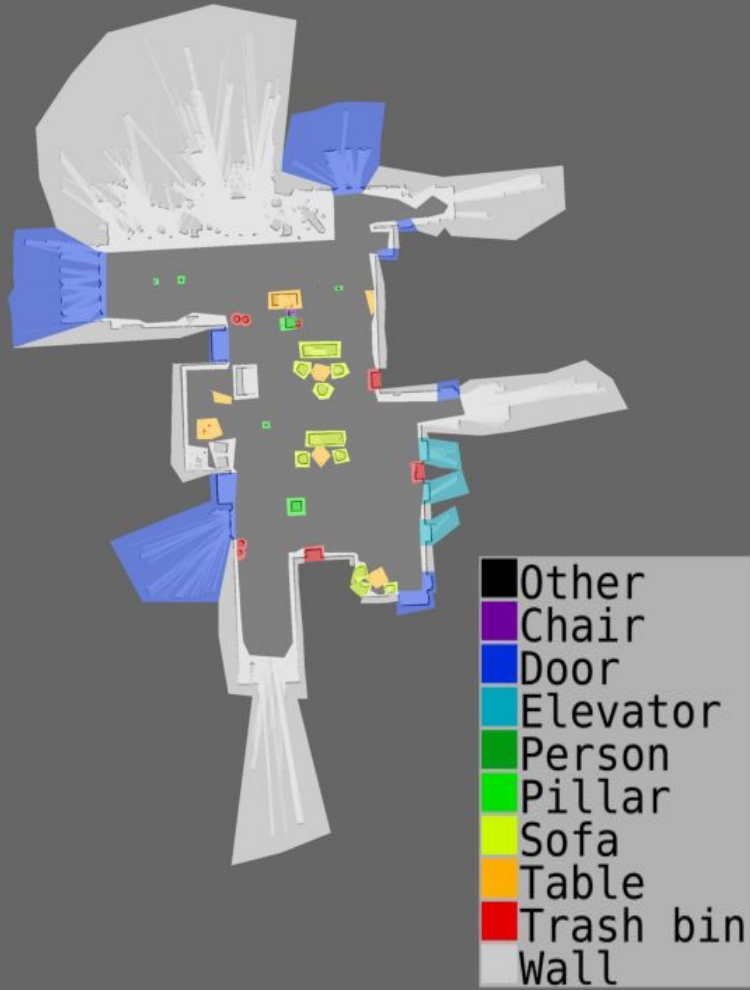}
            \label{fig:lobby}
    }%
    %\vspace{0.01cm}
    \subfloat[Engineering corridor, Temple]{
            \centering
            \includegraphics[width=0.23\textwidth]{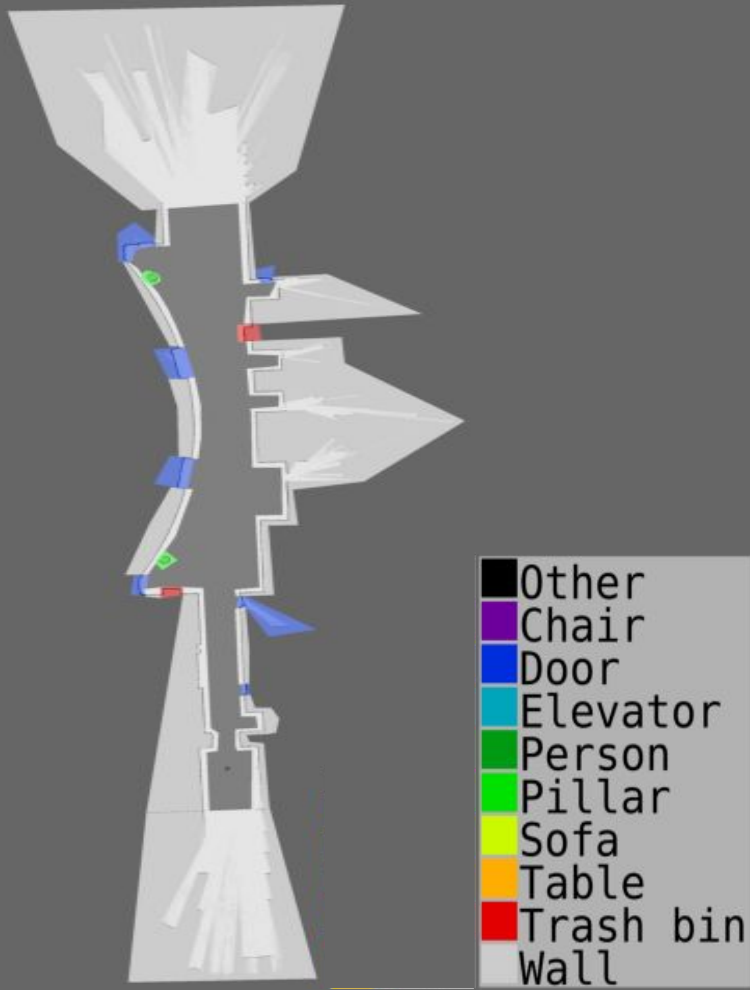}
            \label{fig:corridor}
    }%
    \subfloat[Engineering 4th-floor, Temple]{
            \centering
            \includegraphics[width=0.23\textwidth]{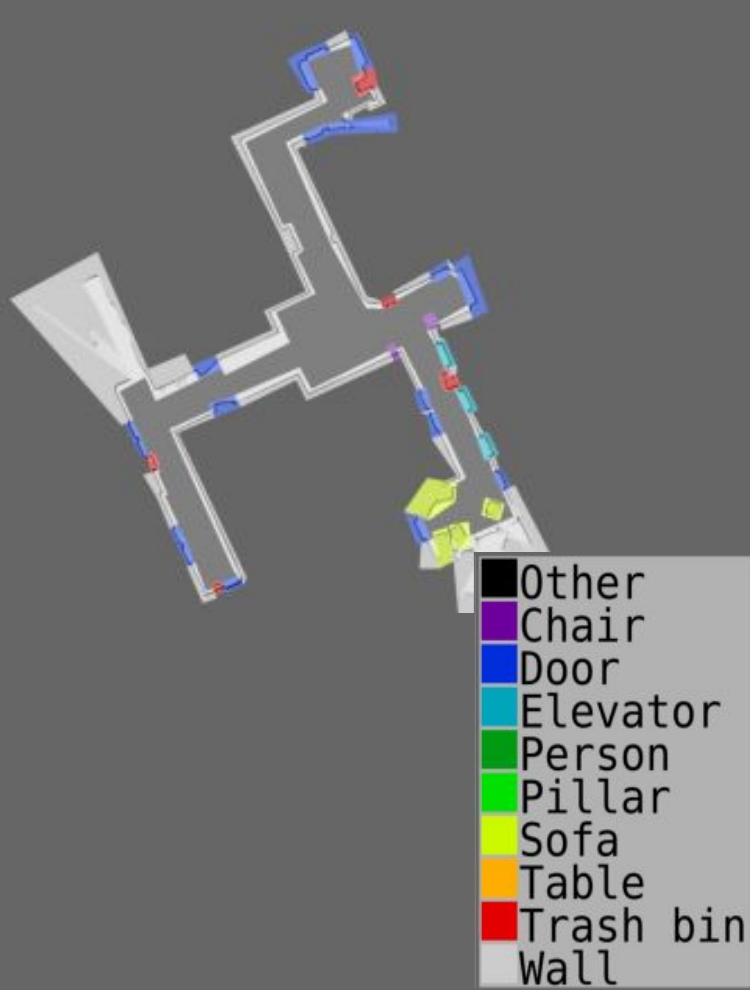}
            \label{fig:4th_floor}
    }%
    \subfloat[Engineering 6th-floor, Temple]{
            \centering
            \includegraphics[width=0.23\textwidth]{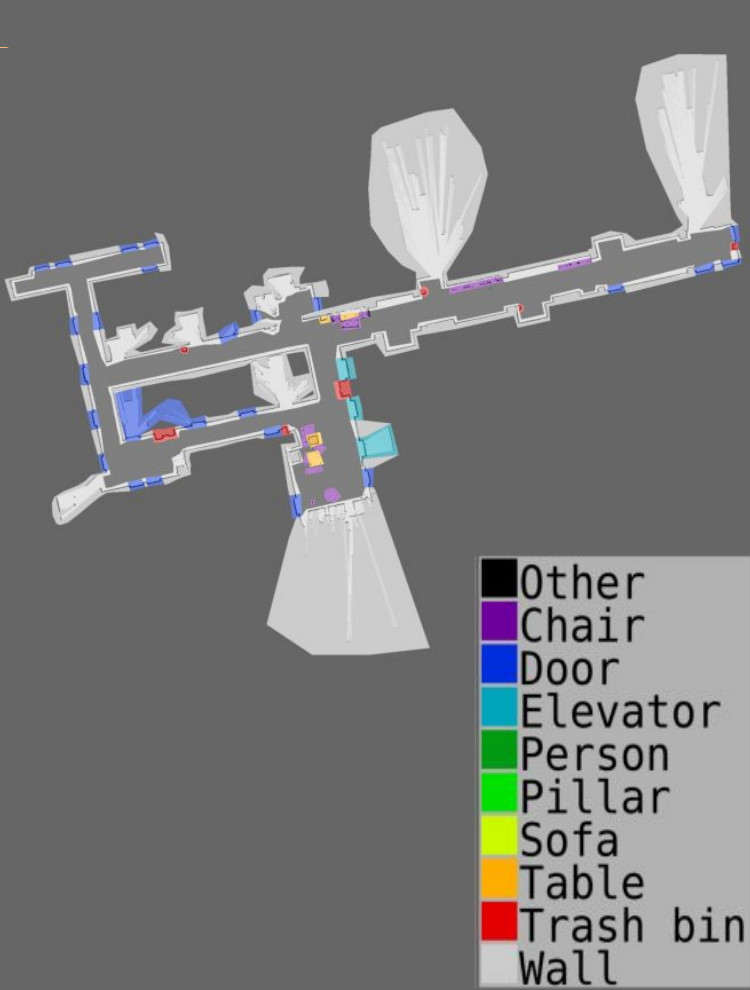}
            \label{fig:6th_floor}
    }%
    \vspace{0.01cm}
    \subfloat[Engineering 8th-floor, Temple]{
            \centering
            \includegraphics[width=0.23\textwidth]{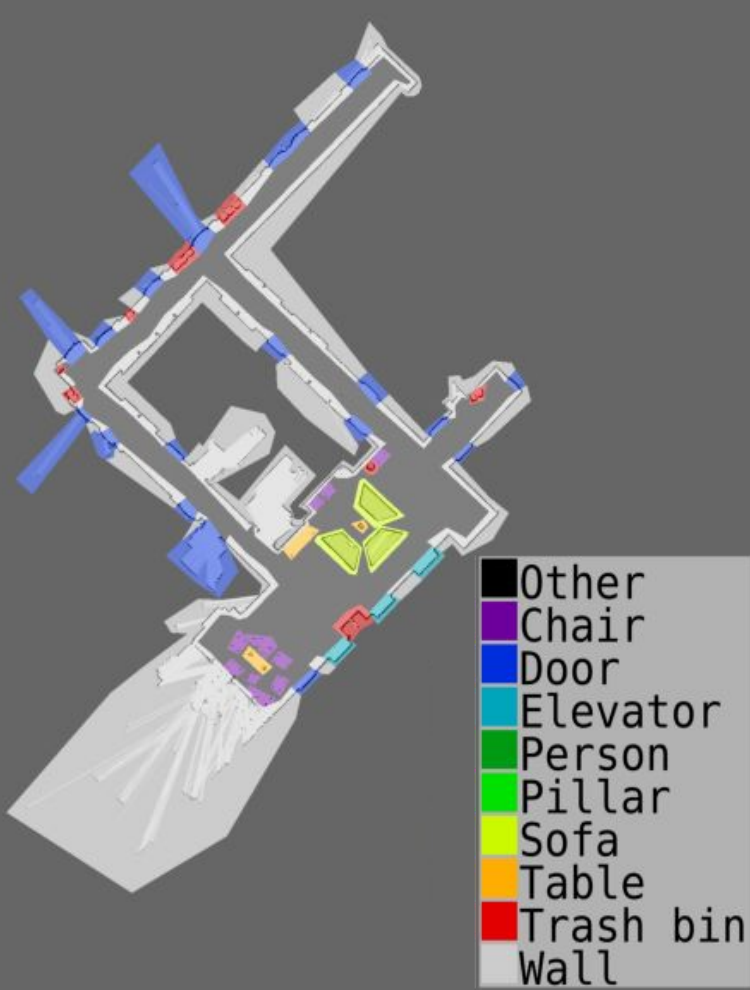}
            \label{fig:8th_floor}
    }%
    \subfloat[Engineering 9th-floor, Temple]{
            \centering
            \includegraphics[width=0.23\textwidth]{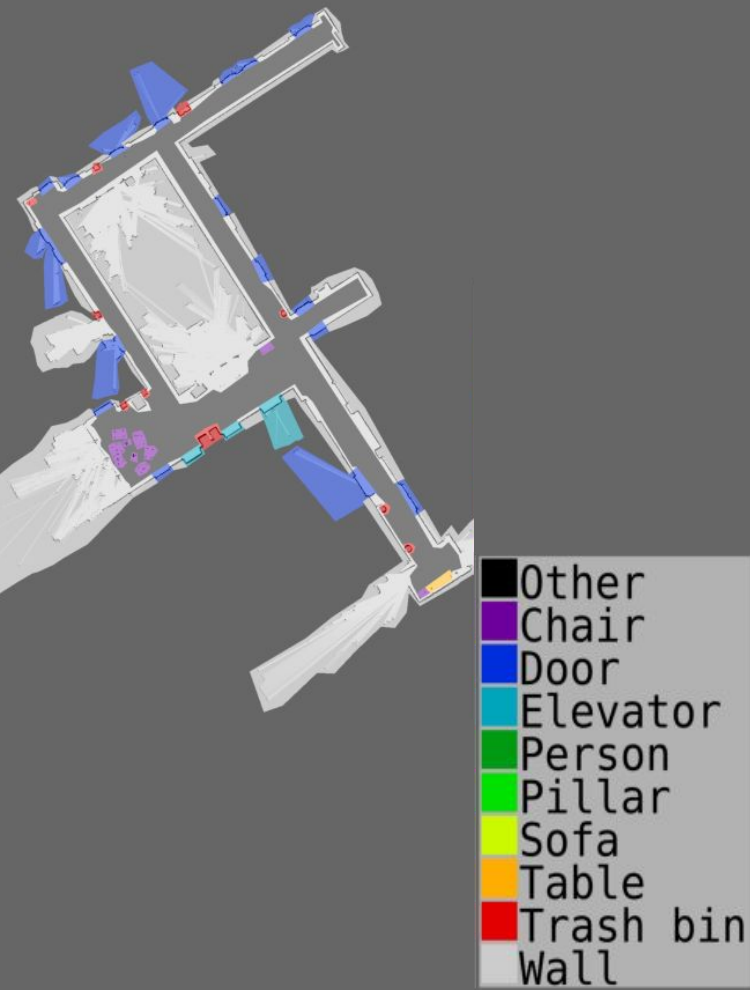}
            \label{fig:9th_floor}
    }%
    \subfloat[SERC lobby, Temple]{
            \centering
            \includegraphics[width=0.23\textwidth]{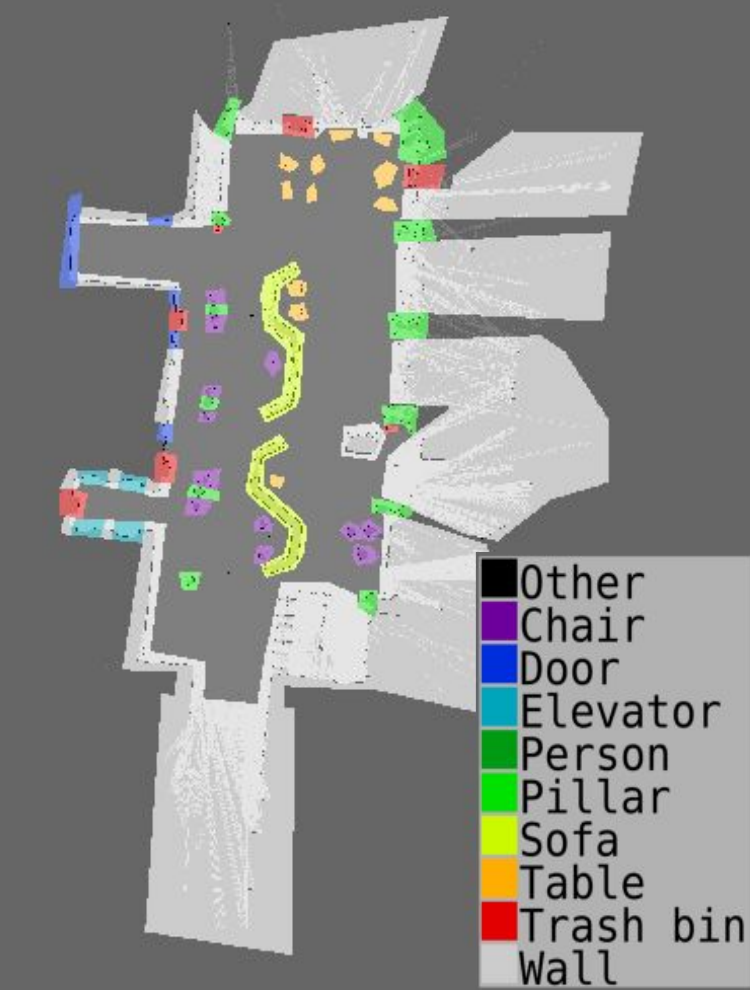}
            \label{fig:serc}
    }%
    \subfloat[Gladfelter lobby, Temple]{
            \centering
            \includegraphics[width=0.23\textwidth]{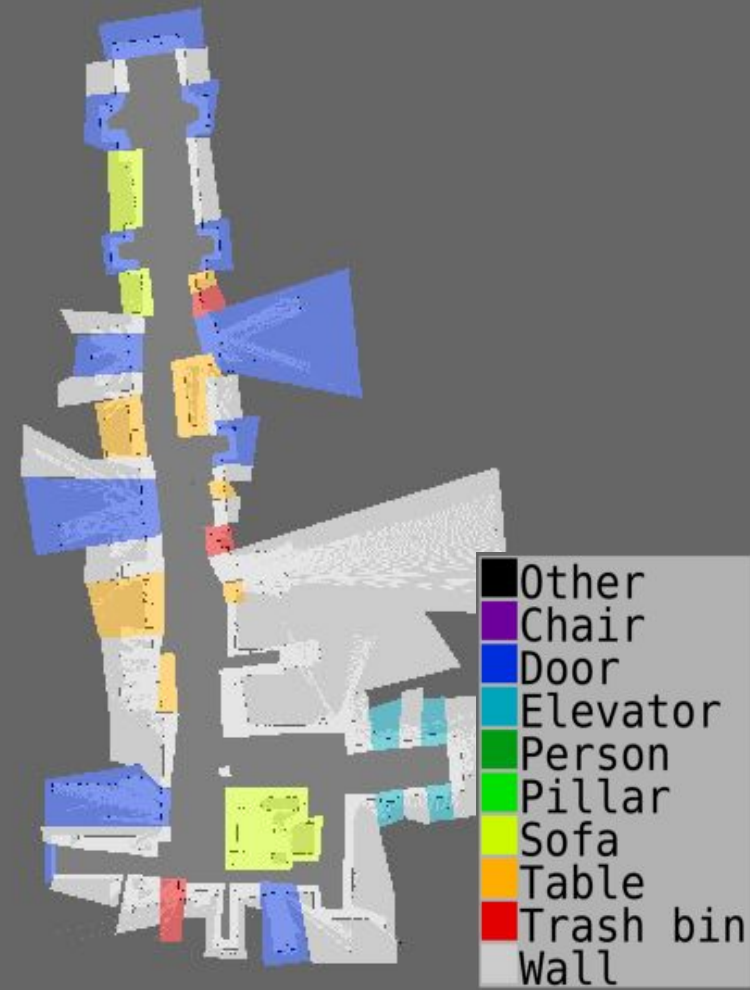}
            \label{fig:gladfelter}
    }%
    \vspace{0.01cm}
    \subfloat[Mazur lobby, Temple]{
            \centering
            \includegraphics[width=0.23\textwidth]{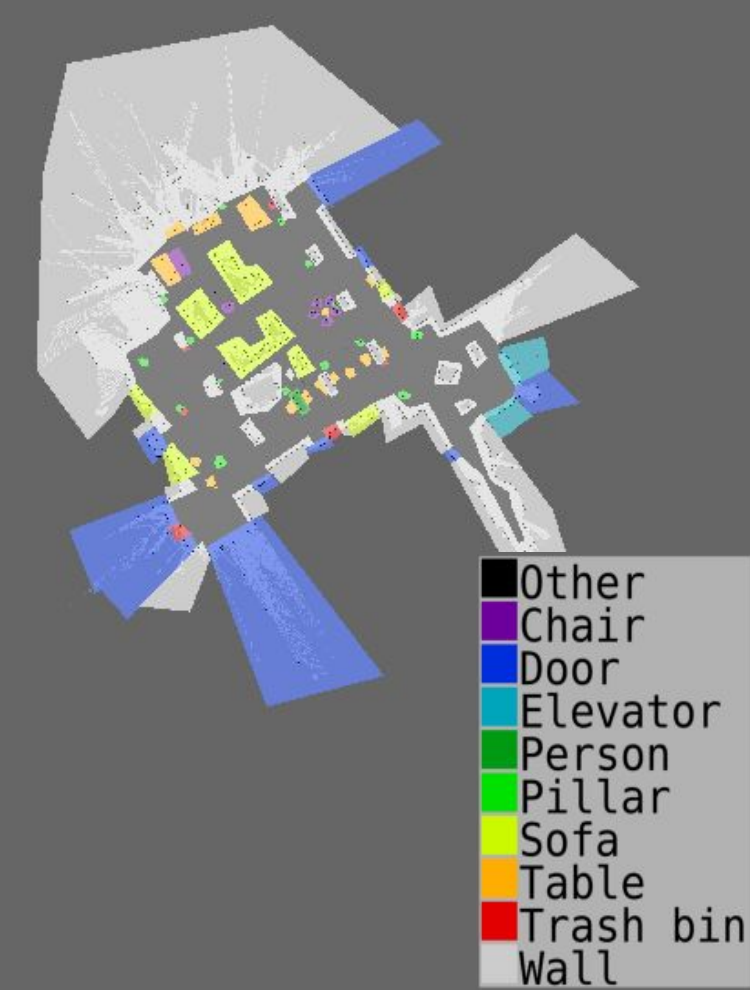}
            \label{fig:mazur}
    }%
    \subfloat[Chow Yei Ching 4th-floor, HKU]{
            \centering
            \includegraphics[width=0.23\textwidth]{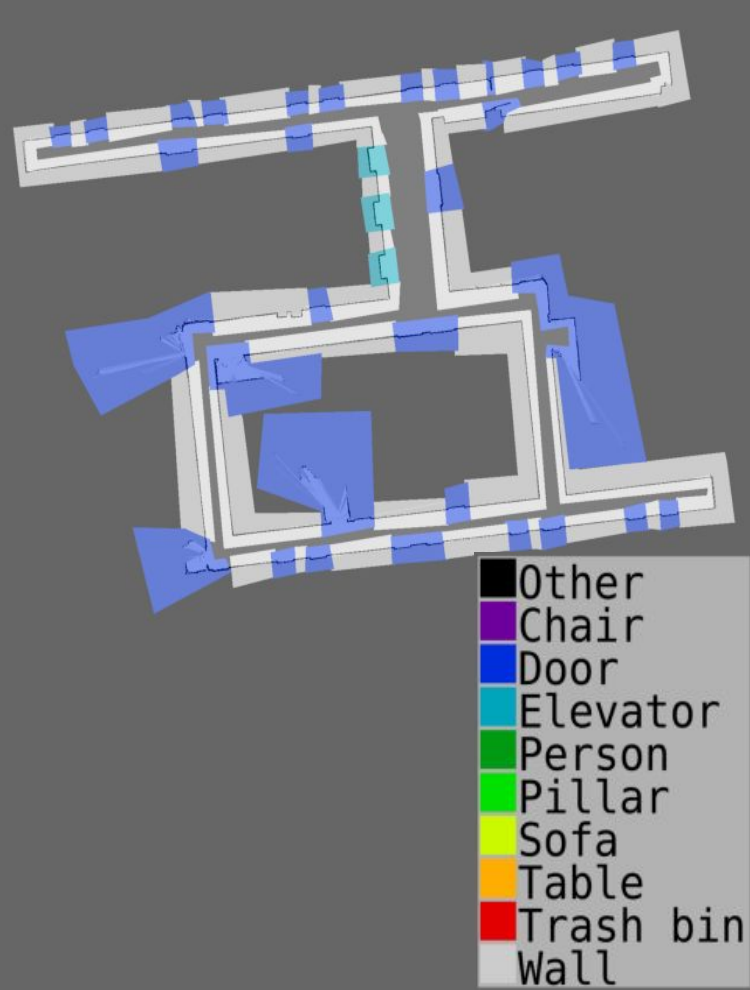}
            \label{fig:cyc}
    }%
    \subfloat[Jockey Club 3rd-floor, HKU]{
            \centering
            \includegraphics[width=0.23\textwidth]{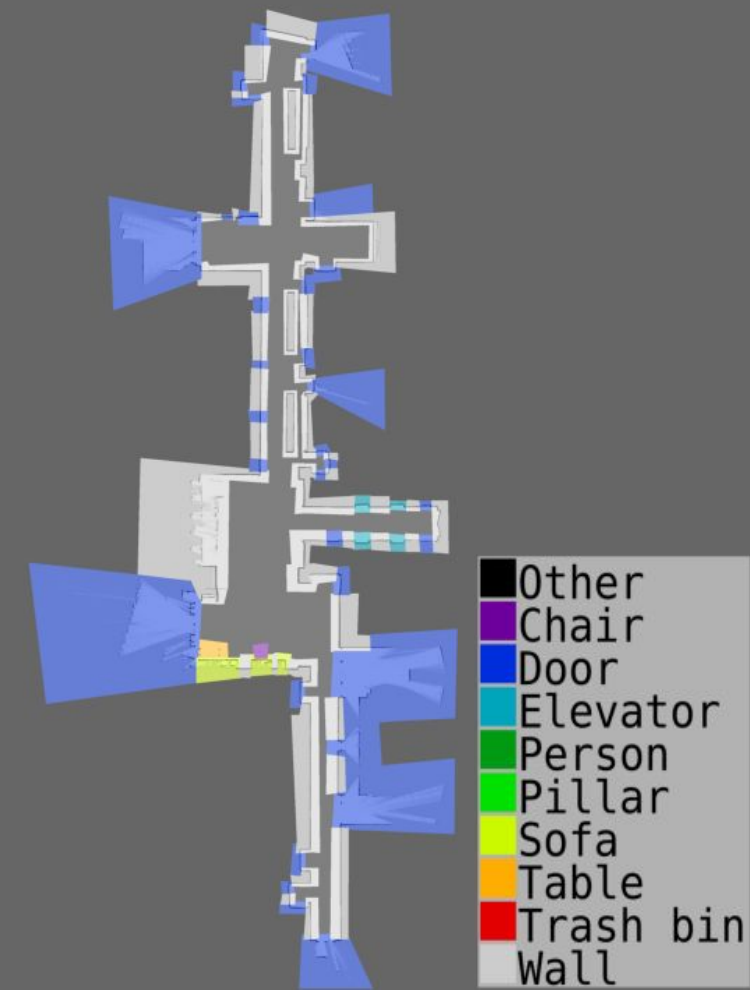}
            \label{fig:jcb}
    }%
    \subfloat[Centennial Campus lobby, HKU]{
            \centering
            \includegraphics[width=0.23\textwidth]{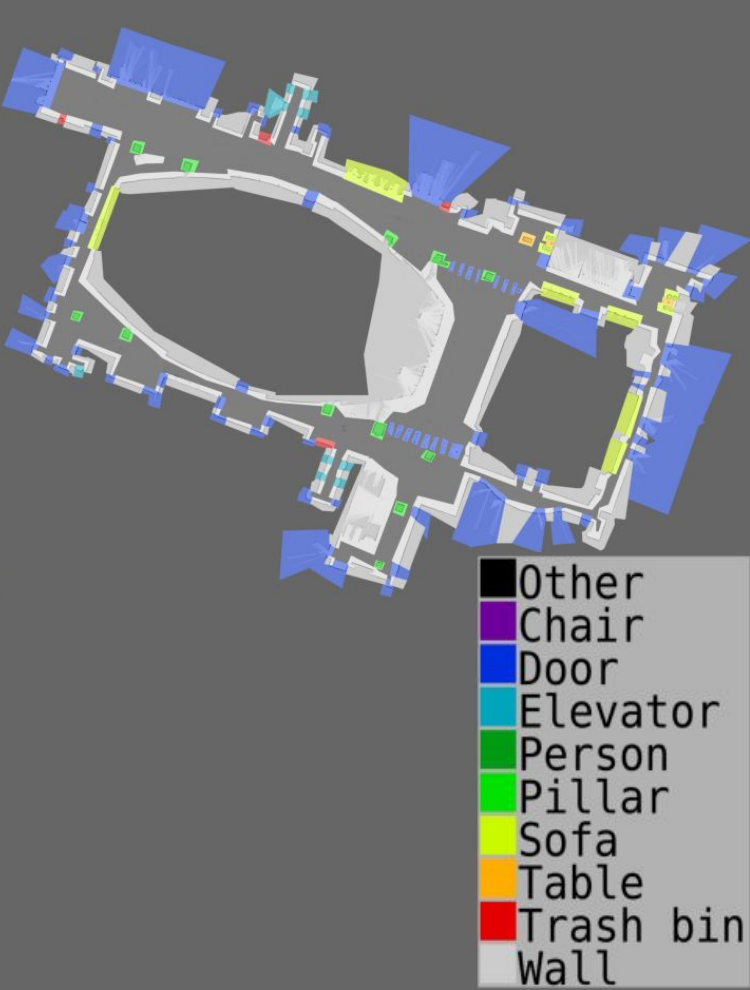}
            \label{fig:centennial}
    }%
    \caption{Floor plans of the dataset collection environments, depicting nine indoor settings across four buildings at Temple University and three additional environments from three buildings at the University of Hong Kong.
    }
    \label{fig:floorplan}
\end{figure*}

%%%%%%%%%%%%%%%%%%%%%%%%%%%%%%%%%%%%%%%%%%%%%%%%%%%%%%%%%%%%%%%%%%%%%%%%%%%%%%%%
\section{Semantic2D Dataset}
\label{sec:semantic2d_dataset}
This section introduces our Semantic2D dataset, a 2D lidar semantic dataset for mobile robotic applications, and SALSA, our semi-automatic labeling framework for semantic annotation. 
We also demonstrate how our semantic labeling tools can be applied to other public 2D lidar datasets. Finally, we examine the limitations of both the Semantic2D dataset and the SALSA framework.

\subsection{Semantic2D}
\label{subsec:semantic2d}
The Semantic2D dataset was collected using two distinct robot platforms equipped with three different lidar sensors. 
\tabref{tab:sensor_config} summarizes the key characteristics of these robots and sensors. Data acquisition spanned twelve indoor environments across seven buildings on two university campuses, as illustrated in \figref{fig:semantic2d_overview}.

\subsubsection{Dataset Collection}
\label{subsubsubsec:dataset_collection}
Data collection employed teleoperation: a PS4 joystick controlled a Clearpath Jackal robot with a Hokuyo UTM-30LX-EW lidar through nine environments across four buildings at Temple University.
Separately, a PS3 joystick maneuvered a customized robot platform equipped with both a WLR-716 and RPLIDAR-S2 lidar through three environments across three buildings at the University of Hong Kong. 
Please see \figref{fig:floorplan} for floorplan details. All environments contained naturally moving pedestrians. 
Each location was pre-mapped using the \texttt{gmapping} ROS package prior to data collection.\footnote{While \texttt{gmapping}'s underlying SLAM algorithm necessitated this approach, alternative pose graph SLAM methods could potentially correct full pose histories.}

\subsubsection{Dataset Content}
\label{subsubsubsec:dataset_content}
During teleoperation, we captured 2D lidar scans (range and intensity data), occupancy maps, and robot poses (from the \texttt{amcl} ROS package). The dataset comprises \unit[131]{minutes}  of raw sensor data recorded at \unit[20]{Hz} across twelve indoor environments, totaling \unit[188,007]{data tuples}. 
These were partitioned into training (70\%), validation (10\%), and testing (20\%) subsets. To prevent environmental or temporal bias, data from each scene was split according to these ratios before being combined, with scene-level splits performed uniformly at random.

\subsubsection{Dataset Statistics}
\label{subsubsubsec:dataset_statistics}
\begin{figure}[t]
    \centering
    \includegraphics[width=0.46\textwidth]{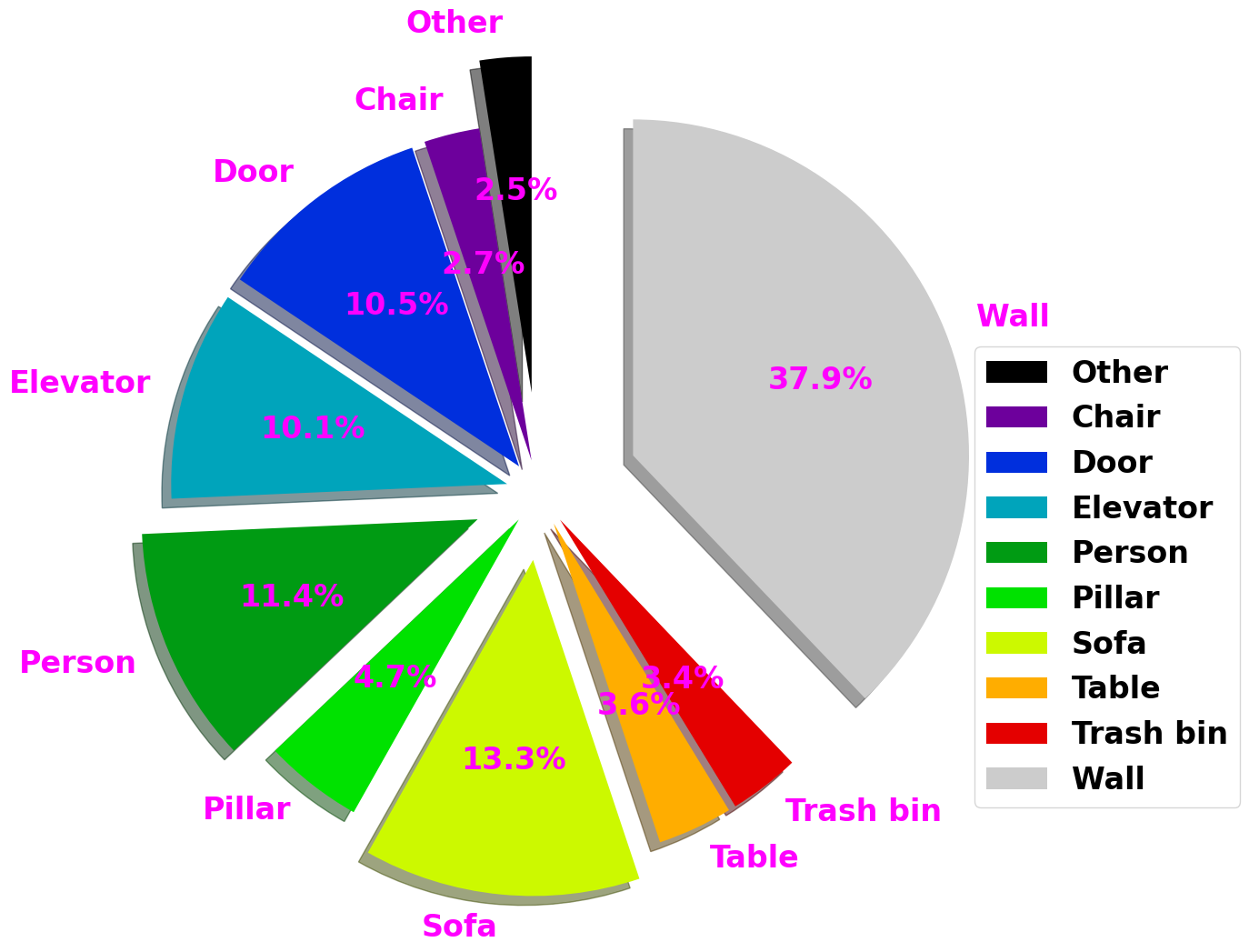}
    \caption{The percentage of each class in the Semantic2D dataset.
            }
    \label{fig:class_percentage}
\end{figure}

As shown in \figref{fig:floorplan}, Semantic2D features annotations for nine common indoor object categories: chairs, doors, elevators, persons, pillars, sofas, tables, trash cans, and walls, with unclassified objects labeled as ``Other''. 
Class distribution analysis (\figref{fig:class_percentage}) reveals walls (37.9\%), sofas (13.3\%), persons (11.4\%), and doors (10.5\%) as the predominant categories, while the ``Other'' class constitutes less than 3\% of the dataset.

\subsubsection{Additional Data}
Although not utilized for 2D lidar segmentation, we captured supplementary sensor data including IMU readings, RGB/depth images (from Stereolabs ZED2 or Intel RealSense D435 cameras), odometry, pedestrian tracking (via Zed2 driver), and joystick velocity commands. 
These \texttt{rosbag}-recorded streams may facilitate research on robot control or navigation tasks. We additionally recorded nominal paths to predefined waypoints, computed by the \texttt{move\_base} ROS node.

\begin{figure*}[t]
    \centering
    \includegraphics[width=0.99\textwidth]{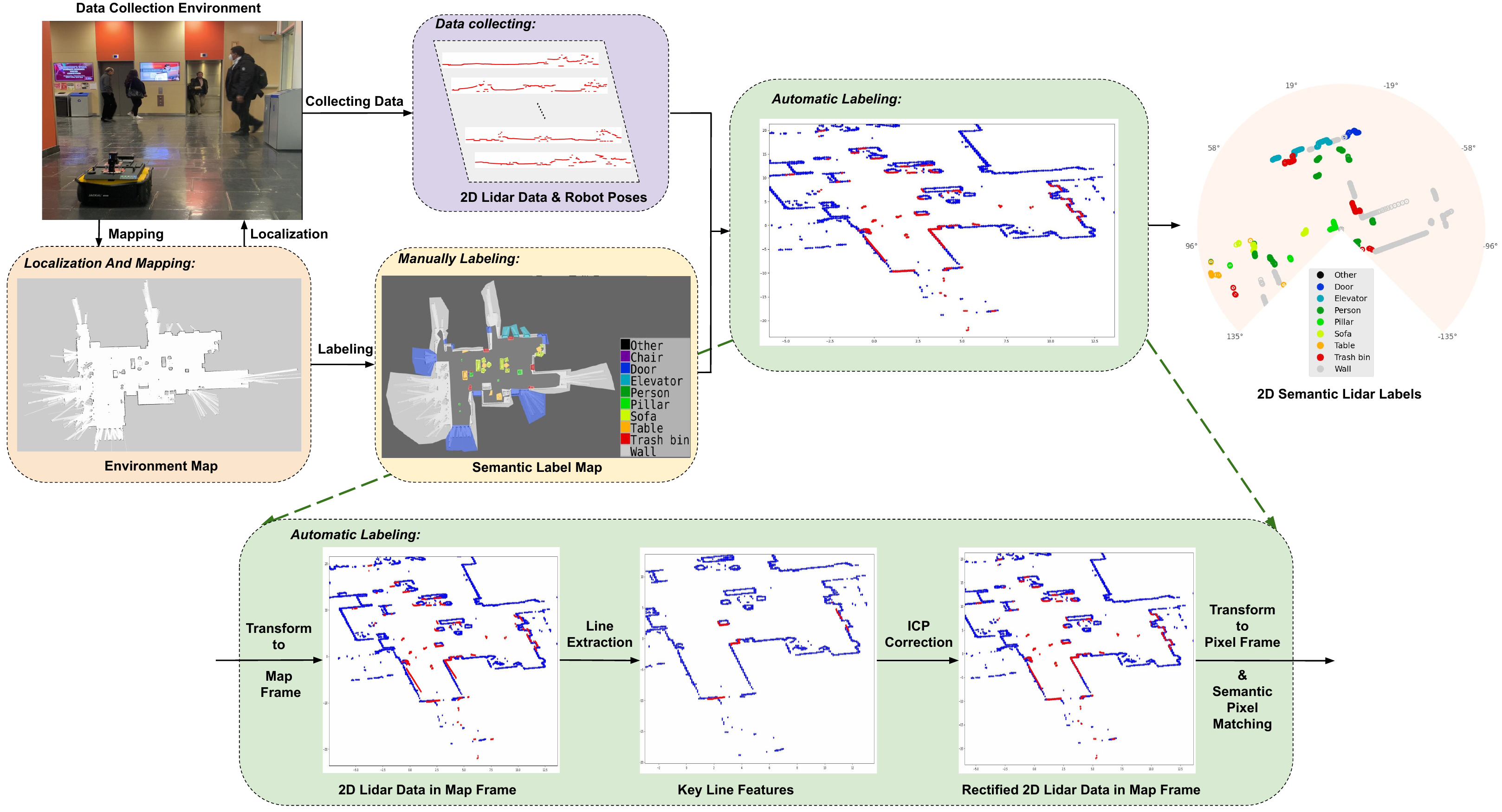}
    \caption{SALSA: a semi-automatic semantic labeling framework for the Semantic2D dataset that only requires manual labeling of a pre-mapped environment.
            }
    \label{fig:semi_automatic_labeling}
\end{figure*}

\subsection{Semi-Automatic Labeling Framework}
\label{subsec:semi_automated_labeling_framework}
With over 300 million lidar points in our Semantic2D dataset, manual labeling is infeasible. We therefore developed SALSA (Semi-Automatic Labeling framework for Semantic Annotation), outlined in \figref{fig:semi_automatic_labeling}, which relies on an accurate initial semantic environment map and precise alignment between this map and individual lidar scans. 
This approach significantly reduces labeling effort while maintaining high-quality results.

To create the initial semantic environment map for each scene, we annotate the pre-mapped occupancy grid map (used for data collection) with the LabelMe tool~\citep{torralba2010labelme}. 
The resulting manually labeled semantic maps are shown in \figref{fig:floorplan}.
We then align lidar scans with these semantic environment maps to assign labels to each scan point. 
Points not aligning with mapped structures are labeled as ``Person,''\footnote{This approach is valid because people were the only moving objects in our data collection environments.} leveraging the fact that misalignments typically correspond to dynamic obstacles.

\begin{figure*}[t]
    \centering
    \subfloat[Engineering lobby, Temple]{
            \centering
            \includegraphics[width=0.24\textwidth]{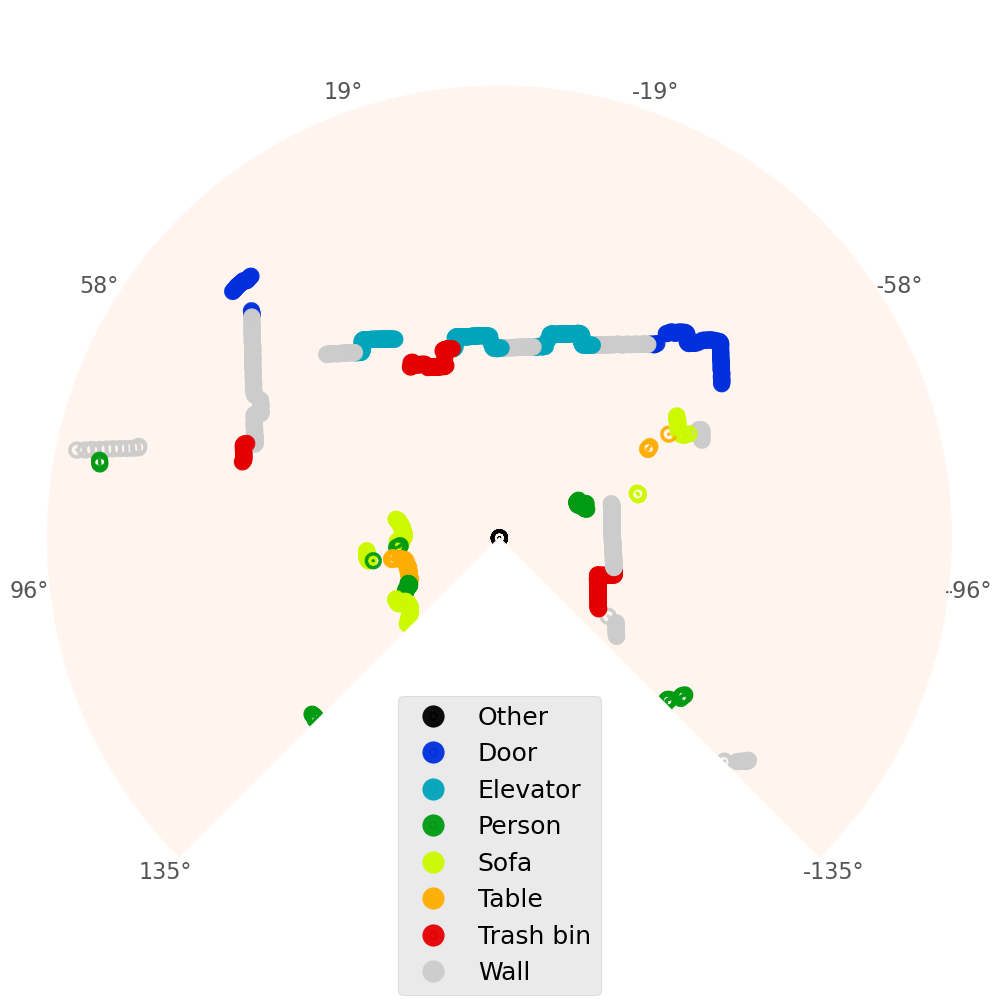}
            \label{fig:lobby_showcase}
    }%
    %\vspace{0.01cm}
    \subfloat[Engineering corridor, Temple]{
            \centering
            \includegraphics[width=0.24\textwidth]{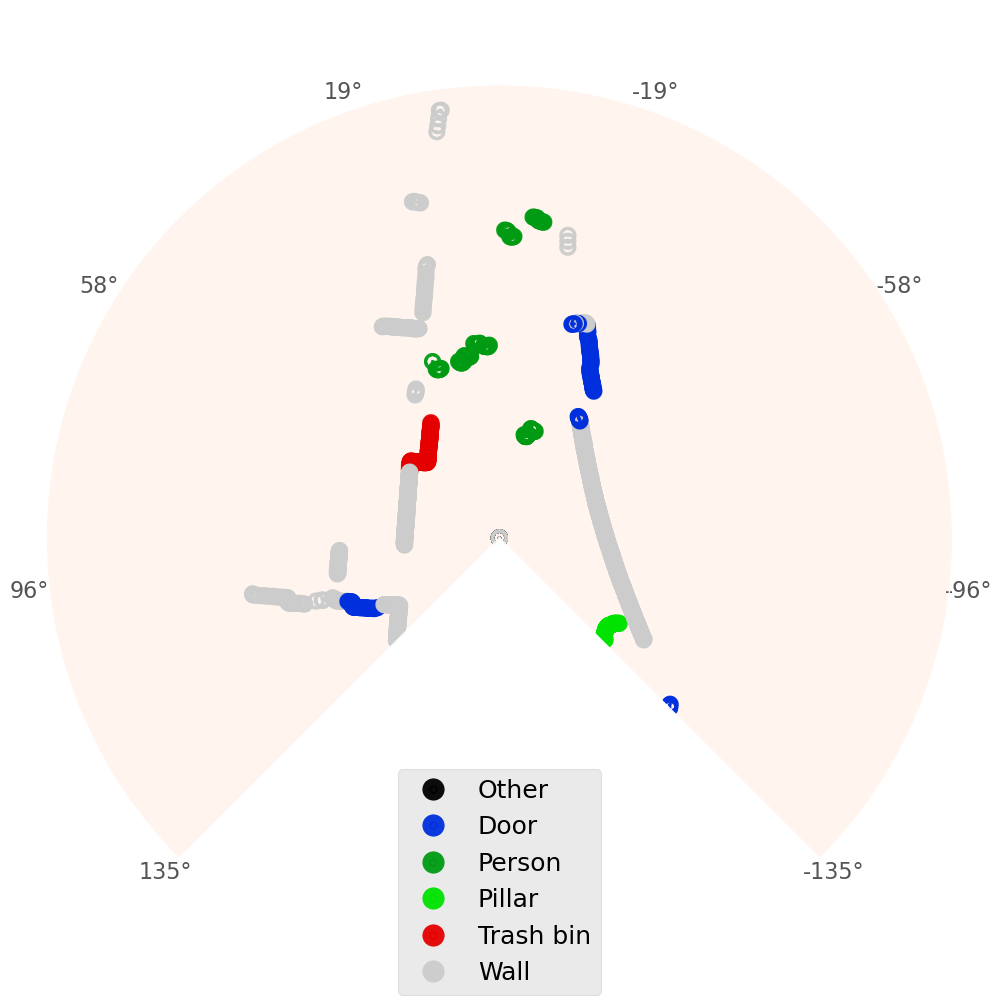}
            \label{fig:corridor_showcase}
    }%
    \subfloat[Engineering 4th-floor, Temple]{
            \centering
            \includegraphics[width=0.24\textwidth]{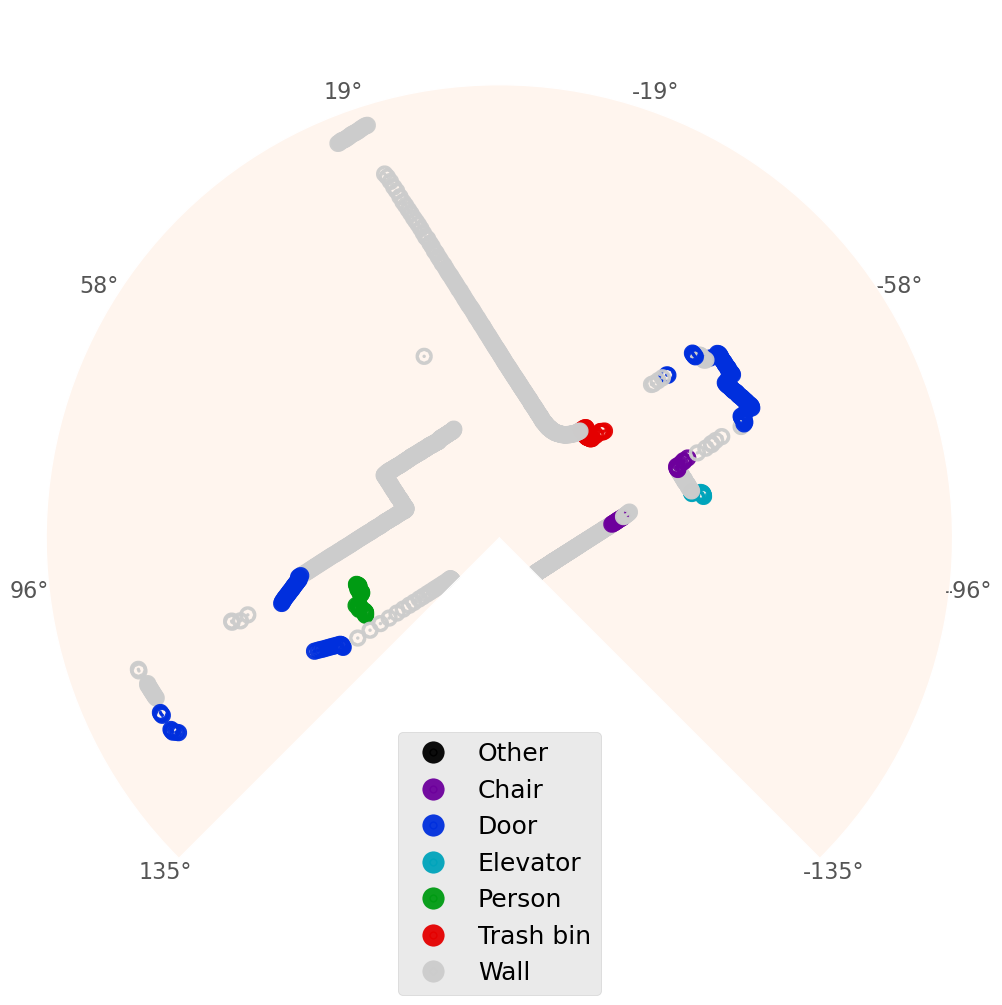}
            \label{fig:4th_floor_showcase}
    }%
    \subfloat[Engineering 6th-floor, Temple]{
            \centering
            \includegraphics[width=0.24\textwidth]{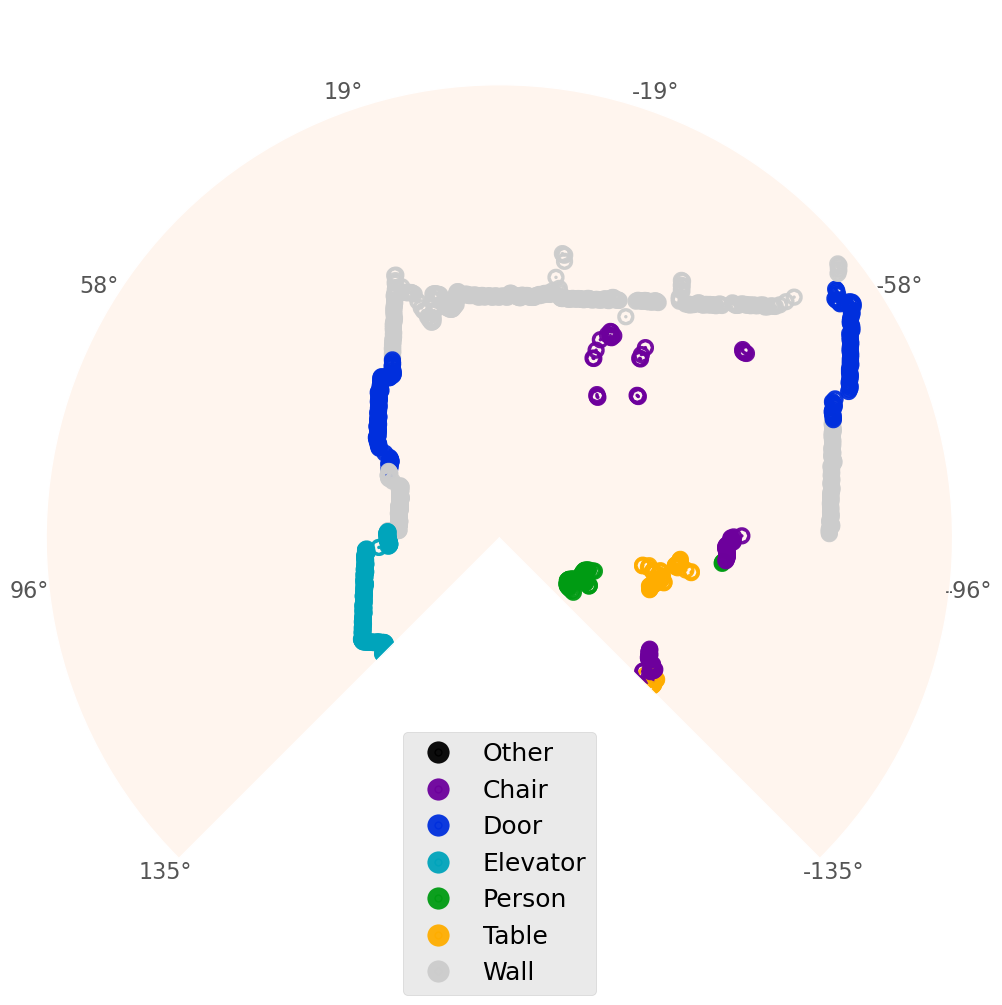}
            \label{fig:6th_floor_showcase}
    }%
    \vspace{0.01cm}
    \subfloat[Engineering 8th-floor, Temple]{
            \centering
            \includegraphics[width=0.24\textwidth]{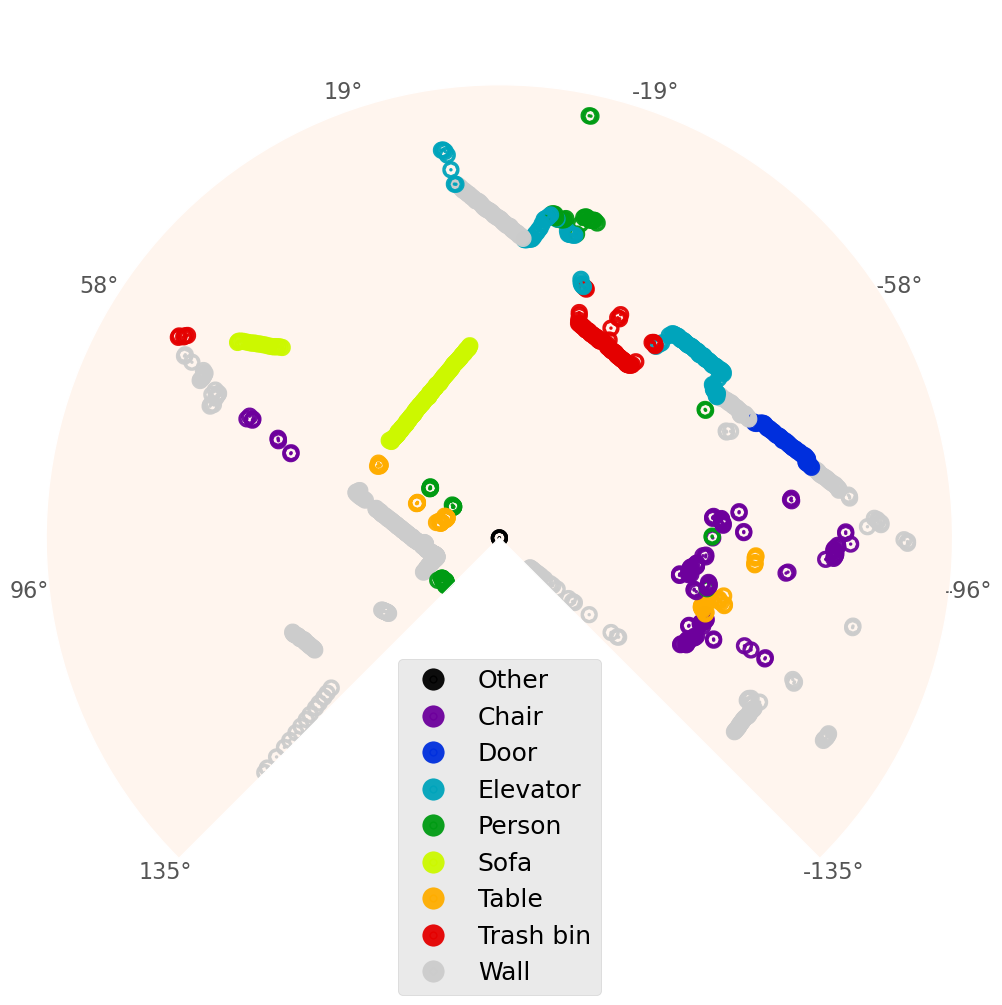}
            \label{fig:8th_floor_showcase}
    }%
    \subfloat[Engineering 9th-floor, Temple]{
            \centering
            \includegraphics[width=0.24\textwidth]{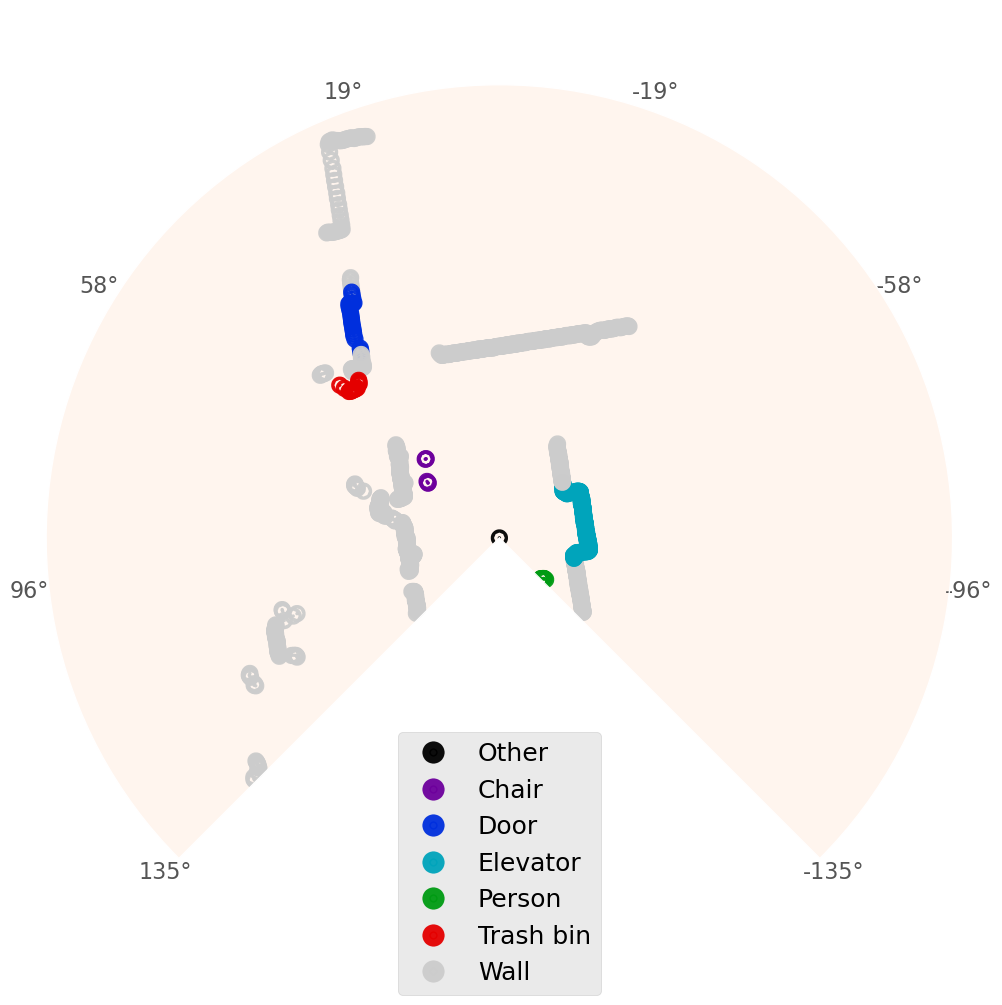}
            \label{fig:9th_floor_showcase}
    }%
    \subfloat[SERC lobby, Temple]{
            \centering
            \includegraphics[width=0.24\textwidth]{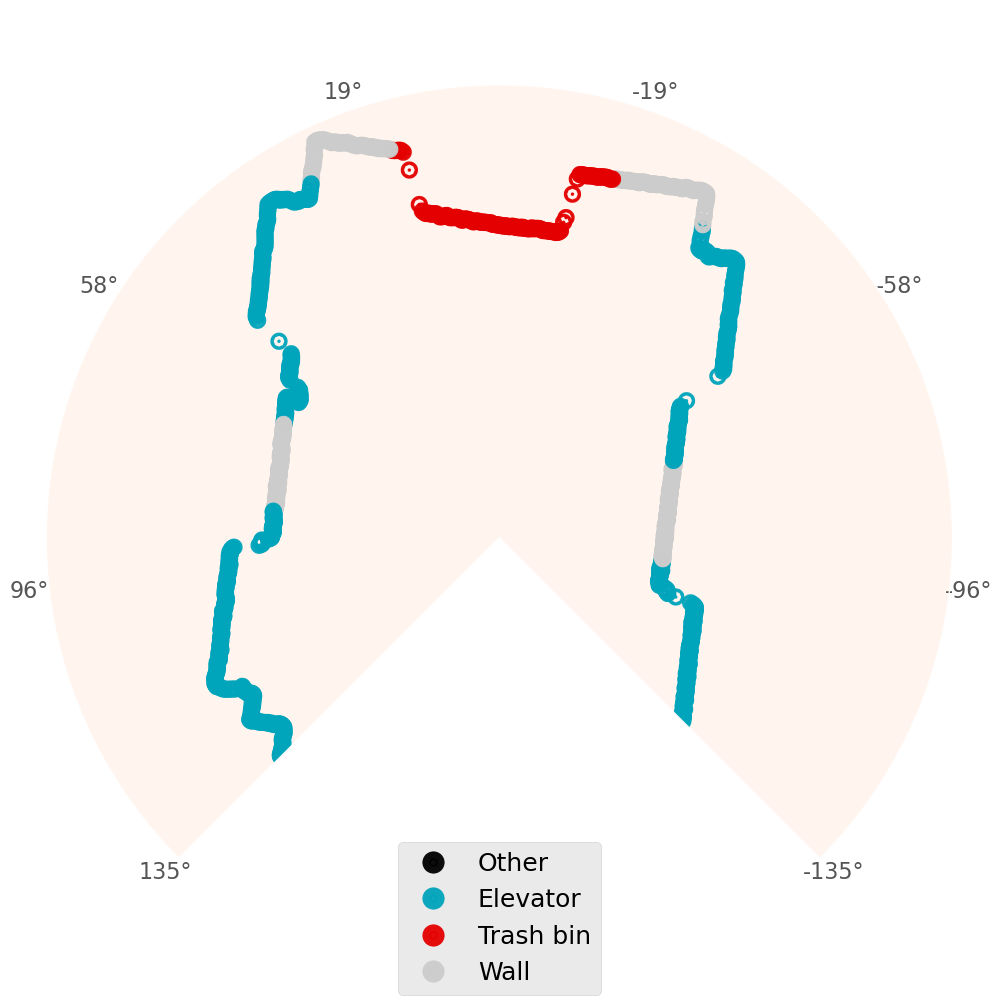}
            \label{fig:serc_showcase}
    }%
    \subfloat[Gladfelter lobby, Temple]{
            \centering
            \includegraphics[width=0.24\textwidth]{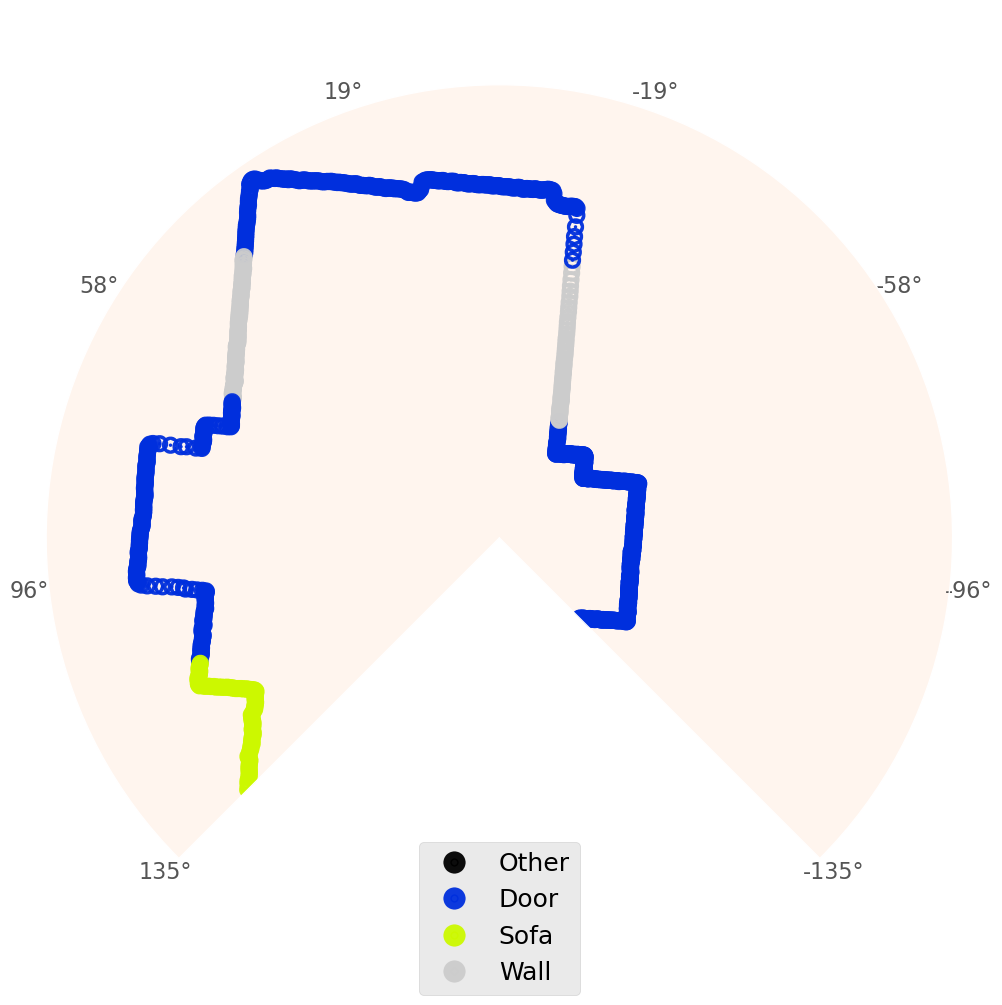}
            \label{fig:gladfelter_showcase}
    }%
    \vspace{0.01cm}
    \subfloat[Mazur lobby, Temple]{
            \centering
            \includegraphics[width=0.24\textwidth]{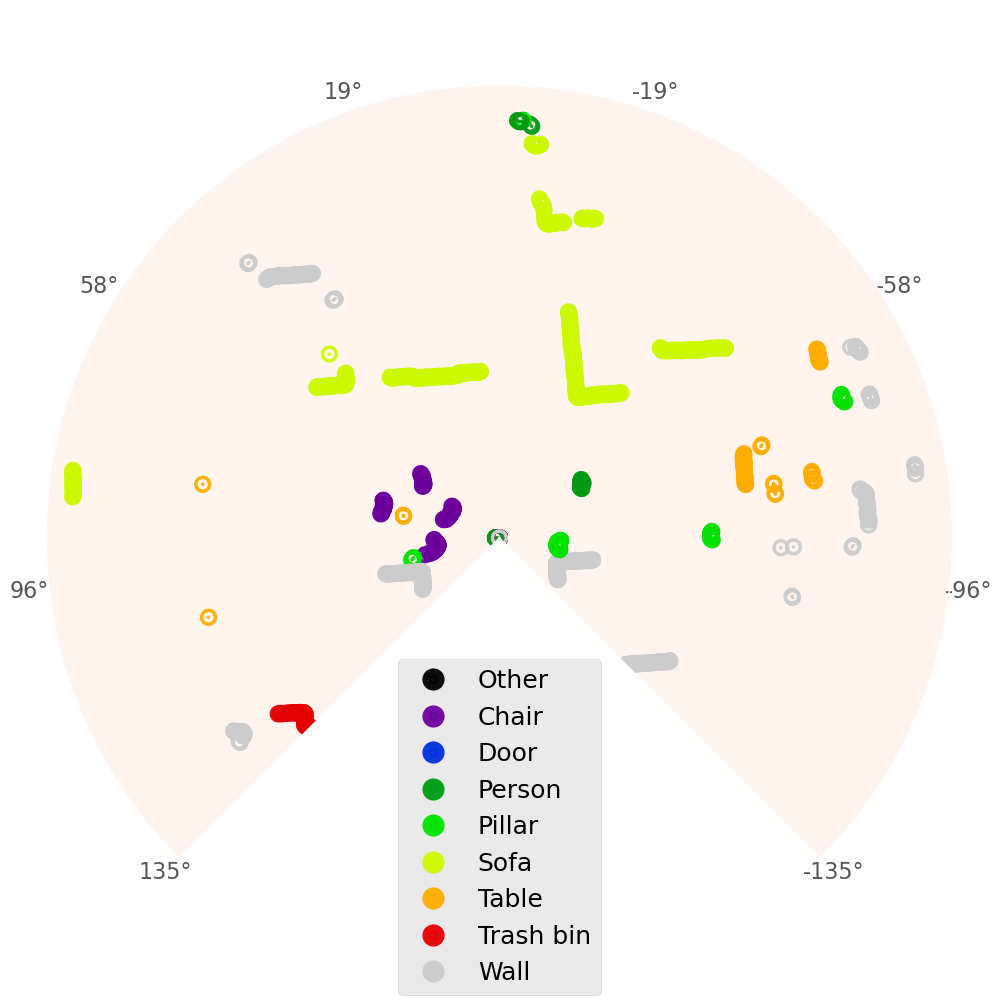}
            \label{fig:mazur_showcase}
    }%
    % \vspace{0.01cm}
    \subfloat[Chow Yei Ching 4th-floor, HKU]{
            \centering
            \includegraphics[width=0.24\textwidth]{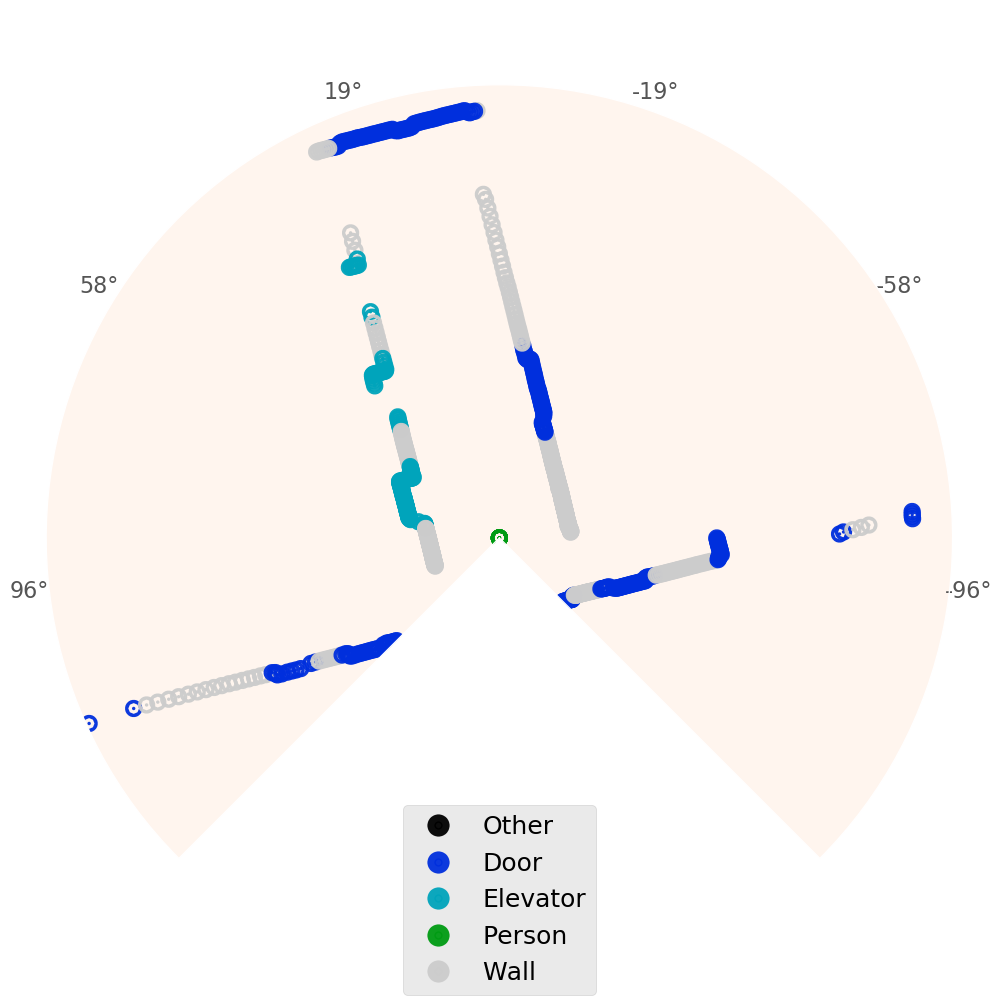}
            \label{fig:cyc_showcase}
    }%
    \subfloat[Jockey Club 3rd-floor, HKU]{
            \centering
            \includegraphics[width=0.24\textwidth]{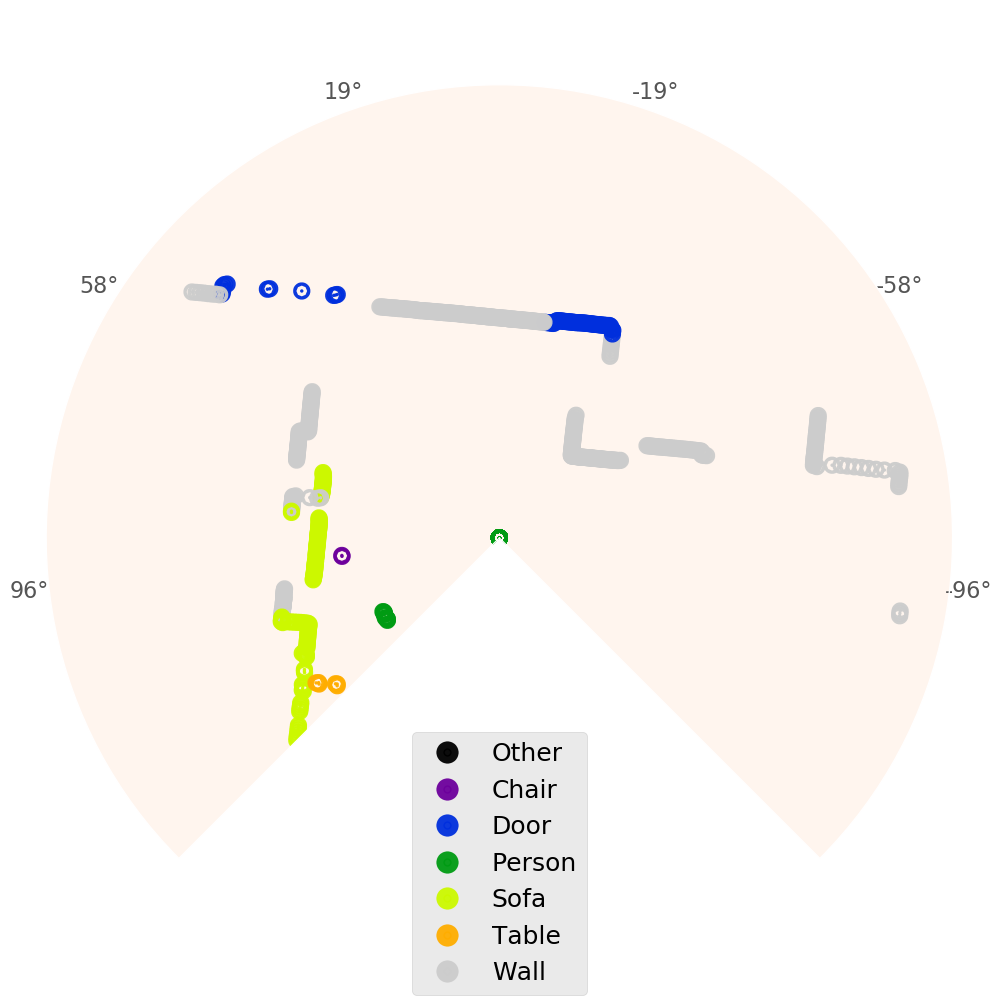}
            \label{fig:jcb_showcase}
    }%
    \subfloat[Centennial Campus lobby, HKU]{
            \centering
            \includegraphics[width=0.24\textwidth]{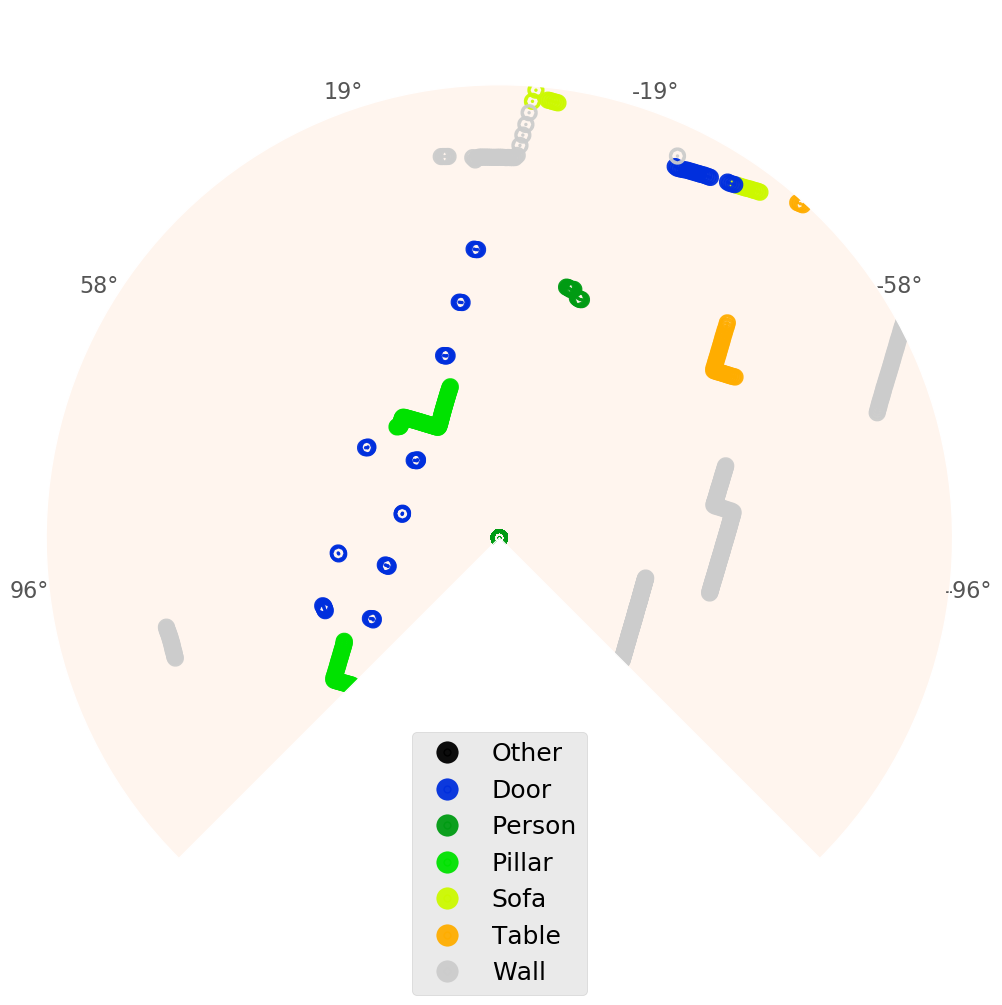}
            \label{fig:centennial_showcase}
    }%
    \caption{
    Semantic label visualization for the Semantic2D dataset, with color-coded class assignments
    }
    \label{fig:semantic2d_labeling_showcase}
\end{figure*}

We found that the pose estimates from \texttt{amcl} lacked sufficient accuracy for our labeling requirements. Using these estimates as initial conditions (see the first box in \figref{fig:semi_automatic_labeling}), we implement the following refinement pipeline:
\begin{enumerate}[noitemsep]
    \item \textbf{Feature Extraction}: We filter out dynamic objects (e.g., people) unsuitable for scan alignment. Inspired by FLIRT features~\citep{tipaldi2010flirt}, we extract stable line features (e.g., walls) from 2D lidar scans instead of using all raw points. 
    These static features, which constitute significant portions of each scan, provide robust reference points for alignment~\citep{pfister2003weighted} (second box in \figref{fig:semi_automatic_labeling}).

    \item \textbf{Scan Alignment}: Using the extracted stable line features and initial \texttt{amcl} pose estimates, we apply the Iterative Closest Point (ICP) algorithm~\citep{thrun2002probabilistic} to refine alignment, leading to substantially improved registration quality (third box in \figref{fig:semi_automatic_labeling}).

    \item \textbf{Semantic Labeling}: The refined alignment enables precise semantic label transfer from map to scan points. 
    Points intersecting labeled objects inherit corresponding semantic labels; points in free space are labeled as ``Person''; all others receive the ``Other'' label.
    Resulting labeled scans for each environment are shown in \figref{fig:semantic2d_labeling_showcase}.

\end{enumerate}

In summary, SALSA reduces the labeling burden from annotating individual lidar points to annotating a single map per scene, achieving substantial time savings while maintaining label quality. 
This framework provides researchers with an efficient pipeline for semantic annotation of 2D lidar data, facilitating advancements in 2D lidar-based scene understanding.

\subsection{Semantic Labeling Application Case}
\label{subsec:semantic_labeling_application_case}
While our Semantic2D dataset was collected using specific robotic platforms (a Jackal robot and a customized robot with three lidar types), researchers may question whether our SALSA labeling framework can be applied to data from other robot models. 
To address this concern, we demonstrate SALSA's applicability on two additional datasets: the synthetic OGM-Turtlebot2 dataset~\citep{xie2023sogmp, xie2025scope, sogmp_dataset} and the real-world MIT Stata Center dataset~\citep{fallon2013stata}.

\begin{figure*}[t]
    \centering
    \subfloat[Dataset collection environment]{
            \centering
            \includegraphics[width=0.38\textwidth]{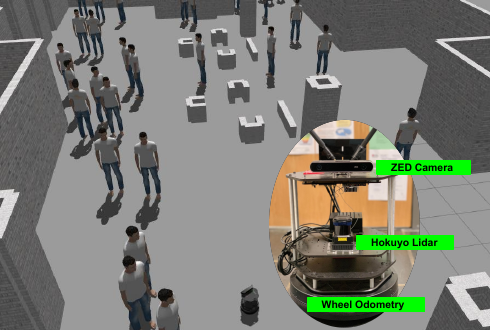}
            \label{fig:omg_turtlebot2_dataset}
    }%
    %\vspace{0.01cm}
    \subfloat[Gazebo Lobby map]{
            \centering
            \includegraphics[width=0.3\textwidth]{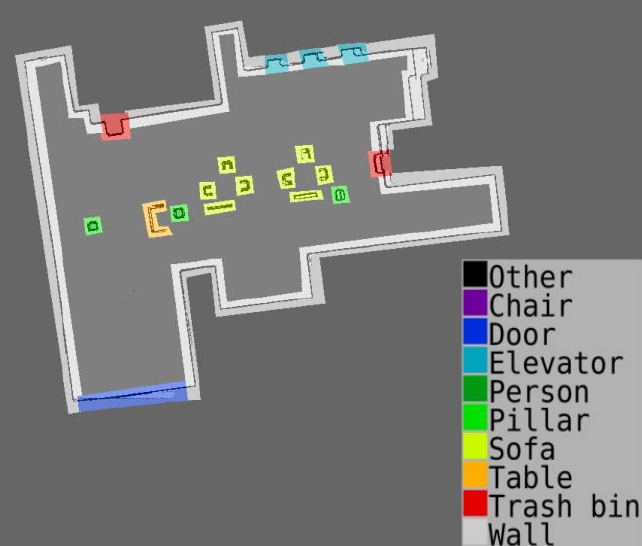}
            \label{fig:gazebo_lobby_map}
    }%
    \subfloat[Semantic labeling example]{
            \centering
            \includegraphics[width=0.3\textwidth]{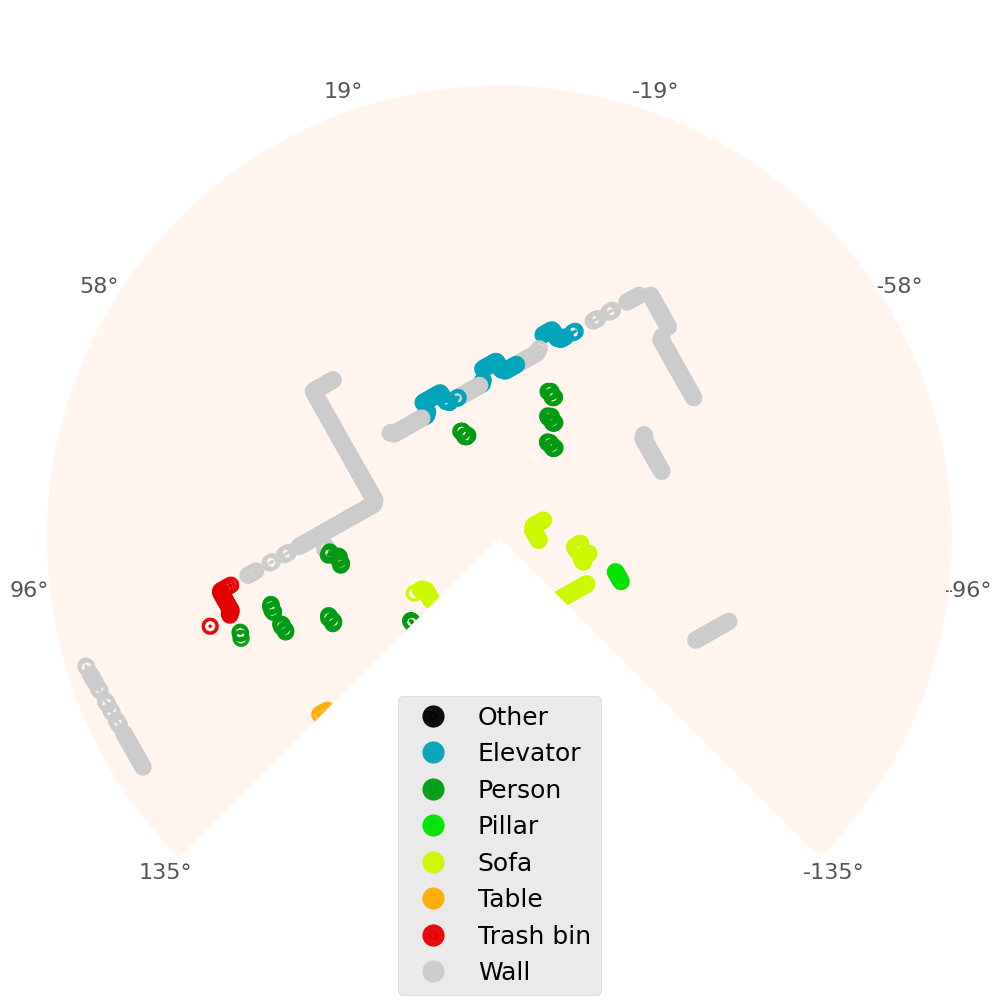}
            \label{fig:gazebo_lobby_showcase}
    }%
    \caption{
    Semantic segmentation results from applying the SALSA labeling framework to the OGM-Turtlebot2 dataset, with color-coded class assignments.
    }
    \label{fig:semantic_labeling_application1}
\end{figure*}

\begin{figure*}[t]
    \centering
    \subfloat[Dataset collection environment]{
            \centering
            \includegraphics[width=0.38\textwidth]{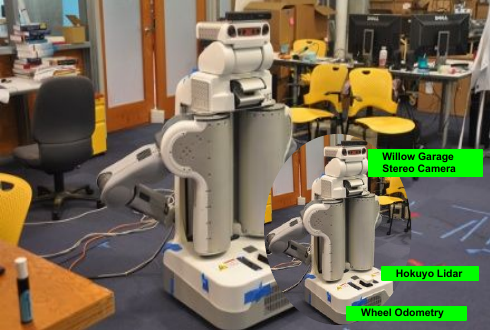}
            \label{fig:mit_stata_center_dataset}
    }%
    %\vspace{0.01cm}
    \subfloat[Gazebo Lobby map]{
            \centering
            \includegraphics[width=0.3\textwidth]{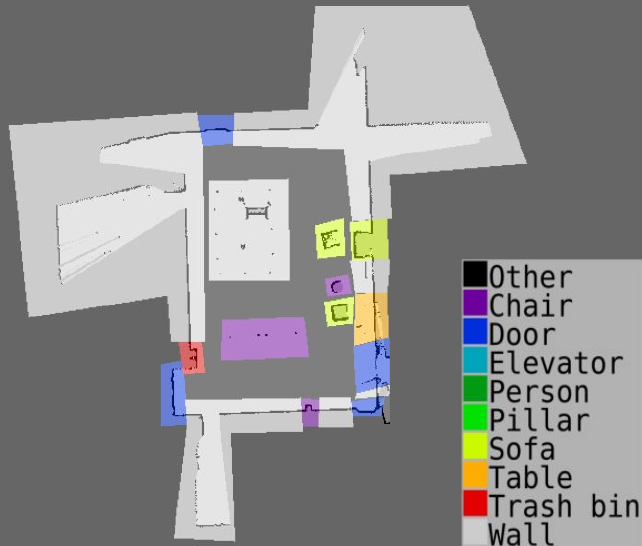}
            \label{fig:mit_stata_center_map}
    }%
    \subfloat[Semantic labeling example]{
            \centering
            \includegraphics[width=0.3\textwidth]{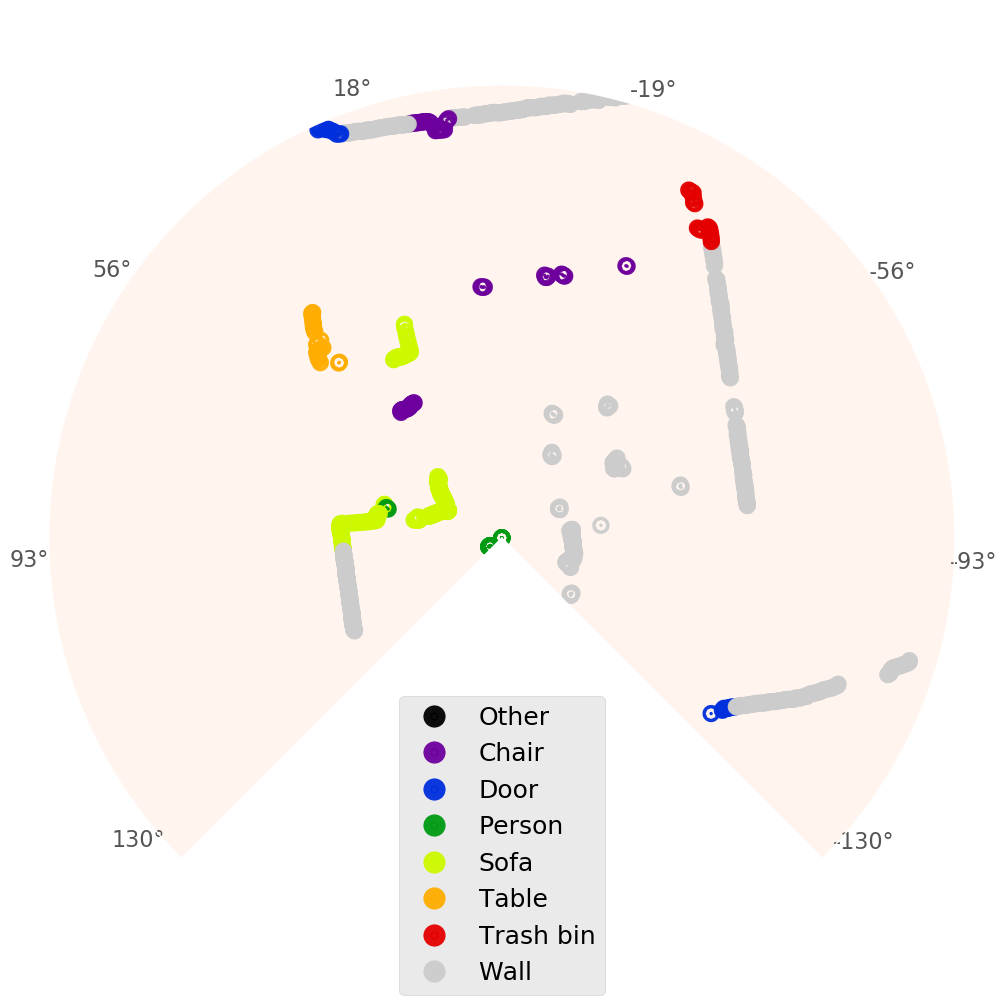}
            \label{fig:mit_stata_center_showcase}
    }%
    \caption{Semantic segmentation results from applying the SALSA labeling framework to the MIT Stata Center dataset, with color-coded class assignments.
    }
    \label{fig:semantic_labeling_application2}
\end{figure*}

The OGM-Turtlebot2 dataset features a simulated Turtlebot2 robot with a 2D lidar navigating an indoor Gazebo lobby environment (\figref{fig:omg_turtlebot2_dataset}) populated by 34 moving pedestrians. 
The robot follows random paths between start and goal points. 
The MIT Stata Center dataset involves a PR2 robot equipped with a 2D Hokuyo lidar\footnote{This lidar has different specifications with our Hokuyo UTM-30LX-EW (\eg, 260\degree \ field of view and 1040 points).} navigating a 10-story academic building (\figref{fig:mit_stata_center_dataset}).

\Cref{fig:gazebo_lobby_map,fig:mit_stata_center_map} show the manually annotated semantic maps for each dataset, while \figrefs{fig:gazebo_lobby_showcase}{fig:mit_stata_center_showcase} present examples of semantically annotated lidar scans generated by our SALSA framework. 
These results demonstrate that SALSA produces accurate and reliable annotations across different robotic platforms, making it readily applicable for researchers working with diverse 2D lidar data.

\subsection{Limitations}
\label{subsec:limitations}
As pioneering contributions, our Semantic2D dataset and SALSA labeling framework have two main limitations.
First, the dataset was collected using only two robot platforms with three types of lidar sensors in campus indoor environments. 
This limited scope reflects our primary objective: to establish a complete 2D lidar semantic segmentation pipeline (from dataset creation and labeling algorithms to segmentation methods and applications) and encourage the research community to expand it. 
As demonstrated in Section~\ref{subsec:semantic_labeling_application_case}, researchers can readily apply our semi-automatic labeling framework to their own data -- such as the OGM-Turtlebot2 dataset~\citep{xie2023sogmp,xie2025scope} and MIT Stata Center dataset~\citep{fallon2013stata} -- and contribute their labeled datasets to our repository.\footnote{Dataset contributions can be submitted via \url{https://github.com/TempleRAIL/semantic2d}}
Through such community efforts, we can collectively enhance the diversity and scale of semantic 2D lidar datasets.

Second, SALSA currently labels all dynamic objects not present in the map as ``Person,'' which may not accurately represent other moving entities like dogs, cats, or bicycles. 
However, since our primary goal is to establish a foundational workflow for 2D lidar semantic segmentation encompassing datasets, labeling frameworks, segmentation algorithms, and applications, we prioritize the overall pipeline's completeness over refining this specific labeling detail. 
Moreover, this limitation can be mitigated by integrating RGB-based object detection. 
For instance, as in our prior work~\citep{xie2021towards}, one can calibrate the lidar and camera sensors, apply detection algorithms like YOLOv3~\citep{redmon2018yolov3} to RGB images, and then project the detected object categories onto corresponding lidar points within the bounding boxes.

%%%%%%%%%%%%%%%%%%%%%%%%%%%%%%%%%%%%%%%%%%%%%%%%%%%%%%%%%%%%%%%%%%%%%%%%%%%%%%%%
\section{Stochastic Semantic Segmentation}
\label{sec:stochastic_semantic_segmentation}
Leveraging the proposed Semantic2D dataset, we design a fine-grained, hardware-friendly stochastic semantic segmentation algorithm for 2D lidar based on a variational autoencoder (VAE) architecture. 
We then demonstrate its superior performance compared to coarse-grained geometry-based algorithms and through ablation studies.

\begin{figure}[t]
    \centering
    \includegraphics[width=0.4\textwidth]{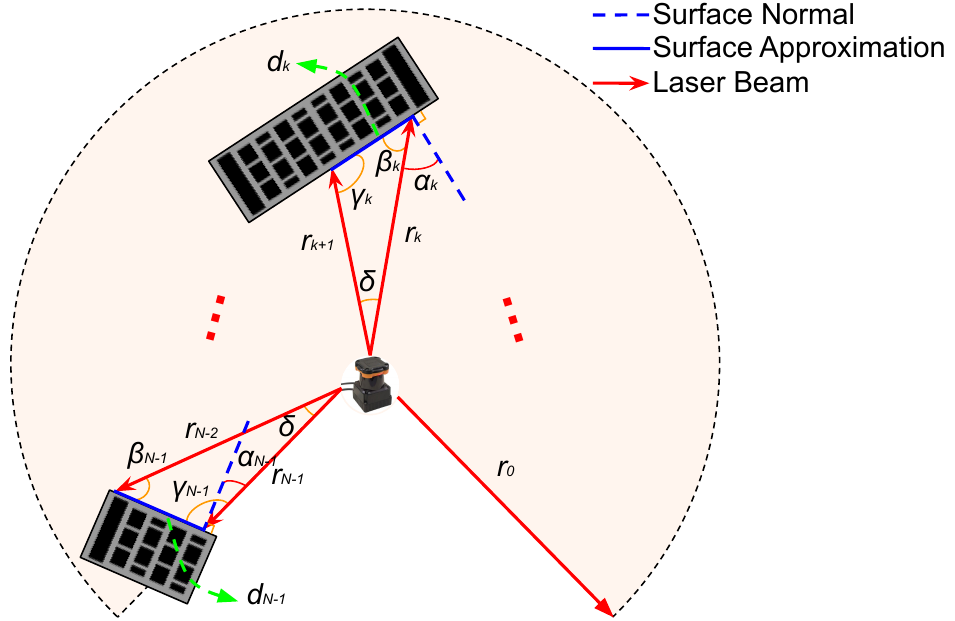}
    \caption{Geometric illustration of how to obtain the angle of incidence $\alpha_k$ for each laser beam.  
    }
    \label{fig:incident_angle}
\end{figure}

\subsection{Problem Formulation}
\label{subsec:problem_formulation}
We consider a mobile robot equipped solely with a 2D lidar sensor, which must perceive and semantically understand its environment to enable autonomous navigation. 
The core challenge is to assign a semantic label to each individual 2D lidar point. Let $\mathbf{Y}_{t}$ and $\mathbf{C}_{t}$ denote the 2D lidar measurement data (i.e., range, bearing, and intensity) and the corresponding semantic category label, respectively, at time t. 
The maximum likelihood 2D semantic segmentation problem is formulated as:
\begin{equation}
    \mathbf{C}_{t}^* = \argmax_{\mathbf{C}_{t}} p_{\theta}(\mathbf{C}_{t} \mid \mathbf{Y}_{t}) \triangleq f_{\theta}(\mathbf{Y}_{t}),
    \label{eq:pf}
\end{equation}
where $\theta$ represents the parameters of the segmentation model $f(\cdot)$. 
At inference time, the objective is to determine the most probable semantic class. 
During training, with ground-truth labels available, the goal is to find the optimal parameters $\theta^*$. 
We implement the segmentation model $f_{\theta}(\cdot)$ using a deep neural network.

\begin{figure*}[t]
    \centering
    \includegraphics[width=0.85\textwidth]{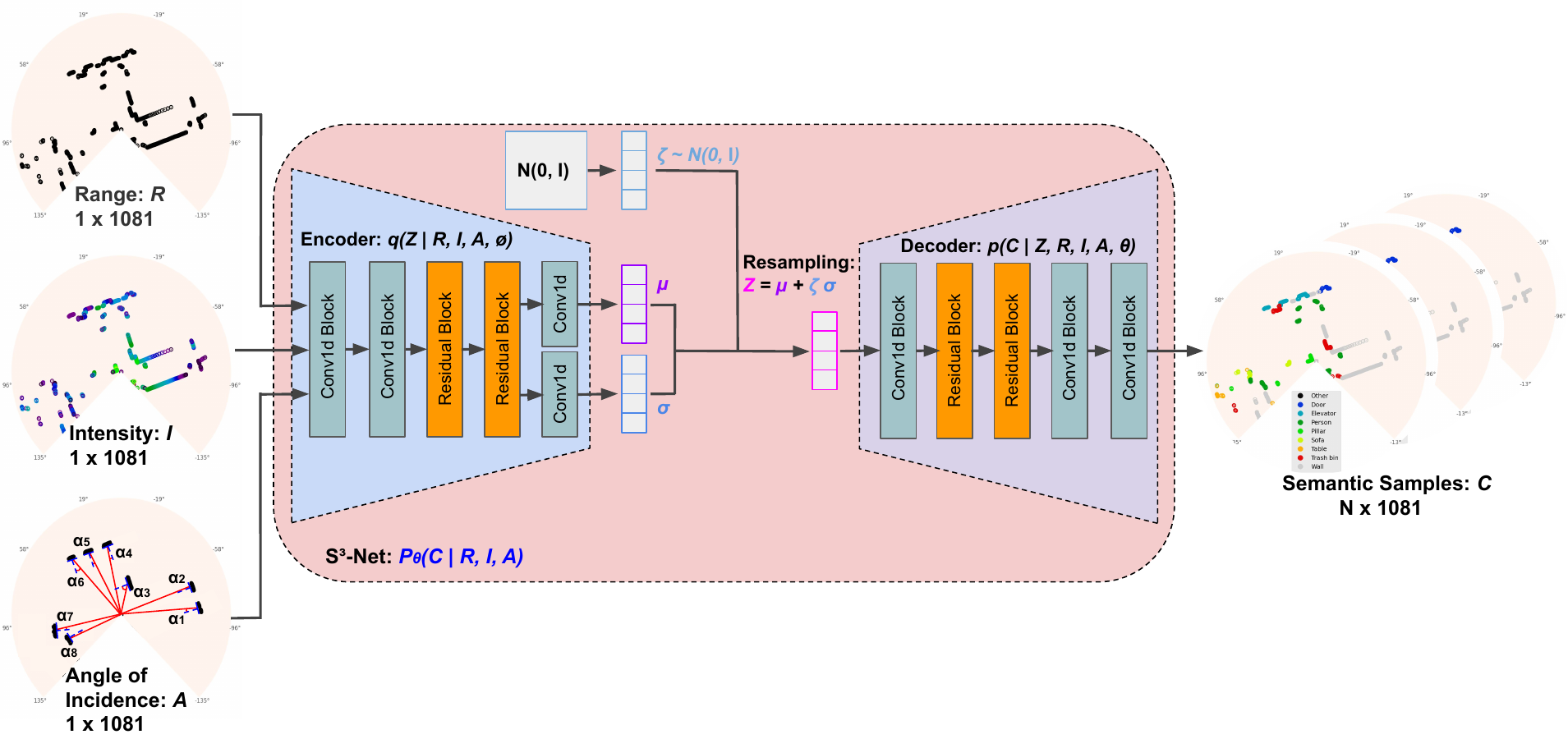}
    \caption{Architecture of S$^3$-Net: a 1D convolutional variational autoencoder that processes 2D lidar range and intensity data to generate semantic labels for each scan point.
            }
    \label{fig:s3_net}
\end{figure*}

\subsection{Input Data}
\label{subsec:input_data}
The design of effective input representations is crucial for deep learning algorithms.
We therefore investigate optimal feature combinations to construct the input representation $\mathbf{Y}_{t}$ for our semantic segmentation model $f_{\theta}(\mathbf{Y}_{t})$.

\subsubsection{Raw Lidar Measurements}
The standard 2D lidar beam model provides raw range and intensity measurements at each time step. 
Let $r_k, b_k, i_k$ denote the range, bearing, and intensity of the $k$-th beam, respectively, for $k = 0, \ldots, N-1$, where $N$ is the total number of beams. 
We aggregate these measurements into vectors $\mathbf{R}_t$ (ranges) and $\mathbf{I}_t$ (intensities).

\subsubsection{Point Cloud}
We also convert the polar coordinate measurements to a 2D point cloud representation, following common practice in 3D lidar semantic segmentation~\citep{yan2024benchmarking}.
The Cartesian coordinates of the $k$-th beam endpoint are given by $\mathbf{p}_k = [r_k \cos b_k, r_k \sin b_k]^\top$, with $\mathbf{P}_{t}$ representing the complete point cloud.

\subsubsection{Angle of Incidence}
\label{subsubsec:angle_of_incidence}
Our prior work on detecting retroreflective markers with lidar~\citep{dames2015experimental} revealed that measured intensity depends on object material properties, range, and angle of incidence. 
For instance, painted drywall exhibits a gradually decreasing intensity with range and incidence angle, while glass and metal show low intensity except near surface normal incidence.

Although material properties are unavailable from lidar, we can estimate the angle of incidence for each beam using local point cloud geometry. 
The primary challenge lies in handling irregular object shapes that complicate surface normal estimation. 
To estimate the incident angle $\alpha_k$ of the $k$-th beam (see \figref{fig:incident_angle}), we approximate the hit surface by the line segment $d_k$ connecting the $k$-th and $(k\!+\!1)$-th beam endpoints. 
This approximation is justified by the fine angular resolution ($\delta$) of 2D lidar sensors (\eg 0.25\degree \ for the Hokuyo UTM-30LX-EW). 
The incident angle is then computed as:
\begin{subequations}
\begin{align}
    d_k = & \ \sqrt{ {r_k}^2 + {r^2_{k+1}} - 2 r_k r_{k+1} \cos(\delta)}, 
    \label{eq:d}  \\
    \beta_k = & \ \arccos\left( \frac{ d^2_k + {r^2_k} - {r^2_{k+1}} }{2 d_k r_k} \right),
    \label{eq:beta}  \\
    \alpha_k = & \ \left\| \frac{\pi}{2} - \beta_k \right\|,
    \label{eq:alpha}
\end{align}
\label{eq:incident_angle}
\end{subequations}
where $\beta_k$ is the beam–surface grazing angle between the $k$-th laser beam and the approximated surface line segment $d_k$. 

For the final $(N\!-\!1)$-th beam, we compute the beam–surface grazing angle $\gamma_{N-1}$ using the $(N\!-\!2)$-th beam and set $\alpha_{N-1} = \pi/2 - \gamma_{N-1}$. We use $\mathbf{A}_t$ denote the vector of all incidence angles.

\subsubsection{Optimal Data Combination}
\label{subsubsec:optimal_data_combination}
We evaluate four input candidates: ranges ($\mathbf{R}$), intensities ($\mathbf{I}$), point clouds ($\mathbf{P}$), and incidence angles ($\mathbf{A}$). 
Ablation studies (Sec.~\ref{subsubsubsubsec:quantitative_results_s3net_data_representation}) across all 15 feature combinations show that the feature combination $\mathbf{Y}_t = \{\mathbf{R}_t, \mathbf{I}_t, \mathbf{A}_t\}$ yields optimal performance for S$^3$-Net. 
Thus, we exclude the 2D point cloud data ($\mathbf{P}$) from the final input configuration.

\subsection{Network Architecture}
\label{subsec:network_architecture}
The optimal lidar measurement combination, $\mathbf{Y}_t = \{\mathbf{R}_t, \mathbf{I}_t, \mathbf{A}_t \}$, serves as input to S$^3$-Net, a deep neural network based on a variational autoencoder (VAE) architecture (\figref{fig:s3_net}). 
We employ a VAE backbone for two key reasons: 1) it aligns with the encoder-decoder structure common in image segmentation networks (e.g., U-Net~\citep{ronneberger2015u}, SegNet~\citep{badrinarayanan2017segnet}, PSPNet~\citep{zhao2017pyramid}), and 2) it provides uncertainty estimates for the output, as demonstrated in prior work~\citep{xie2023sogmp, xie2025scope}.

The input $\mathbf{Y}_t$ is a multi-channel 1D array. 
Since the channels have different units and scales with intensity values being manufacturer-dependent, we apply standardized normalization~\citep{xie2021towards} to enhance generalization. 
The normalized data is processed by 1D convolutional layers, each followed by batch normalization and ReLU activation. 
The VAE backbone generates semantic segmentation samples with per-point uncertainty estimates. Further architectural details are consistent with our previous VAE-based occupancy grid map prediction work~\citep{xie2023sogmp, xie2025scope}.

\subsection{Training Loss}
\label{subsec:training_loss}
While Kullback-Leibler (KL) divergence loss is standard for training VAEs~\citep{kingma2013auto}, we augment it with segmentation-specific losses to enhance performance.
Inspired by findings that combining Cross-Entropy (CE) loss~\citep{zhang2018generalized} (optimizing classification accuracy) and Lovasz-Softmax (LS) loss~\citep{berman2018lovasz} (optimizing mean Intersection-over-Union) improves 3D point cloud segmentation~\citep{yang2021semantic}, we incorporate both into our loss function. 
Additionally, we apply median frequency balancing~\citep{eigen2015predicting} to address class imbalance (\eg prevalent walls versus sparse chairs).

Our final hybrid loss $\mathcal{L}_{\text{seg}}$ combines weighted components:
\begin{equation}
     \mathbf{L_{seg}} = \beta_1 \mathbf{L_{ce}} + \beta_2 \mathbf{L_{ls}} + \beta_3 \mathbf{L_{kl}},
    \label{eq:loss}
\end{equation}
where $\beta_1, \beta_2, \beta_3$ are weighting coefficients. Following parameter settings in~\citep{yang2021semantic, xie2023sogmp, xie2025scope}, we use $[\beta_1, \beta_2, \beta_3] = [1, 1, 0.01]$.

%%%%%%%%%%%%%%%%%%%%%%%%%%%%%%%%%%%%%%%%%%%%%%%%%%%%%%%%%%%%%%%%%%%%%%%%%%%%%%%%
\begin{table*}[t]
    \small\sf\centering
    \caption{Segmentation results (\%, mean $\pm$ std) on the Semantic2D dataset}
    \scalebox{0.6}{
    \begin{tabular}{l|c|c|l|l l l l l l l l l}
        \toprule
        \textbf{Method} & \textbf{FPS} $\uparrow$ & \textbf{Metric} $\uparrow$ & \textbf{All Classes} & \textbf{Chair} & \textbf{Door} & \textbf{Elevator} & \textbf{Person} & \textbf{Pillar} & \textbf{Sofa} & \textbf{Table} & \textbf{Trash bin} & \textbf{Wall} \\ 
        \midrule

         \multirow{2}{*}{\begin{tabular}[c]{@{}c@{}} Line extraction~\citep{pfister2003weighted}  \end{tabular}} &
         \multirow{2}{*}{\begin{tabular}[c]{@{}c@{}} 17.82 \end{tabular}} 
         & CA & 26.56 & - & 26.56 & \textbf{26.56}  & - & - & - & - & - & 26.56 \\
         & & IoU & 25.14 & - & 25.14 & \textbf{25.14}  & - & - & - & - & - & 25.14 \\
         \midrule 

         \multirow{2}{*}{\begin{tabular}[c]{@{}c@{}} Leg detection~\citep{bellotto2008multisensor}  \end{tabular}} &
         \multirow{2}{*}{\begin{tabular}[c]{@{}c@{}} 0.56 \end{tabular}} 
         & CA & 7.57 & - & - & -  & 7.57 & - & - & - & - & -  \\ 
         & & IoU &  2.95 & - & - & -  & 2.95 & - & - & - & - & -  \\ 
         \midrule 
         \midrule

         \multirow{2}{*}{\begin{tabular}[c]{@{}c@{}} S$^3$-Net (\textbf{\textcolor{blue}{R}}) \end{tabular}} &
         \multirow{2}{*}{\begin{tabular}[c]{@{}c@{}} \textbf{305.63} \end{tabular}} 
         & CA & 34.51$\pm$0.85 & 13.54$\pm$0.42 & 57.67$\pm$1.70 & 18.63$\pm$0.69 & 56.95$\pm$1.05 & 24.90$\pm$0.50 & 9.88$\pm$0.39 & 15.63$\pm$0.53 & 52.00$\pm$1.54 & \textbf{61.41$\pm$0.79} \\ 
         & & IoU &  21.13$\pm$0.64 & 7.21$\pm$0.31 & 28.43$\pm$1.08 & 14.16$\pm$0.63  & 27.70$\pm$0.65 & 17.47$\pm$0.71 & 8.48$\pm$0.33 & 8.11$\pm$0.40 & 24.29$\pm$0.98 & 54.34$\pm$0.71 \\ 
         \midrule 

         \multirow{2}{*}{\begin{tabular}[c]{@{}c@{}} S$^3$-Net (I) \end{tabular}} &
         \multirow{2}{*}{\begin{tabular}[c]{@{}c@{}} 273.96 \end{tabular}} 
         & CA & 29.14$\pm$1.01 & 8.74$\pm$0.51 & 47.50$\pm$1.84 & 18.02$\pm$0.73  & 49.63$\pm$1.36 & 21.80$\pm$0.62 & 10.63$\pm$0.48 & 11.68$\pm$0.73 & 47.78$\pm$1.90 & 46.45$\pm$0.96 \\ 
         & & IoU &  17.12$\pm$0.62 & 5.21$\pm$0.34 & 22.44$\pm$0.85 & 12.59$\pm$0.53  & 21.23$\pm$0.62 & 14.68$\pm$0.64 & 8.98$\pm$0.41 & 5.36$\pm$0.37 & 20.08$\pm$0.89 & 43.52$\pm$0.91 \\ 
         \midrule 

         \multirow{2}{*}{\begin{tabular}[c]{@{}c@{}} S$^3$-Net (P) \end{tabular}} &
         \multirow{2}{*}{\begin{tabular}[c]{@{}c@{}} 293.50 \end{tabular}} 
         & CA & 31.09$\pm$0.96 & 11.68$\pm$0.48 & 52.78$\pm$1.95 & 15.74$\pm$0.58  & 56.84$\pm$1.18 & 23.16$\pm$0.56 & 7.08$\pm$0.37 & 14.65$\pm$0.65 & 46.42$\pm$1.76 & 51.43$\pm$1.14 \\  
         & & IoU &  17.73$\pm$0.67 & 6.07$\pm$0.33 & 22.97$\pm$1.07 & 11.18$\pm$0.53  & 24.44$\pm$0.64 & 15.22$\pm$0.71 & 6.03$\pm$0.32 & 7.10$\pm$0.42 & 20.17$\pm$0.97 & 46.38$\pm$1.03 \\  
         \midrule 

         \multirow{2}{*}{\begin{tabular}[c]{@{}c@{}} S$^3$-Net (A) \end{tabular}} &
         \multirow{2}{*}{\begin{tabular}[c]{@{}c@{}} 302.78 \end{tabular}} 
         & CA & 30.79$\pm$1.09 & 12.54$\pm$0.60 & 49.06$\pm$2.12 & 13.56$\pm$0.73  & 55.93$\pm$1.30 & 22.78$\pm$0.78 & 6.98$\pm$0.41 & 13.92$\pm$0.77 & 44.34$\pm$2.04 & 58.02$\pm$1.07 \\ 
         & & IoU &  17.79$\pm$0.75 & 6.43$\pm$0.38 & 21.07$\pm$1.18 & 9.76$\pm$0.65 & 26.77$\pm$0.77 & 13.78$\pm$0.84 & 6.07$\pm$0.38 & 6.34$\pm$0.47 & 18.78$\pm$1.09 & 51.08$\pm$0.95 \\ 
         \midrule 
         \midrule

         \multirow{2}{*}{\begin{tabular}[c]{@{}c@{}} S$^3$-Net (\textbf{\textcolor{blue}{R+I}})\end{tabular}} &
         \multirow{2}{*}{\begin{tabular}[c]{@{}c@{}} 293.71 \end{tabular}} 
         & CA & 37.25$\pm$0.91 & 14.07$\pm$0.63  & 63.76$\pm$1.79 & 25.05$\pm$0.49 & 55.43$\pm$1.02 & 26.44$\pm$0.61  & 16.91$\pm$0.46 & 16.68$\pm$0.77 & 58.74$\pm$1.48 & 58.10$\pm$0.91 \\ 
         & & IoU & 24.24$\pm$0.75 & 7.28$\pm$0.42 & 31.60$\pm$1.06 & 20.07$\pm$0.57 & 30.02$\pm$0.79 & 18.70$\pm$0.81 & \textbf{14.65$\pm$0.45} & 7.89$\pm$0.53 & 33.08$\pm$1.20 & 54.86$\pm$0.88 \\ 
         \midrule 

         \multirow{2}{*}{\begin{tabular}[c]{@{}c@{}} S$^3$-Net (R+P)\end{tabular}} &
         \multirow{2}{*}{\begin{tabular}[c]{@{}c@{}} 266.70 \end{tabular}} 
         & CA & 34.50$\pm$0.93 & 12.99$\pm$0.42  & 59.27$\pm$1.84 & 21.66$\pm$0.70 & 57.64$\pm$1.04 & 25.14$\pm$0.51  & 10.30$\pm$0.43 & 14.94$\pm$0.64 & 53.10$\pm$1.64 & 55.43$\pm$1.15 \\ 
         & & IoU & 20.83$\pm$0.77 & 7.44$\pm$0.34 & 28.18$\pm$1.18 & 15.42$\pm$0.67 & 28.07$\pm$0.66 & 17.06$\pm$0.72 & 9.09$\pm$0.39 & 7.95$\pm$0.50 & 24.00$\pm$1.02 & 50.25$\pm$1.07 \\ 
         \midrule 

         \multirow{2}{*}{\begin{tabular}[c]{@{}c@{}} S$^3$-Net (R+A)\end{tabular}} &
         \multirow{2}{*}{\begin{tabular}[c]{@{}c@{}} 298.81 \end{tabular}} 
         & CA & 35.01$\pm$0.98 & 14.99$\pm$0.60 & 58.44$\pm$1.89 & 18.20$\pm$0.72  & 57.88$\pm$0.96 & 25.38$\pm$0.70 & 11.61$\pm$0.42 & 16.66$\pm$0.78 & 52.93$\pm$1.60 & 59.02$\pm$1.15 \\ 
         & & IoU &  21.07$\pm$0.78 & 7.56$\pm$0.42 & 25.95$\pm$1.22 & 12.71$\pm$0.67 & 31.17$\pm$0.77 & 17.37$\pm$0.87 & 9.86$\pm$0.40 & 7.83$\pm$0.50 & 24.47$\pm$1.11 & 52.68$\pm$1.05 \\ 
         \midrule 

         \multirow{2}{*}{\begin{tabular}[c]{@{}c@{}} S$^3$-Net (I+P)\end{tabular}} &
         \multirow{2}{*}{\begin{tabular}[c]{@{}c@{}} 263.65 \end{tabular}} 
         & CA & 32.72$\pm$0.98 & 11.18$\pm$0.57  & 57.02$\pm$1.98 & 21.83$\pm$0.57 & 54.07$\pm$1.08 & 24.89$\pm$0.53  & 12.42$\pm$0.48 & 12.60$\pm$0.88 & 50.47$\pm$1.75 & 50.02$\pm$0.99 \\ 
         & & IoU & 19.95$\pm$0.71 & 6.27$\pm$0.40 & 25.74$\pm$1.02 & 16.83$\pm$0.56 & 24.02$\pm$0.64 & 16.53$\pm$0.76 & 10.78$\pm$0.43 & 6.39$\pm$0.54 & 25.52$\pm$1.05 & 47.46$\pm$0.95 \\ 
         \midrule 

         \multirow{2}{*}{\begin{tabular}[c]{@{}c@{}} S$^3$-Net (I+A)\end{tabular}} &
         \multirow{2}{*}{\begin{tabular}[c]{@{}c@{}} 282.25 \end{tabular}} 
         & CA & 36.56$\pm$0.94 & 14.80$\pm$0.67 & 60.34$\pm$1.75 & 22.97$\pm$0.70 & 55.37$\pm$1.26 & 27.11$\pm$0.55 & 14.63$\pm$0.54 & 15.23$\pm$0.88 & 60.94$\pm$1.63 & 57.65$\pm$0.95 \\ 
         & & IoU &  23.31$\pm$0.75 & 7.24$\pm$0.43 & 30.12$\pm$1.12 & 19.85$\pm$0.54 & 27.32$\pm$0.71 & 17.54$\pm$0.81 & 13.87$\pm$0.50 & 8.07$\pm$0.59 & 32.29$\pm$1.14 & 53.53$\pm$0.95 \\ 
         \midrule 

         \multirow{2}{*}{\begin{tabular}[c]{@{}c@{}} S$^3$-Net (P+A)\end{tabular}} &
         \multirow{2}{*}{\begin{tabular}[c]{@{}c@{}} 263.77 \end{tabular}} 
         & CA & 32.32$\pm$1.09 & 13.46$\pm$0.59  & 54.11$\pm$2.10 & 15.86$\pm$0.89 & 57.19$\pm$1.11 & 24.18$\pm$0.64  & 8.83$\pm$0.49 & 15.70$\pm$0.87 & 47.81$\pm$1.75 & 53.68$\pm$1.39 \\ 
         & & IoU & 18.85$\pm$0.80 & 6.75$\pm$0.38 & 23.16$\pm$1.20 & 11.05$\pm$0.74 & 28.76$\pm$0.76 & 16.51$\pm$0.83 & 7.46$\pm$0.42 & 6.66$\pm$0.52 & 21.00$\pm$1.07 & 48.30$\pm$1.28 \\ 
         \midrule 
         \midrule

         \multirow{2}{*}{\begin{tabular}[c]{@{}c@{}} S$^3$-Net (R+I+P)\end{tabular}} &
         \multirow{2}{*}{\begin{tabular}[c]{@{}c@{}} 267.77 \end{tabular}} 
         & CA & 36.40$\pm$0.94 & 13.01$\pm$0.58 & 60.77$\pm$1.94 & 25.13$\pm$0.45 & 56.63$\pm$1.10 & 26.85$\pm$0.63 & 16.09$\pm$0.54 & 14.75$\pm$0.81 & 57.66$\pm$1.44 & 56.61$\pm$0.97 \\ 
         & & IoU &  23.41$\pm$0.77 & 7.08$\pm$0.39 & 29.01$\pm$1.05 & 17.14$\pm$0.64 & 28.85$\pm$0.82 & 18.90$\pm$0.85 & 12.24$\pm$0.51 & 6.99$\pm$0.53 & 36.57$\pm$1.19 & 53.91$\pm$0.91 \\ 
         \midrule
         
         \multirow{2}{*}{\begin{tabular}[c]{@{}c@{}} \textbf{\textcolor{blue}{S$^3$-Net (R+I+A)}}\end{tabular}} &
        \multirow{2}{*}{\begin{tabular}[c]{@{}c@{}} 270.32 \end{tabular}} 
         & CA & \textbf{39.44$\pm$0.90} & \textbf{16.19$\pm$0.68} &  \textbf{67.57$\pm$1.72} & 26.45$\pm$0.53 & 57.36$\pm$0.94 & 27.90$\pm$0.48 & 16.37$\pm$0.57 & \textbf{19.51$\pm$0.92} & \textbf{63.33$\pm$1.36} & 60.29$\pm$0.93 \\ 
         & & IoU & \textbf{26.24$\pm$0.79} & 7.81$\pm$0.43 & 33.70$\pm$1.12 & 22.21$\pm$0.60  & \textbf{32.40$\pm$0.83} & \textbf{20.66$\pm$0.86} & 14.02$\pm$0.53 & 8.29$\pm$0.52 & \textbf{40.16$\pm$1.32} & \textbf{56.92$\pm$0.90} \\ 
         \midrule

         \multirow{2}{*}{\begin{tabular}[c]{@{}c@{}} S$^3$-Net (R+P+A)\end{tabular}} &
         \multirow{2}{*}{\begin{tabular}[c]{@{}c@{}} 279.82 \end{tabular}} 
         & CA & 34.78$\pm$1.02 & 14.90$\pm$0.59 & 59.21$\pm$1.99 & 18.21$\pm$0.81 & \textbf{58.65$\pm$1.04} & 25.19$\pm$0.66 & 11.94$\pm$0.44 & 16.69$\pm$0.84 & 53.51$\pm$1.58 & 54.81$\pm$1.19 \\ 
         & & IoU &  20.91$\pm$0.78 & 7.64$\pm$0.41 & 26.04$\pm$1.12 & 13.38$\pm$0.74 & 30.56$\pm$0.75 & 18.00$\pm$0.85 & 10.19$\pm$0.41 & 7.94$\pm$0.52 & 24.67$\pm$1.08 & 49.76$\pm$1.12 \\ 
         \midrule

         \multirow{2}{*}{\begin{tabular}[c]{@{}c@{}} S$^3$-Net (I+P+A)\end{tabular}} &
         \multirow{2}{*}{\begin{tabular}[c]{@{}c@{}} 268.24 \end{tabular}} 
         & CA & 37.20$\pm$0.97 & 14.19$\pm$0.69 & 62.20$\pm$1.94 & 23.70$\pm$0.66 & 56.79$\pm$1.14 & 27.71$\pm$0.54 & 15.13$\pm$0.49 & 16.92$\pm$0.96 & 61.32$\pm$1.26 & 57.04$\pm$1.03 \\ 
         & & IoU &  23.98$\pm$0.80 & 7.35$\pm$0.43 & 29.56$\pm$1.08 & 19.16$\pm$0.67 & 27.95$\pm$0.77 & 19.66$\pm$0.90 & 12.92$\pm$0.46 & 7.97$\pm$0.59 & 37.44$\pm$1.27 & 53.77$\pm$1.00 \\ 
         \midrule
         \midrule

         \multirow{2}{*}{\begin{tabular}[c]{@{}c@{}} S$^3$-Net {(R+I+P+A)}\end{tabular}} &
        \multirow{2}{*}{\begin{tabular}[c]{@{}c@{}} 253.90 \end{tabular}} 
         & CA & 38.86$\pm$0.89 & 14.93$\pm$0.61 & 65.67$\pm$1.82 & 26.55$\pm$0.41 & 56.95$\pm$1.06 & \textbf{28.89$\pm$0.52} & \textbf{16.99$\pm$0.57} & 18.64$\pm$0.85 & 59.75$\pm$1.32 & 61.40$\pm$0.86 \\ 
         & & IoU & 25.65$\pm$0.75 & \textbf{8.06$\pm$0.43} & \textbf{34.08$\pm$1.11} & 21.60$\pm$0.50 & 28.62$\pm$0.76 & 19.82$\pm$0.75 & 14.40$\pm$0.52 & \textbf{9.40$\pm$0.58} & 36.71$\pm$1.24 & 58.19$\pm$0.85 \\ 
         \midrule 
         \bottomrule
    \end{tabular}
    }
    \label{tab:segmentation}
\end{table*}

\subsection{Segmentation Results}
\label{subsec:segmentation_results}

\subsubsection{Baselines}
\label{subsubsec:baselines}
While no state-of-the-art general segmentation algorithms exist for 2D lidar, we compare our proposed S$^3$-Net against two geometry-based approaches: line detection~\citep{pfister2003weighted}\footnote{For ground truth, we combine predominant linear features (\ie walls, doors, and elevators) since the detector cannot distinguish between individual classes.} and leg detection~\citep{bellotto2008multisensor}\footnote{For ground truth, we use all person leg points.}. 
To evaluate input data selection, we also include 14 ablation baselines using the same S$^3$-Net architecture with different feature combinations, denoted as S$^3$-Net (\texttt{data-combination}), where \texttt{data-combination} is a subset of $\{\mathbf{R}, \mathbf{I}, \mathbf{P},\mathbf{A} \}$.
For example, S$^3$-Net ($\mathbf{R} + \mathbf{I} + \mathbf{A}$) represents our proposed optimal combination (range, intensity, and incident angle).

All deep neural networks are trained on the Temple Engineering training subsets (70\% of Semantic2D data) and evaluated on corresponding testing subsets (20\%).

\subsubsection{Evaluation Metrics}
\label{subsubsec:evaluation_metrics}
We evaluate semantic segmentation performance using two standard metrics from~\citet{hackel2017semantic3d}:
\begin{itemize}
    \item \textbf{Class Accuracy (CA)}:
        \begin{equation}
             CA_c = \frac{TP_c + TN_c}{TP_c + FP_c + FN_c + TN_c},
            \label{eq:ca}
        \end{equation}
        
    \item \textbf{Intersection over Union (IoU)}:
    \begin{equation}
         IoU_c = \frac{TP_c}{TP_c + FP_c + FN_c},
        \label{eq:iou}
    \end{equation}
\end{itemize}
where $TP_c$, $FP_c$, $FN_c$, and $TN_c$ represent true positives, false positives, false negatives, and true negatives for class $c$, respectively. 
These values correspond to counts of 2D lidar points assigned to class $c$. 
We report both per-class metrics (CA and IoU) and their means across all categories (mCA and mIoU) as percentages, with higher values indicating better performance.

\subsubsection{Quantitative Results of Data Representation}
\label{subsubsubsubsec:quantitative_results_s3net_data_representation}
\tabref{tab:segmentation} presents quantitative segmentation results for our proposed S$^3$-Net ($\mathbf{R}+ \mathbf{I}+\mathbf{A}$) and ablation baselines, revealing six key findings.

First, compared to geometry-based methods like line extraction for static linear objects~\citep{pfister2003weighted} and leg detection for moving pedestrians~\citep{bellotto2008multisensor}, our S$^3$-Net not only enables segmentation of diverse categories (e.g., chairs, tables, sofas) but also achieves higher accuracy (CA and IoU) for both static and dynamic objects.
The sole exception is elevator segmentation, where the line detector shows marginally better performance, though it cannot distinguish elevators from doors or walls.

Second, when using only a single data type, S$^3$-Net ($\mathbf{R}$) with range data yields more accurate segmentation per class than models using point position, intensity, or incident angle.
This establishes lidar range as the most critical feature for 2D semantic segmentation. 
Furthermore, representing this data in polar coordinates (range) proves more amenable to neural network processing than Cartesian point data, as evidenced by superior performance across all metrics.
This supports our design choice for S$^3$-Net and aligns with findings from PolarNet~\citep{zhang2020polarnet}, which also demonstrates the advantage of polar representations.

Third, among two-data combinations, S$^3$-Net ($\mathbf{R}+\mathbf{I}$) achieves superior segmentation (\ie higher mCA, mIoU, CA, and IoU) across nearly all categories, demonstrating that intensity is the second most important feature for 2D lidar segmentation. 
This improvement stems from material-dependent reflectance properties, where intensity variations provide discriminative cues for materials like drywall, wooden doors, and bare metal~\citep{dames2015experimental}. 
Conversely, combining point position with range (S$^3$-Net ($\mathbf{P}+\mathbf{R}$)) yields no improvement over range alone, as both represent the same geometric information in different coordinate systems.

Fourth, when three or more data types are available, our proposed S$^3$-Net ($\mathbf{R}+\mathbf{I}+\mathbf{A}$) achieves the best segmentation performance across nearly all categories, outperforming other three- or four-data combinations.\footnote{This aligns with our earlier finding that adding point data ($\mathbf{P}$) is redundant, as it provides no performance gain over range-based representations.} 
This result highlights the importance of incident angle as the third most critical feature for 2D lidar segmentation.
The improvement arises because lidar intensity measurements can be affected by range, material properties, and incidence angle.
By incorporating incident angle data, our method implicitly corrects intensity-related errors, thereby enhancing segmentation accuracy. 
This is consistent with recent work~\citep{viswanath2023off}, where explicit intensity correction using incidence information improved 3D lidar semantic segmentation.

Fifth, our VAE architecture demonstrates strong output consistency despite its stochastic nature.
Since the VAE's output depends on sampled noise $\zeta$ (see \figref{fig:s3_net}), we evaluated label consistency by generating 32 outputs per input. 
The results in \tabref{tab:segmentation} show mean CA and IoU values with standard deviations predominantly below 2\%, confirming the model's reliability across stochastic samples.

Finally, computational efficiency tests on a resource-constrained Intel i5-8250U CPU (1.60GHz) show that S$^3$-Net achieves inference speeds up to 300 FPS, significantly surpassing geometry-based methods (line extraction~\citep{pfister2003weighted} and leg detection~\citep{bellotto2008multisensor}, which max at 18 FPS). 
This demonstrates real-time capability for semantic segmentation on mobile robots, effectively augmenting standard 2D lidar with semantic perception.

\begin{figure}[t]
    \centering
    \includegraphics[width=0.45\textwidth]{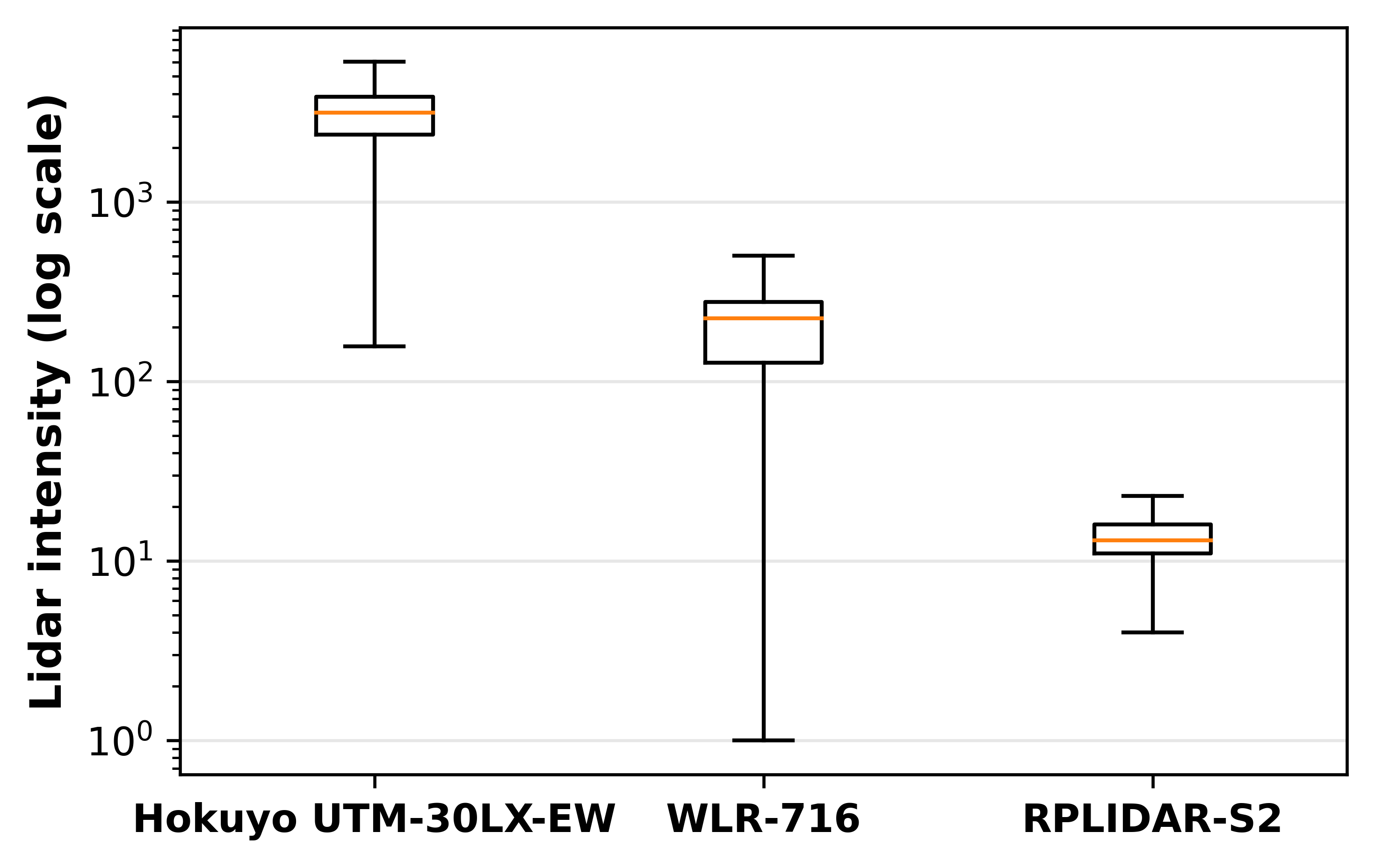}
    \caption{
             Intensity value statistics for three 2D lidar sensors (Hokuyo UTM-30LX-EW, WLR-716, and RPLIDAR-S2), showing orders of magnitude variation in average intensity readings, which poses a key challenge for generalizing semantic segmentation models across different lidar hardware.
            }
    \label{fig:lidar_intensity_plot}
\end{figure}

\begin{figure*}[t]
    \centering
    \subfloat[CA]{
        %\begin{minipage}{0.25\linewidth}
            \centering
            \includegraphics[width=0.4\textwidth]{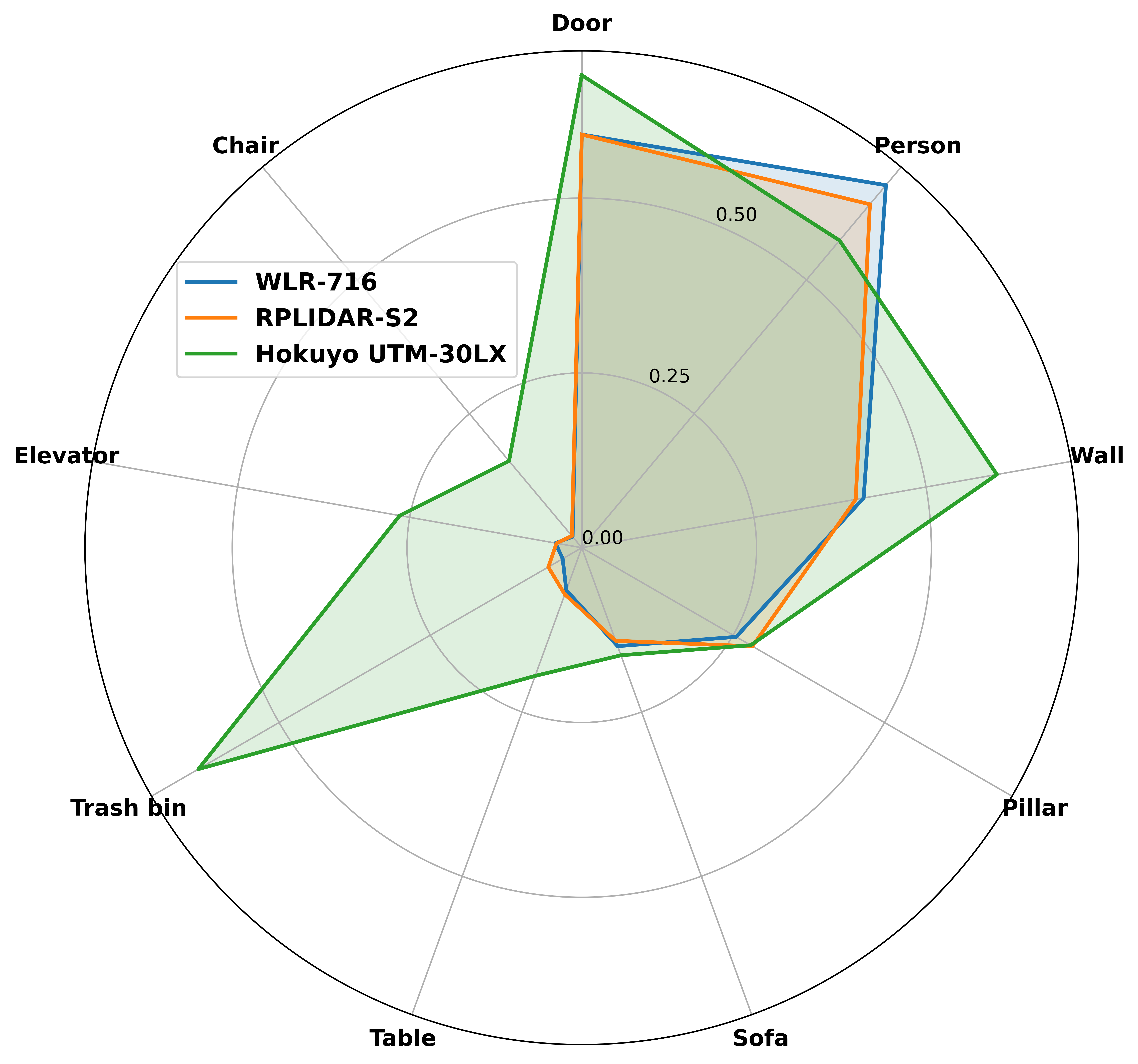}
            \label{fig:ca_wlr_rplidar}
            %\caption{World Map}
        %\end{minipage}%
    }%
    \hspace{0.001cm}
    \subfloat[IoU]{
        %\begin{minipage}{0.25\linewidth}
            \centering
            \includegraphics[width=0.4\textwidth]{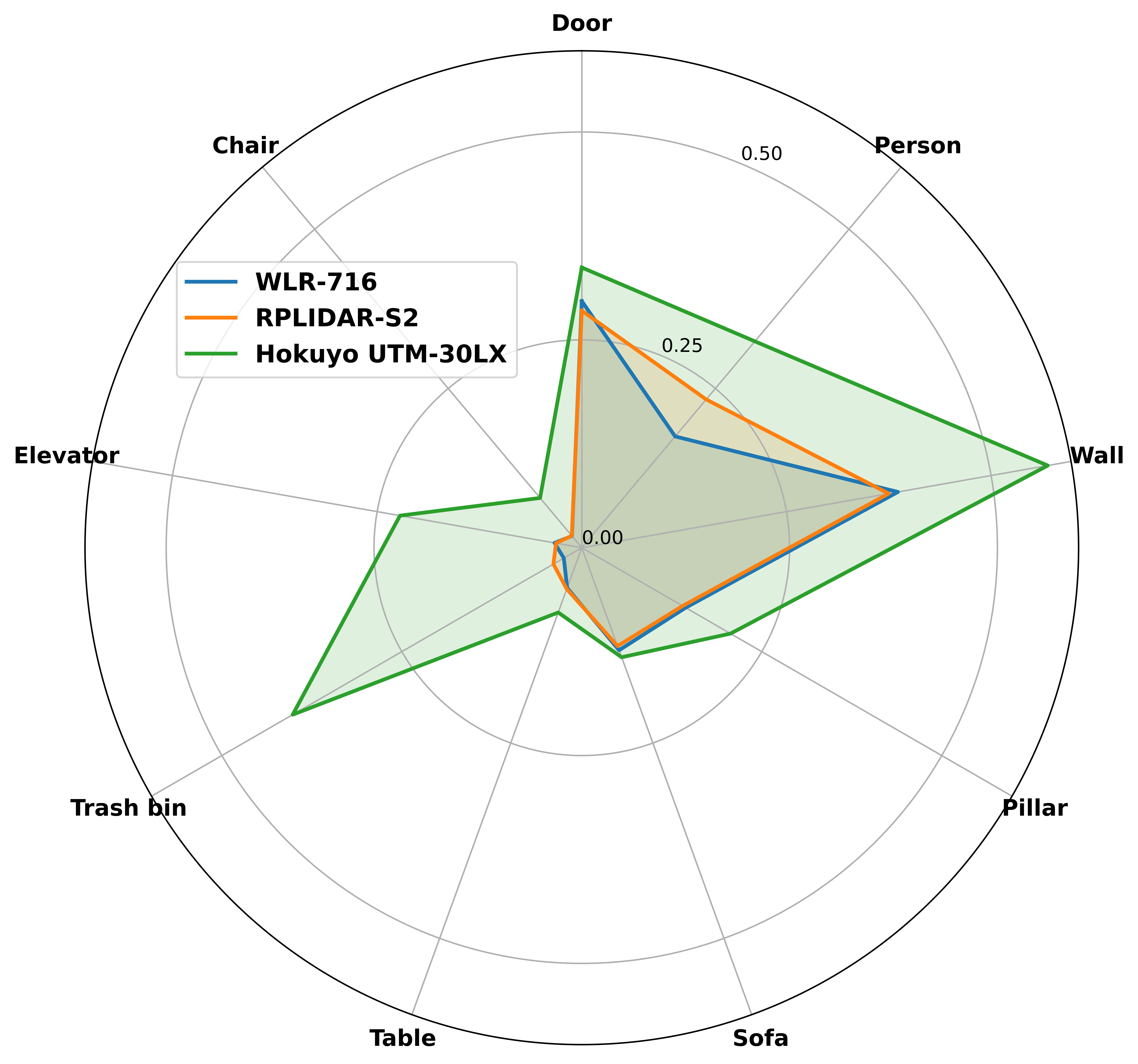}
            \label{fig:iou_wlr_rplidar}
            %\caption{World Map}
        %\end{minipage}%
    }%
    \caption{
    Semantic segmentation generalization to WLR-716 and RPLIDAR-S2 lidar sensors in HKU environments without retraining, showing consistent performance profiles despite sensor differences. 
    Hokuyo UTM-30LX results from Temple Engineering environments are included for reference.
    }
    \label{fig:ca_iou_wlr_rplidar}
\end{figure*}

\begin{figure*}[t]
    \centering
    \subfloat[CA]{
        %\begin{minipage}{0.25\linewidth}
            \centering
            \includegraphics[width=0.4\textwidth]{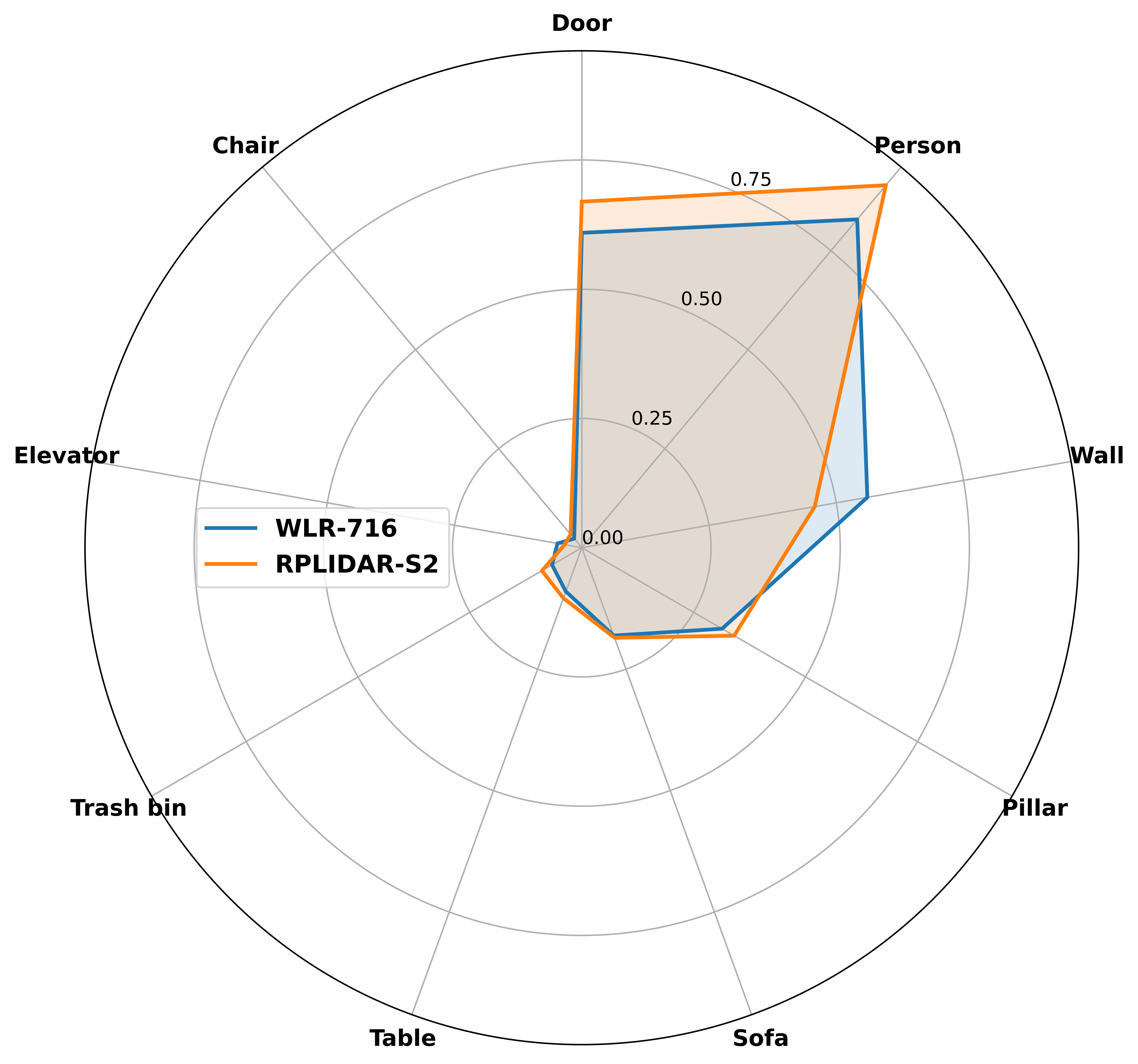}
            \label{fig:ca_wlr_rplidar_retraining}
            %\caption{World Map}
        %\end{minipage}%
    }%
    \hspace{0.001cm}
    \subfloat[IoU]{
        %\begin{minipage}{0.25\linewidth}
            \centering
            \includegraphics[width=0.4\textwidth]{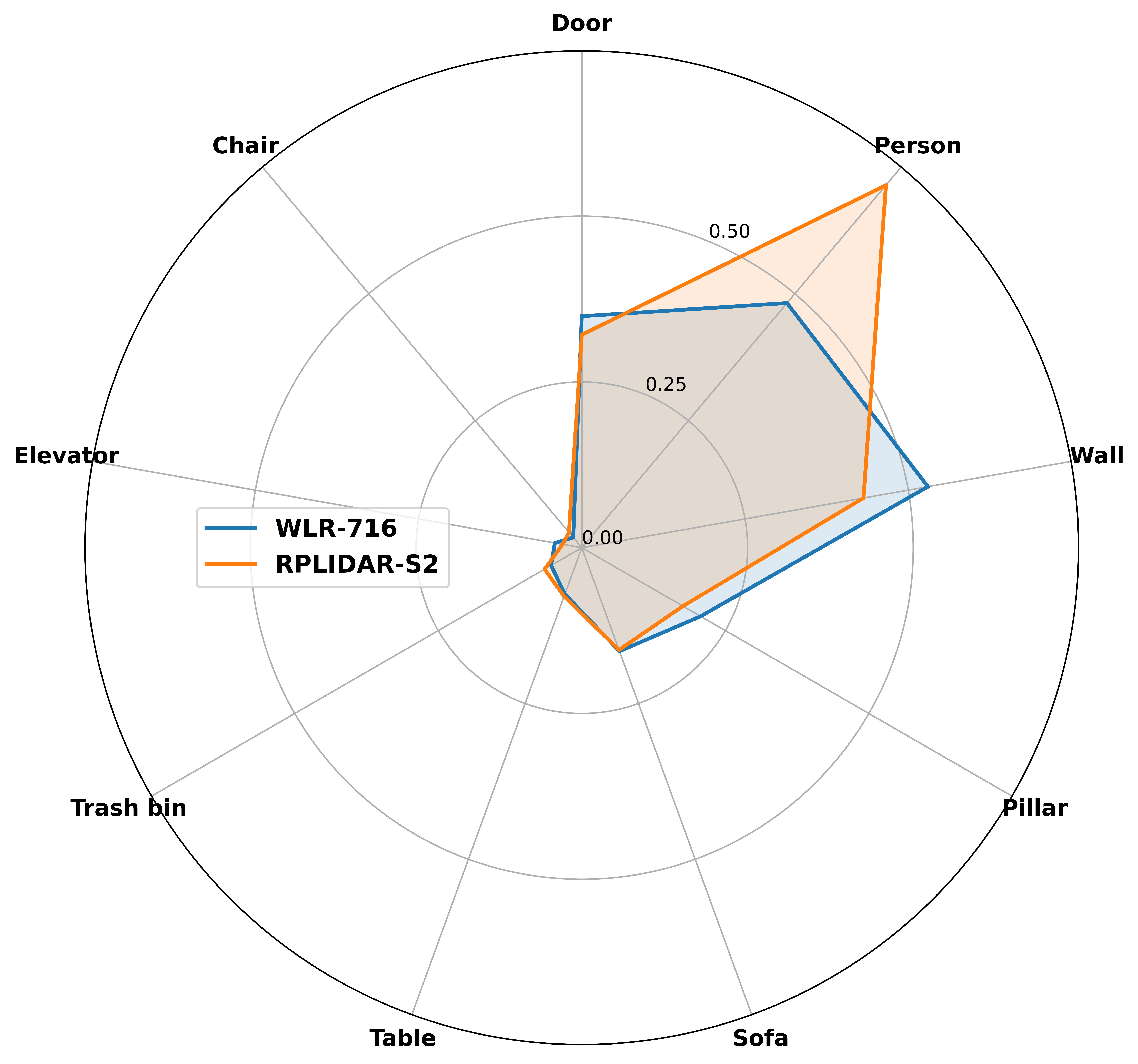}
            \label{fig:iou_wlr_rplidar_retraining}
            %\caption{World Map}
        %\end{minipage}%
    }%
    \caption{
    Semantic segmentation results for WLR-716 and RPLIDAR-S2 lidar sensors in HKU environments after retraining, demonstrating similar CA and IoU performance profiles across object categories.
    }
    \label{fig:ca_iou_wlr_rplidar_retraining}
\end{figure*}

\begin{figure*}[t]
    \centering
    \includegraphics[width=0.95\textwidth]{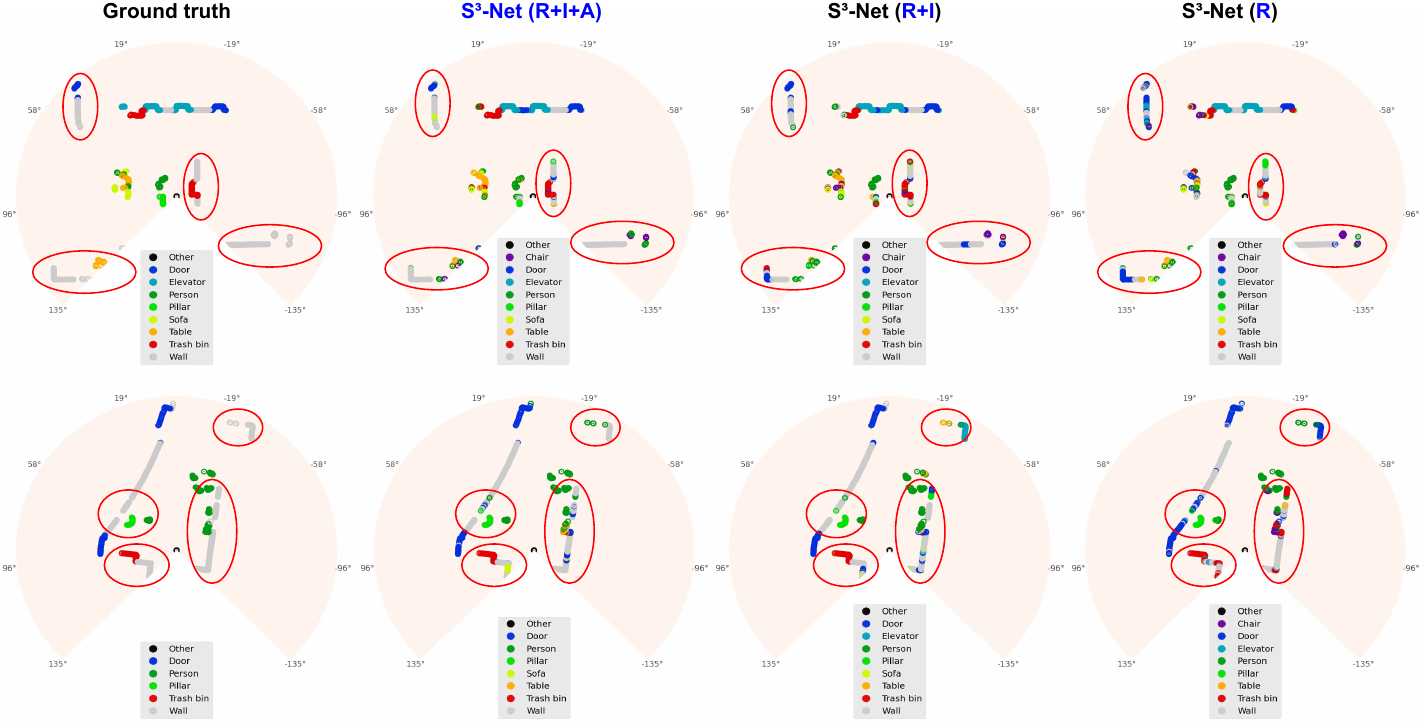}
    \caption{
    Stochastic semantic segmentation results from multiple variations of S$^3$-Net on the Semantic2D dataset, showing color-coded class labels with red ellipses highlighting significant segmentation errors.
            }
    \label{fig:semantic2d_lidar_showcase}
\end{figure*}

\subsubsection{Quantitative Results of Lidar Types}
\label{subsubsubsubsec:quantitative_results_s3net_lidar_types}
Although our S$^3$-Net ($\mathbf{R}+\mathbf{I}+\mathbf{A}$) achieves strong segmentation performance with Hokuyo UTM-30LX-EW lidar, we aim to generalize its use across different 2D lidar sensors, such as WLR-716 and RPLIDAR-S2, without full retraining. 
A primary challenge is configuration variation, especially in intensity values, which differ considerably across brands and models, as illustrated in \figref{fig:lidar_intensity_plot}.

\paragraph{Direct Model Transfer}
We fine-tuned the S$^3$-Net model trained on Hokuyo UTM-30LX-EW lidar data and evaluated it on datasets collected with WLR-716 and RPLIDAR-S2 sensors in three HKU buildings.
Since these lidars differ in point count and field of view, we first reprojected their range/angle data to match the Hokuyo configuration through coordinate transformation and point interpolation. 
We applied sensor-specific standardized normalization to account for differing value ranges in distance, intensity, and incident angle measurements. 
The model was then fine-tuned using only a small validation subset (10\% of data) from the HKU Building dataset for WLR-716 and RPLIDAR-S2, respectively, before final evaluation on their respective test subsets (20\% of data).

\Cref{fig:ca_iou_wlr_rplidar} compares the CA and IoU performance of the fine-tuned WLR-716 and RPLIDAR-S2 models against the original Hokuyo Temple Engineering results. 
Both sensors achieve nearly identical semantic segmentation performance on the HKU dataset, with the exception of slightly lower IoU for WLR-716. 
We attribute this difference to WLR-716's sparser point cloud (811 points versus 1,792 points in RPLIDAR-S2) and lower angular resolution, which may introduce interpolation artifacts.
The comparable performance demonstrates that S$^3$-Net ($\mathbf{R}+\mathbf{I}+\mathbf{A}$) can be effectively adapted to different lidar sensors while maintaining segmentation quality.

Notably, both adapted models show improved CA and IoU for doors, walls, and people compared to chairs, elevators, trash cans, and tables. 
This performance pattern aligns with environmental differences between Temple and HKU buildings: HKU environments contain fewer instances of the latter categories, as visible in \figrefs{fig:cyc}{fig:centennial}. 
These results validate S$^3$-Net's cross-platform capability and its reasonable adaptation to varying environmental conditions across different lidar sensors.

\paragraph{Retrained Models}
Given the promising generalization of S$^3$-Net, initially trained on Hokuyo UTM-30LX-EW lidar, to WLR-716 and RPLIDAR-S2 sensors via fine-tuning, we further investigate the impact of varying lidar configurations (e.g., range, horizontal FOV, angular resolution) on segmentation performance.
To this end, we retrain S$^3$-Net separately for each sensor using their respective training subsets (70\% of the HKU dataset) and evaluate on corresponding test subsets (20\%). 
The HKU dataset was collected in identical building environments using both sensors mounted on the same robot (see \figref{fig:semantic2d_overview}), thereby minimizing environmental and platform-related confounding factors.

\Cref{fig:ca_iou_wlr_rplidar_retraining} presents the CA and IoU performance of the retrained models, revealing three key observations. 
First, the retrained models exhibit similar performance profiles to the fine-tuned versions (see \figref{fig:ca_iou_wlr_rplidar}) but achieve higher accuracy across all categories—most notably for ``person'' segmentation.
This improvement stems from the use of native sensor data during retraining, which eliminates artifacts introduced by coordinate reprojection and point interpolation during fine-tuning. 
This suggests that small objects (e.g., human legs) are more sensitive to such spatial transformations than large objects (e.g., walls, sofas).
Second, similar to the fine-tuning results, the retrained models for both sensors achieve comparable overall performance, though RPLIDAR-S2 yields significantly higher accuracy for ``person'' segmentation while slightly underperforming on ``wall'' segmentation. 
This indicates that higher angular resolution and point density (as with RPLIDAR-S2) particularly benefit small-object segmentation.

In summary, lidar configuration, especially angular resolution and point density, has a greater impact on small-object segmentation performance.
Higher-resolution sensors provide distinct advantages for semantic segmentation of fine structures, such as human legs, while large objects remain robust to sensor variations.

\subsubsection{Qualitative Results}
\label{subsubsec:qualitative_results}
\Cref{fig:semantic2d_lidar_showcase} and the accompanying multimedia material present semantic segmentation results from our proposed S$^3$-Net ($\mathbf{R}+\mathbf{I}+\mathbf{A}$) and two strong ablation baselines: S$^3$-Net ($\mathbf{R}$) and S$^3$-Net ($\mathbf{R}+\mathbf{I}$).
The results demonstrate that S$^3$-Net ($\mathbf{R}+\mathbf{I}+\mathbf{A}$), which incorporates range, intensity, and incident angle data, yields more accurate segmentation than the baselines.
This is evidenced by fewer mis-segmented points within the highlighted regions (red ellipses). 
These qualitative findings confirm the ability of 2D lidar to achieve semantic scene understanding without a camera, enabling enhanced performance in various lidar-based mobile robotics applications.

%%%%%%%%%%%%%%%%%%%%%%%%%%%%%%%%%%%%%%%%%%%%%%%%%%%%%%%%%%%%%%%%%%%%%%%%%%%%%%%%
\section{Semantic2D Applications}
\label{sec:segmentation_applications}
Our S$^3$-Net enables 2D lidar sensors to provide semantic information, allowing mobile robots to achieve high-level scene understanding without cameras, as illustrated in \figref{fig:semantic2d_overview}.
This capability creates opportunities to enhance existing 2D lidar-based robotic applications, with a non-comprehensive list below: 
\begin{itemize}
    \item \textbf{Object Tracking}: 
    Our framework supports identification, tracking, and semantic labeling of both static objects (tables, sofas, trash bins) and dynamic pedestrians by combining lidar geometry with semantic labels from S$^3$-Net. 
    The Semantic2D dataset provides all necessary sensor data, ground truth maps, and pedestrian annotations.

    \item \textbf{Mapping}: 
    Semantic maps can be constructed using lidar geometric data and S$^3$-Net semantic labels. The dataset includes ground truth semantic maps and tools for creating custom mappings.

    \item \textbf{Localization}: 
    Semantic localization algorithms can be developed using the dataset's comprehensive measurements (lidar data, odometry, semantic maps) and ground truth robot poses.

    \item \textbf{Navigation}: 
    The framework enables semantic-aware navigation control, including integration with natural language interfaces~\citep{srivastava2024speechguided}. 
    The dataset provides complete perception data (lidar, RGB images, pedestrian tracks, poses, paths) and control data (velocity commands, trajectories).
\end{itemize}

While these applications demonstrate the broad utility of our work, we focus specifically on semantic mapping and navigation to validate the effectiveness of the Semantic2D dataset and S$^3$-Net segmentation.

\begin{table*}[t]
    \small\sf\centering
    \caption{Semantic mapping results (\%) in Temple Engineering Environments}
    \scalebox{0.75}{
    \begin{tabular}{l|c|c|l|l l l l l l l l l}
        \toprule
        \textbf{Environment} & \textbf{SSIM} $\uparrow$ & \textbf{Metric} $\uparrow$ & \textbf{All Classes} & \textbf{Chair} & \textbf{Door} & \textbf{Elevator} & \textbf{Person} & \textbf{Pillar} & \textbf{Sofa} & \textbf{Table} & \textbf{Trash bin} & \textbf{Wall} \\ 
        \midrule

         \multirow{2}{*}{\begin{tabular}[c]{@{}c@{}} Engineering lobby  \end{tabular}} &
         \multirow{2}{*}{\begin{tabular}[c]{@{}c@{}} 87.85 \end{tabular}} 
         & CA & 74.16 & 65.28 & 95.34 & 93.83  & - & 98.19 & 75.72 & 92.67 & 98.54 & 47.86 \\
         & & IoU & 47.23 & 4.22 & 53.60 & 91.08  & - & 84.11 & 70.52 & 22.79 & 53.18 & 45.60 \\
         \midrule 

         \multirow{2}{*}{\begin{tabular}[c]{@{}c@{}} Engineering 8th-floor \end{tabular}} &
         \multirow{2}{*}{\begin{tabular}[c]{@{}c@{}} 84.16 \end{tabular}} 
         & CA & 64.49 & 100.00 & 88.67 & 98.69  & - & - & 69.60 & 87.83 & 89.14 & 46.49  \\ 
         & & IoU & 30.73 & 9.35 & 42.03 & 58.69  & - & - & 50.75 & 20.32 & 50.77 & 44.65  \\ 

     \bottomrule
    \end{tabular}
    }
    \label{tab:semantic_mapping}
\end{table*}

\begin{table*}[t]
    \small\sf\centering
    \caption{Category confusion for semantic mapping mismatches in Temple Engineering environments}
    \scalebox{0.85}{
    \begin{tabular}{l|l|c|l}
        \toprule
        \textbf{Environment} & \textbf{Confused class pair} & \textbf{\% of confused cells} & \textbf{Typical relation} \\
        \midrule

        \multirow{6}{*}{\begin{tabular}[c]{@{}c@{}} Engineering lobby \end{tabular}} 
          & door-wall      & \textbf{32.2} & Adjacent vertical structures \\
          & door-trash bin & 12.8 & Both near walls / entrances \\
          & chair-table    & 11.5 & Similar furniture category \\
          & sofa-wall      &  8.7 & Similar line structures \\
          & table-wall     &  6.9 & Similar line structures \\
         
        \midrule
        \multirow{6}{*}{\begin{tabular}[c]{@{}c@{}} Engineering 8th-floor \end{tabular}} 
          & door-wall      & \textbf{28.3} & Adjacent vertical structures \\
          & door-trash bin & 20.1 & Both near walls / entrances \\
          & sofa-wall      & 11.7 & Similar line structures \\
          & chair-table    &  5.5 & Similar furniture category \\
          & door-elevator  &  4.5 & Similar openings/doorways \\
          
        \bottomrule
    \end{tabular}
    }
    \label{tab:category_confusion}
\end{table*}

\subsection{Semantic Mapping}
\label{subsec:semantic_mapping}
For semantic mapping, we assume known robot poses and employ a modified inverse sensor model~\citep{thrun2003learning} to generate semantic occupancy grid maps. 
Our approach incorporates two key modifications: 1) dynamic objects (e.g., pedestrians) are filtered from lidar points prior to mapping using semantic labels, and 2) semantic labels are assigned to occupied grid cells via majority voting. 
We evaluate this algorithm in the Temple Engineering lobby and $8$th-floor environments using Semantic2D dataset, where S$^3$-Net provides per-point semantic category information during the mapping process.

\begin{figure*}[t]
    \centering
    \subfloat[Engineering lobby, Temple]{
        %\begin{minipage}{0.25\linewidth}
            \centering
            \includegraphics[width=0.4\textwidth]{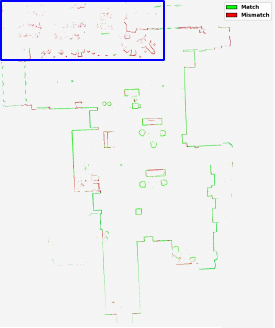}
            \label{fig:mismatch_lobby}
            %\caption{World Map}
        %\end{minipage}%
    }%
    \hspace{0.001cm}
    \subfloat[Engineering 8th-floor, Temple]{
        %\begin{minipage}{0.25\linewidth}
            \centering
            \includegraphics[width=0.4\textwidth]{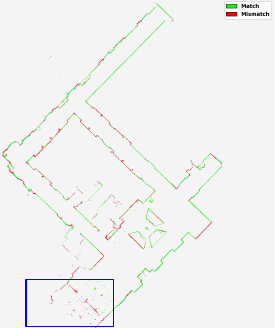}
            \label{fig:mismatch_8th}
            %\caption{World Map}
        %\end{minipage}%
    }%
    \caption{Final semantic mapping mismatch analysis across floorplans, with blue boxes highlighting cluttered and noisy regions. 
    }
    \label{fig:mismatch_mapping}
\end{figure*}

\begin{figure*}[t]
    \centering
    \subfloat[t]{
            \centering
            \includegraphics[width=0.2352\textwidth]{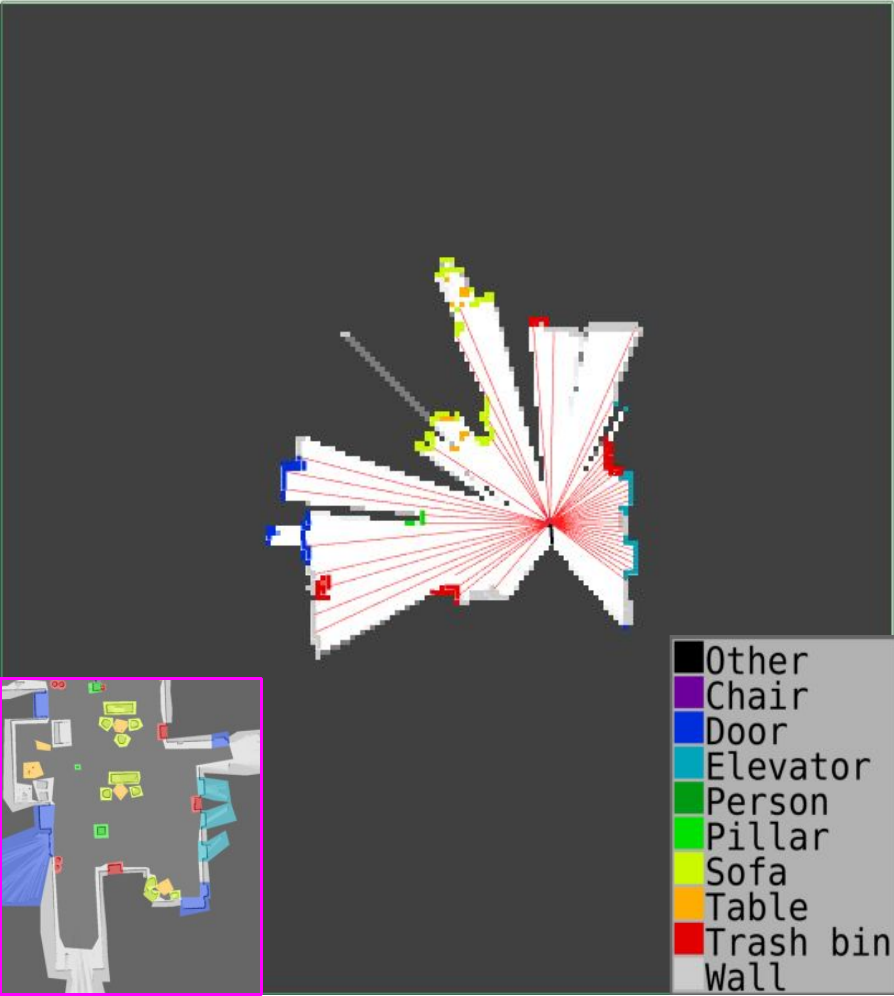}
            \label{fig:lobby_mapping1}
    }%
    %\vspace{0.01cm}
    \subfloat[t+1]{
            \centering
            \includegraphics[width=0.2352\textwidth]{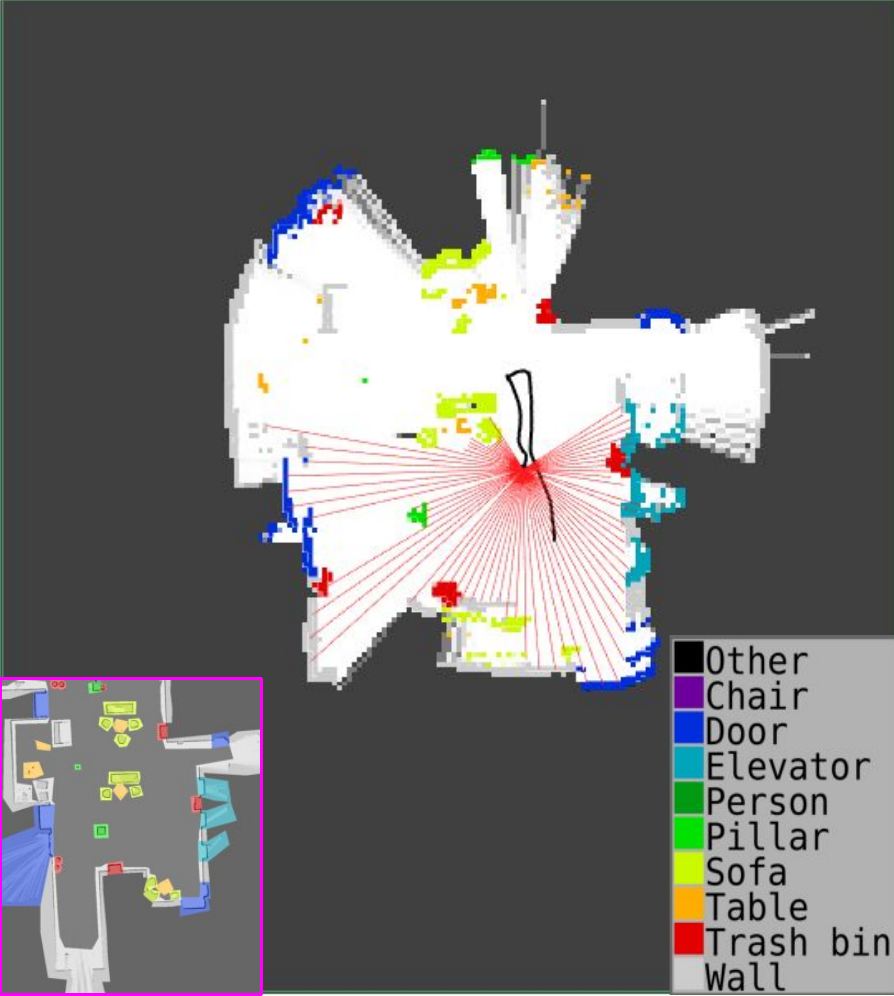}
            \label{fig:lobby_mapping2}
    }%
    \subfloat[t+2]{
            \centering
            \includegraphics[width=0.2352\textwidth]{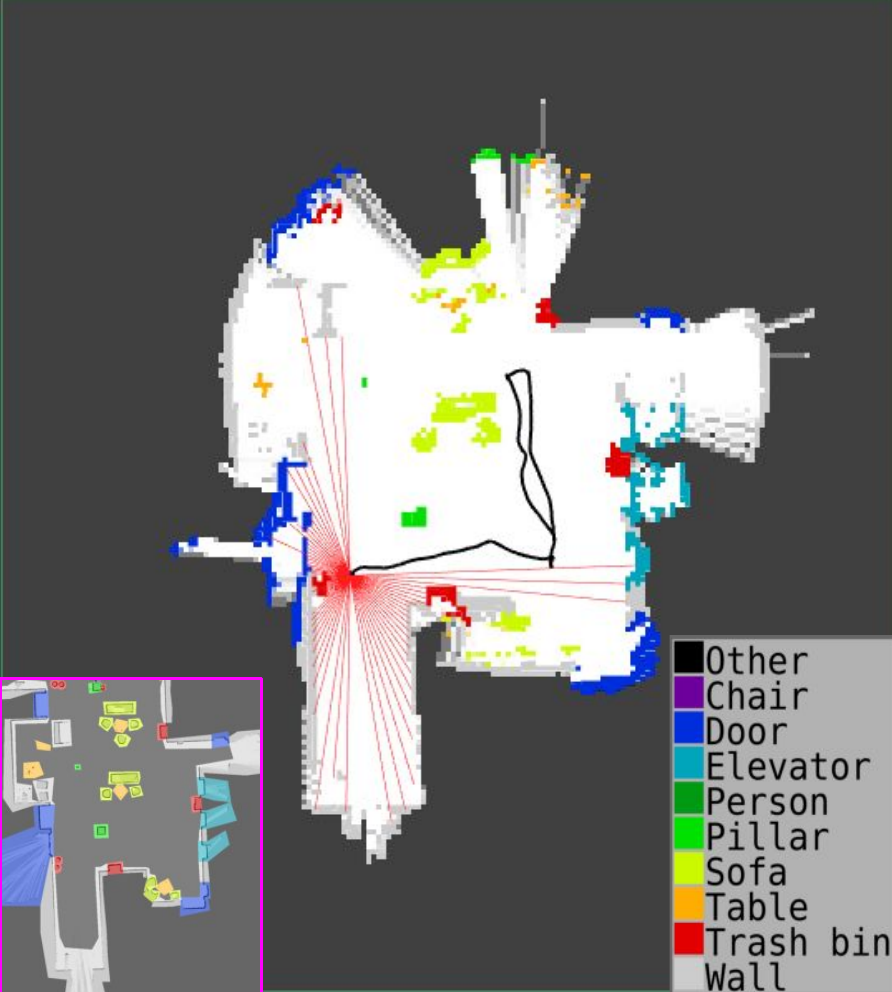}
            \label{fig:lobby_mapping3}
    }%
    \subfloat[t+3]{
            \centering
            \includegraphics[width=0.2352\textwidth]{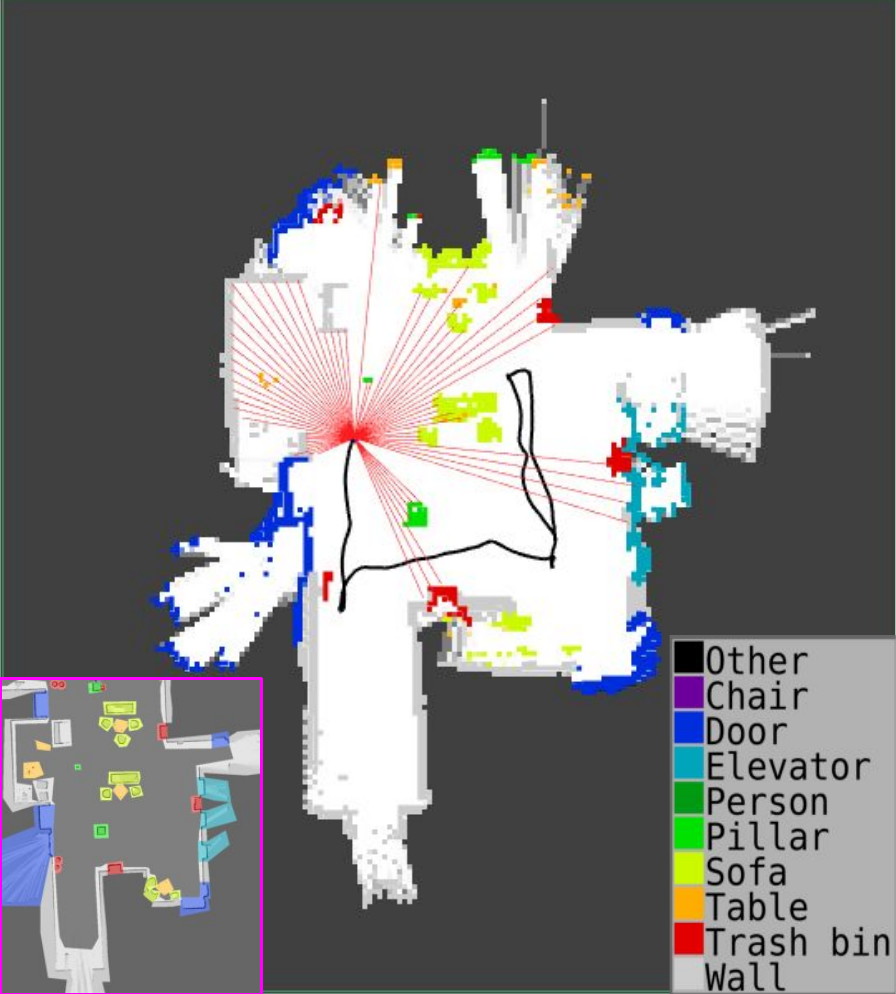}
            \label{fig:lobby_mapping4}
    }%
    \caption{
    Semantic mapping results in the Engineering lobby environment showing color-coded class labels, with 2D lidar beams (red), robot trajectory (black), and ground truth reference (magenta box).
    }
    \label{fig:semantic_mapping_lobby}
\end{figure*}

\begin{figure*}[t]
    \centering
    \subfloat[t]{
            \centering
            \includegraphics[width=0.2352\textwidth]{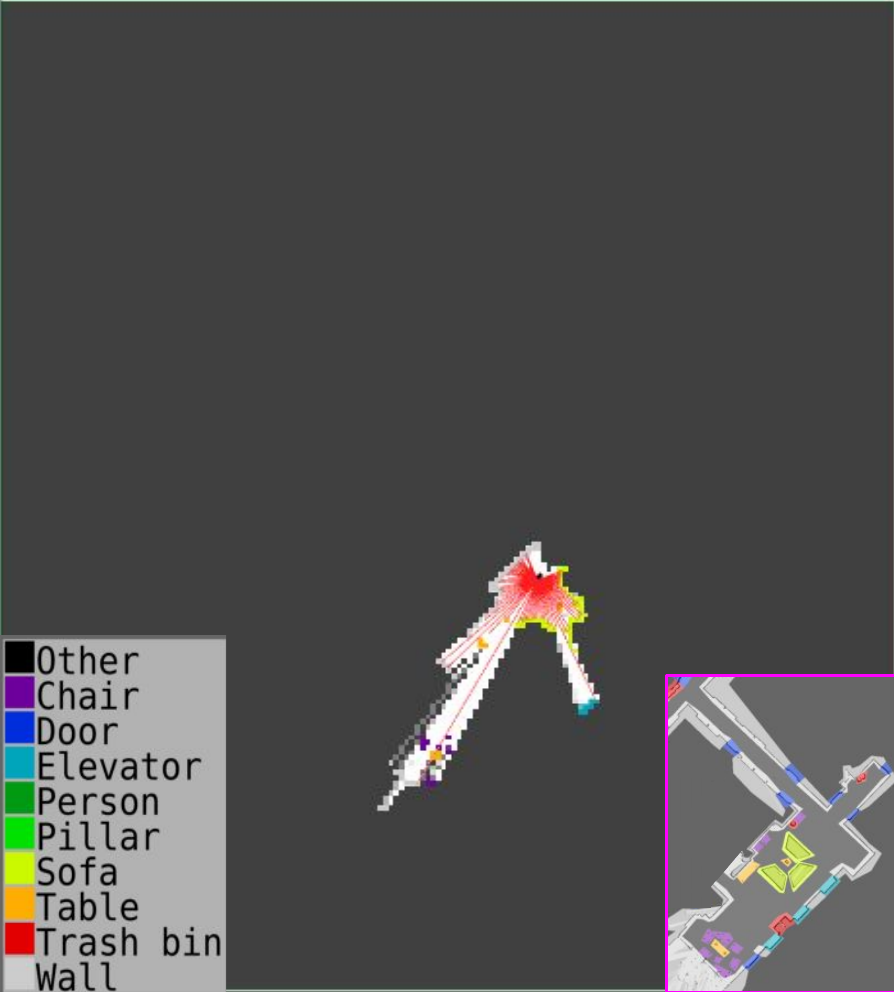}
            \label{fig:eng_8th_mapping1}
    }%
    %\vspace{0.01cm}
    \subfloat[t+1]{
            \centering
            \includegraphics[width=0.2352\textwidth]{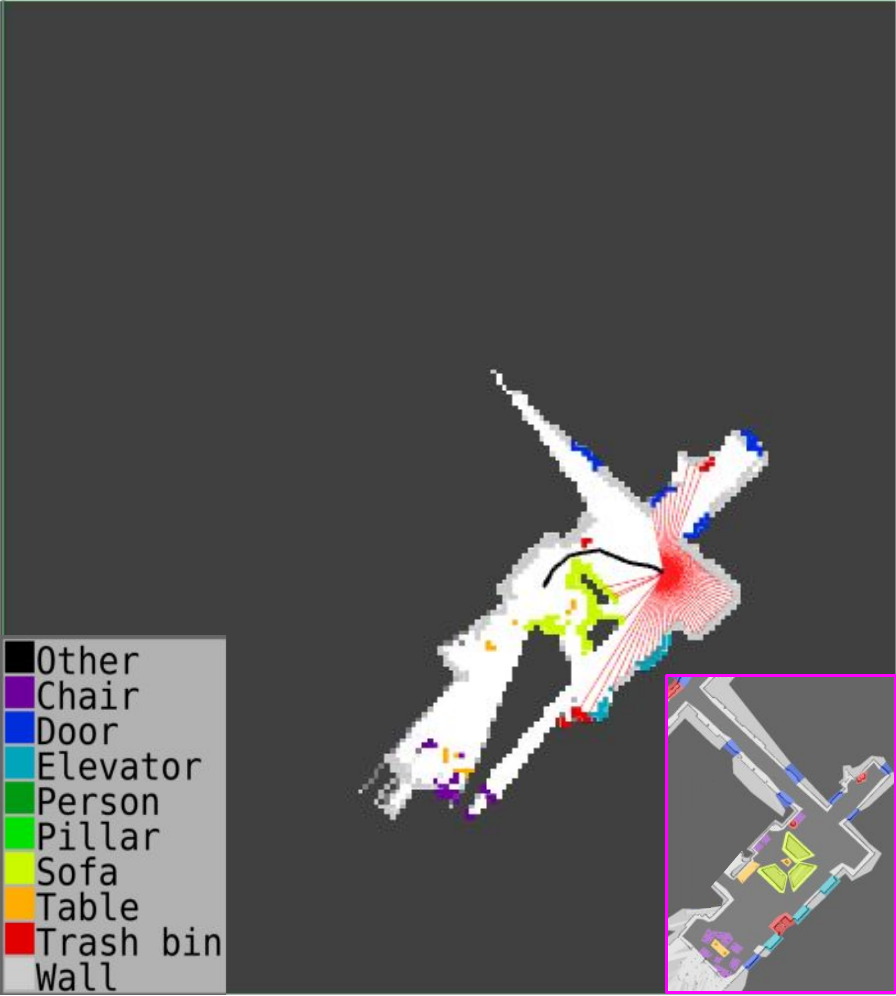}
            \label{fig:eng_8th_mapping2}
    }%
    \subfloat[t+2]{
            \centering
            \includegraphics[width=0.2352\textwidth]{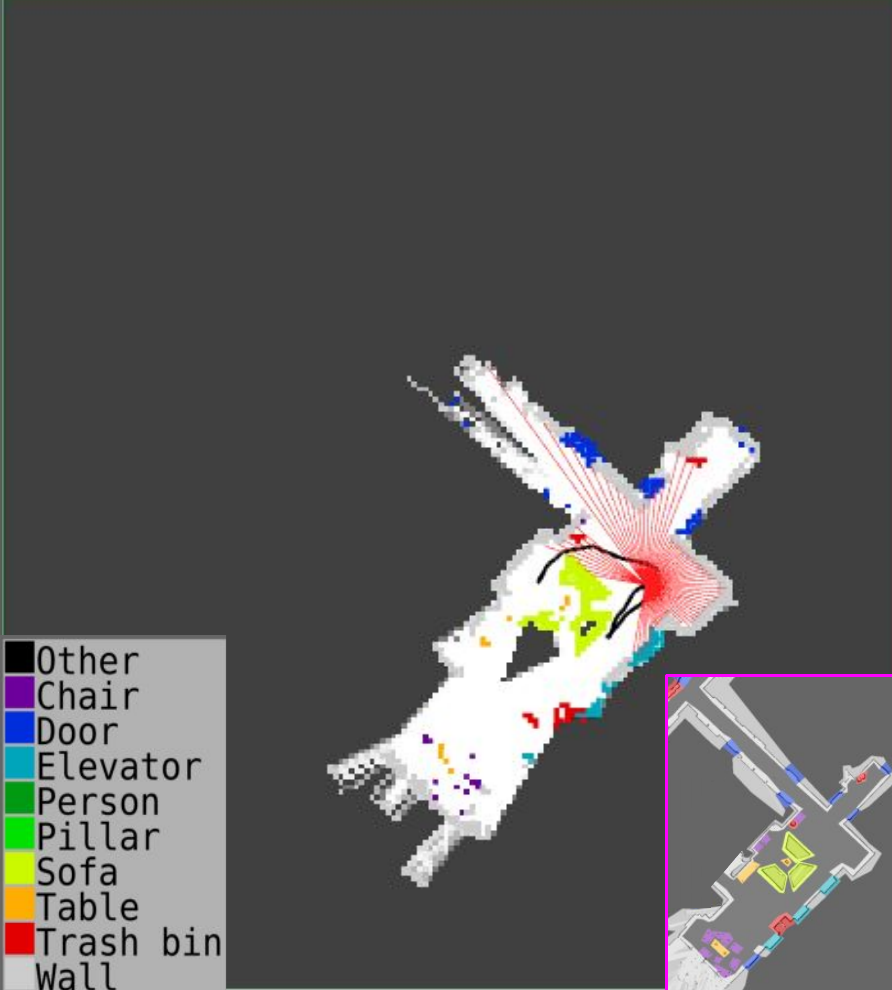}
            \label{fig:eng_8th_mapping3}
    }%
    \subfloat[t+3]{
            \centering
            \includegraphics[width=0.2352\textwidth]{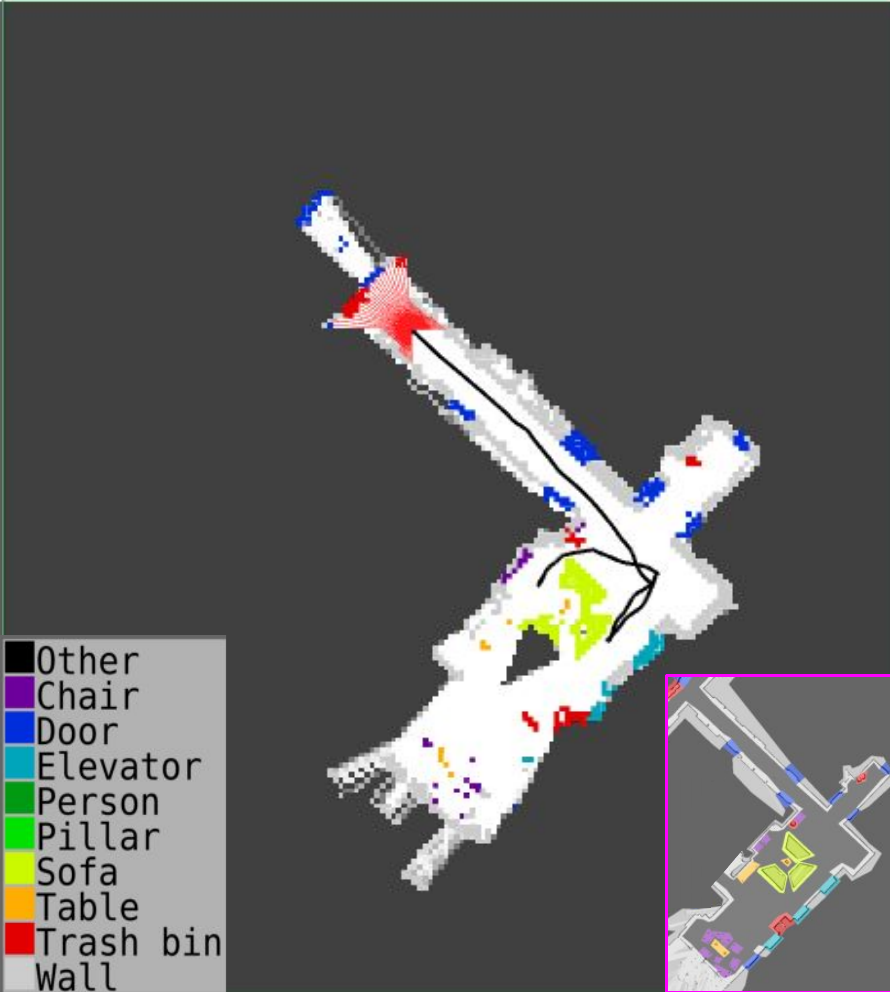}
            \label{fig:eng_8th_mapping4}
    }%
    \caption{
    Semantic mapping results in the Engineering 8th-floor environment showing color-coded class labels, with 2D lidar beams (red), robot trajectory (black), and ground truth reference (magenta box).
    }
    \label{fig:semantic_mapping_eng_8th}
\end{figure*}

\subsubsection{Quantitative Results}
\label{subsubsubsec:quantitative_results_mapping}
We evaluate the semantic mapping results against manually annotated ground truth maps using the Structural Similarity Index Measure (SSIM)~\citep{wang2004image} and semantic segmentation metrics (CA and IoU). 
\tabref{tab:semantic_mapping} presents quantitative results for the Engineering lobby and $8$th-floor environments. 
Our semantic mapping algorithm, leveraging S$^3$-Net, achieves accurate and reasonable performance, as evidenced by three key observations. 
First, the algorithm attains high scores (SSIM up to 87\%, CA up to 74\%, IoU up to 47\%), indicating strong similarity to ground truth and effective per-cell segmentation.
Second, mapping for elevators and trash cans is most reliable due to their distinctive iron material, aligning with S$^3$-Net’s use of intensity and incidence angle data. 
Third, generated maps correctly exclude ``person'' categories, consistent with the practice of omitting dynamic obstacles from environmental maps.

Despite strong overall performance, we analyze error sources using mismatch regions shown in \figref{fig:mismatch_mapping}. 
Although red mismatched areas are more extensive than expected, over 70\% of errors are explainable. \tabref{tab:category_confusion} lists the top five category confusion pairs by prevalence in mismatched cells. 
The dominant ``door–wall'' confusion arises from ambiguous boundaries between adjacent vertical structures. 
The second most common, ``door–trash bin,'' occurs near walls and entrances due to spatial proximity. 
The third and fourth pairs, ``chair–table'' (furniture-internal) and ``sofa–wall'' (line-structure merging),  reflect semantic or geometric similarities. The fifth pair, ``table–wall'' or ``door–elevator,'' stems from analogous shapes or functional roles.

These top confusions result from spatial adjacency or geometric/functional similarity rather than random error. 
The remaining errors ($<30\%$) originate from cluttered regions or annotation noise, as seen in the blue box of \figref{fig:mismatch_mapping}. 
In summary, our analysis confirms that the semantic maps are accurate, with most mismatches following expected patterns attributable to environmental complexity.

\subsubsection{Qualitative Results}
\label{subsubsubsec:qualitative_results_mapping}
\Cref{fig:semantic_mapping_lobby,fig:semantic_mapping_eng_8th} and the accompanying multimedia materials illustrate the semantic maps generated by our occupancy grid mapping algorithm using 2D lidar data. 
Compared to the ground truth floor plans of the Engineering lobby and $8$th-floor environments (\figrefs{fig:lobby}{fig:8th_floor}), the results demonstrate that our algorithm successfully constructs semantically annotated maps, despite the lower resolution relative to the original labeled maps.
These qualitative outcomes confirm that our 2D lidar-based semantic understanding workflow is effective for semantic occupancy grid mapping, and can be extended to other semantic-aware applications such as object tracking, localization, and navigation.

%%%%%%%%%%%%%%%%%%%%%%%%%%%%%%%%%%%%%%%%%%%%%%%%%%%%%%%%%%%%%%%%%%%%%%%%%%%%%%%%
\begin{figure*}[t]
    \centering
    %% WMSE:
    \subfloat[Semantic Pfeiffer]{
        %\begin{minipage}{0.25\linewidth}
            \centering
            \includegraphics[width=0.46\textwidth]{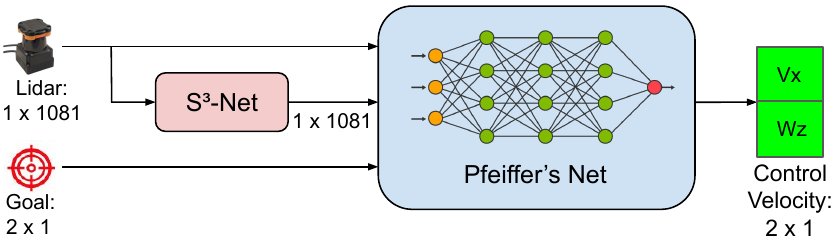}
            \label{fig:pfeiffer}
            %\caption{World Map}
        %\end{minipage}%
    }%
    \hspace{0.001cm}
    \subfloat[Semantic CNN]{
        %\begin{minipage}{0.25\linewidth}
            \centering
            \includegraphics[width=0.46\textwidth]{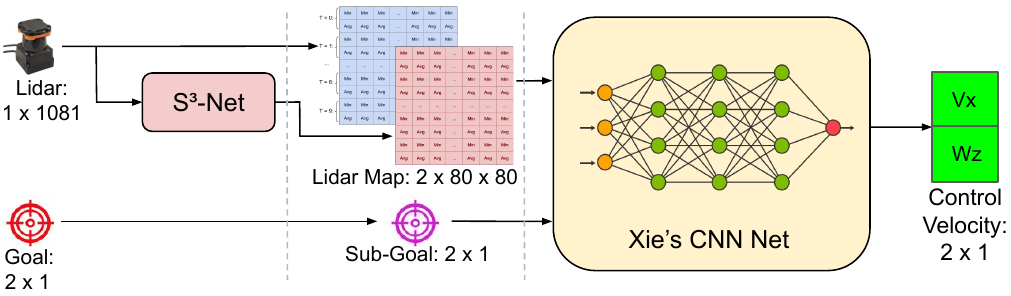}
            \label{fig:cnn}
            %\caption{World Map}
        %\end{minipage}%
    }%
    \caption{
Architectures of two supervised learning-based navigation policies, highlighting their diverging input representations: the Pfeiffer policy~\citep{pfeiffer2017perception} uses raw lidar range data, while the CNN policy~\citep{xie2021towards} processes preprocessed lidar history maps.
    }
    \label{fig:semantic_navigation}
\end{figure*}

\subsection{Semantic Navigation}
\label{subsec:semantic_navigation}
We next demonstrate how semantic information from S$^3$-Net ($\mathbf{R}+\mathbf{I}+\mathbf{A}$) enhances learning-based navigation policies, improving upon lidar-only autonomous navigation.

\subsubsection{Baselines and Training}
\label{subsubsec:baselines_and_training}
While most learning-based navigation policies~\citep{pfeiffer2017perception, long2018towards, fan2020distributed, guldenring2020learning, xie2021towards, perez2020robot, xie2023drlvo} rely solely on lidar range data, we integrate semantic information from S$^3$-Net to enhance scene understanding. 
We select two representative supervised learning policies as baselines: Pfeiffer~\citep{pfeiffer2017perception} and Xie's CNN~\citep{xie2021towards}.
For fair comparison, we exclude pedestrian kinematic data and use only lidar history as input to the CNN policy. 
For each baseline, we incorporate semantic labels as additional input channels, as illustrated in \figref{fig:semantic_navigation}.

Integrating semantic data is straightforward for the end-to-end Pfeiffer policy, which takes raw lidar ranges as input.
We simply add a parallel input channel for semantic labels, pairing each range measurement with its corresponding category (Fig. \ref{fig:pfeiffer}), yielding the Semantic Pfeiffer policy.

For the CNN policy~\citep{xie2021towards}, which downsamples lidar history using minimum and average pooling, we apply identical operations to the semantic data. 
Specifically, for minimum pooling we select the semantic label of the point with minimum range; for average pooling we apply majority voting to all semantic labels in the pooling window. 
The resulting Semantic CNN policy is shown in Fig. \ref{fig:cnn}.

We train all four policies (Pfeiffer, Semantic Pfeiffer, CNN, Semantic CNN) on Temple Engineering environment data from the Semantic2D dataset (Figs. \ref{fig:lobby}–\ref{fig:9th_floor}). 
Evaluation is conducted in a Gazebo simulator~\citep{xie2021towards, xie2023drlvo} featuring a lobby environment with 5 pedestrians (Fig. \ref{fig:omg_turtlebot2_dataset}).\footnote{Although the simulator provides zero intensity values, affecting S$^3$-Net segmentation, the semantic information remains valid for navigation.}

\subsubsection{Evaluation Metrics}
\label{subsubsubsec:evaluation_metrics_navigation}
We evaluate navigation performance using six standard metrics~\citep{loquercio2018dronet, xie2021towards, xie2023drlvo, xie2025scope}:
\begin{itemize}
    \item \textbf{Root mean square error (RMSE)}:
    \begin{equation}
         RMSE = \sqrt{\frac{1}{N} \sum_{i=1}^{N} \left({\bar{\mathbf{u}}_i - \mathbf{u}_i}\right)^2},
        \label{eq:rmse}
    \end{equation}

    \item \textbf{Explained variance ratio (EVA)}:
    \begin{equation}
         EVA = \frac{\sum_{i=1}^{N} [\left({\bar{\mathbf{u}}_i - \mathbf{u}_i}\right) - \mu_{\bar{\mathbf{u}}_i - \mathbf{u}_i}]^2} {\sum_{i=1}^{N} \left(\bar{\mathbf{u}}_i  - \mu_{{\bar{\mathbf{u}}}}\right)^2},
        \label{eq:eva}
    \end{equation}
    
    \item \textbf{Success rate}: the fraction of collision-free trials,
    
    \item \textbf{Average time}: the average travel time of trials,
    
    \item \textbf{Average length}: the average trajectory length of trials,
    
    \item \textbf{Average speed}: the average speed during trials,
\end{itemize}
where $\mathbf{u}$ and $\bar{\mathbf{u}}$ are the policy-generated and ground-truth velocity commands, respectively. 
The first two metrics assess the policy's learning accuracy, while the remaining four quantify navigation performance.

\begin{table}[t]
    \small\sf
    \centering
    \caption{Training results on the robot navigation datasets}
    \scalebox{0.85}{
    \begin{tabular}{l | c c c }
        \toprule
        \textbf{Method} & \textbf{RMSE} $\downarrow$ & \textbf{EVA} $\uparrow$ & \textbf{\# of Params} $\downarrow$ \\
        \midrule
        
        Pfeiffer~\citep{pfeiffer2017perception} & 0.1365 & 0.7369 & 51.53 M  \\
        % \midrule
        
        \textcolor{blue}{Semantic Pfeiffer}  & 0.0980 & 0.8665 & 51.54 M  \\
        \midrule
        % \midrule
         
        CNN~\citep{xie2021towards} & 0.0766 & 0.9199 & \textbf{28.98 M} \\
        % \midrule
        
        \textcolor{blue}{Semantic CNN}  & \textbf{0.0567} & \textbf{0.9565} & 28.99 M  \\
        \bottomrule
    \end{tabular}
    }
    \label{tab:training}
\end{table}

\begin{table*}[t]
    \small\sf\centering
    \caption{Navigation results in 3D simulation dynamic environment}
    \scalebox{0.835}{
    \begin{tabular}{l | l | c c c c}
        \toprule
        \textbf{Environment} & \textbf{Method} & \textbf{Success Rate} $\uparrow$ & \textbf{Average Time (s)} $\downarrow$ & \textbf{Average Length (m)} $\downarrow$ & \textbf{Average Speed (m/s)} $\uparrow$ \\ 
        \midrule

        \multirow{4}{*}{\begin{tabular}[c]{@{}c@{}}Lobby world\end{tabular}} 
         & Pfeiffer~\citep{pfeiffer2017perception} & 0.13 & 5.65 & 1.51 & 0.27  \\  
         & \textcolor{blue}{Semantic Pfeiffer} & 0.48 & 22.84 & 5.78 & 0.25   \\  
         \cmidrule(lr){2-6}  % trim from left/right
         
         & CNN~\citep{xie2021towards} & 0.83 & 45.80 & 10.84 & 0.23    \\ 
         & \textcolor{blue}{Semantic CNN} & \textbf{0.89} & \textbf{15.39} & \textbf{5.77} & \textbf{0.38} \\
         \bottomrule
    \end{tabular}
    }
    \label{tab:navigation}
\end{table*}

\subsubsection{Quantitative Results}
\label{subsubsubsec:quantitative_results_navigation}
\tabref{tab:training} presents the quantitative training results of our Semantic Pfeiffer and Semantic CNN policies compared to their non-semantic counterparts, revealing three key findings.

First, both semantic policies achieve statistically significant improvements in RMSE and EVA over their baseline versions. 
This demonstrates that semantic information from S$^3$-Net enhances prediction accuracy and navigation performance for both raw and preprocessed lidar inputs, validating the utility of our Semantic2D dataset and segmentation approach even with limited training data.

Second, the performance gap between Semantic Pfeiffer and its baseline is substantially larger than that between Semantic CNN and its original version. 
This indicates that policies using raw lidar data benefit more from semantic enrichment than those relying on preprocessed inputs, highlighting the particular value of semantic information for raw-data-based navigation policies~\citep{pfeiffer2017perception, xie2021towards}.

Third, the semantic policies require only minimal parameter increases, maintaining computational efficiency for resource-constrained platforms.

\tabref{tab:navigation} shows consistent trends in deployment results. 
Both semantic policies achieve higher success rates, with Semantic Pfeiffer demonstrating notably greater improvement over its baseline than Semantic CNN. 
The lower absolute performance of Pfeiffer-based policies stems from their limited generalization when trained on small datasets with raw inputs, while the longer navigation times of CNN policies reflect generalization challenges in unseen environments. 
The consistent gains from semantic information highlight its value for improving policy generalization, even with small supervised datasets. 
These results underscore the importance of our complete semantic workflow -- dataset, labeling framework, and segmentation network -- for advancing semantic-aware navigation and related applications.

\begin{figure*}[t]
    \centering
    \subfloat[t]{
            \centering
            \includegraphics[width=0.45\textwidth]{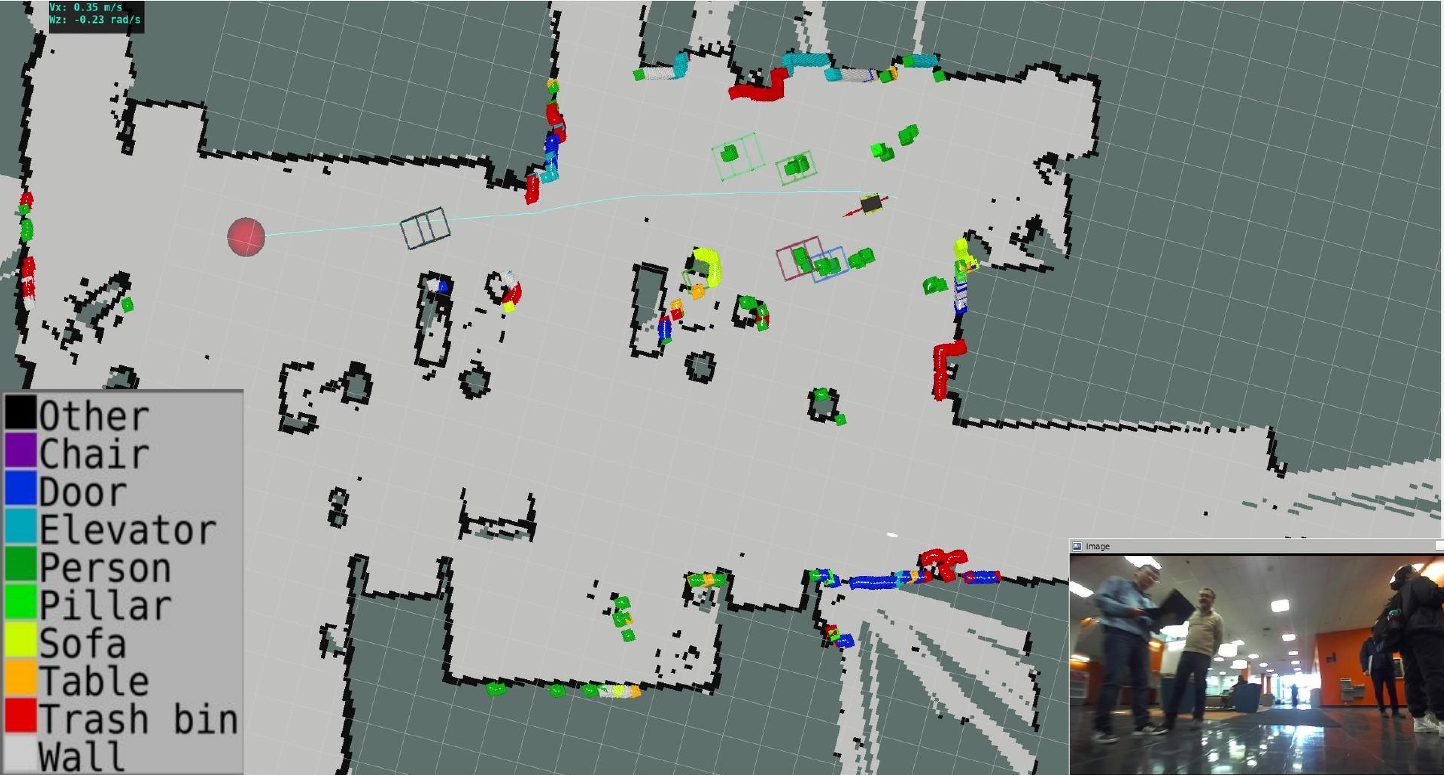}
            \label{fig:semantic_cnn_lobby1}
    }%
    \subfloat[t+1]{
            \centering
            \includegraphics[width=0.45\textwidth]{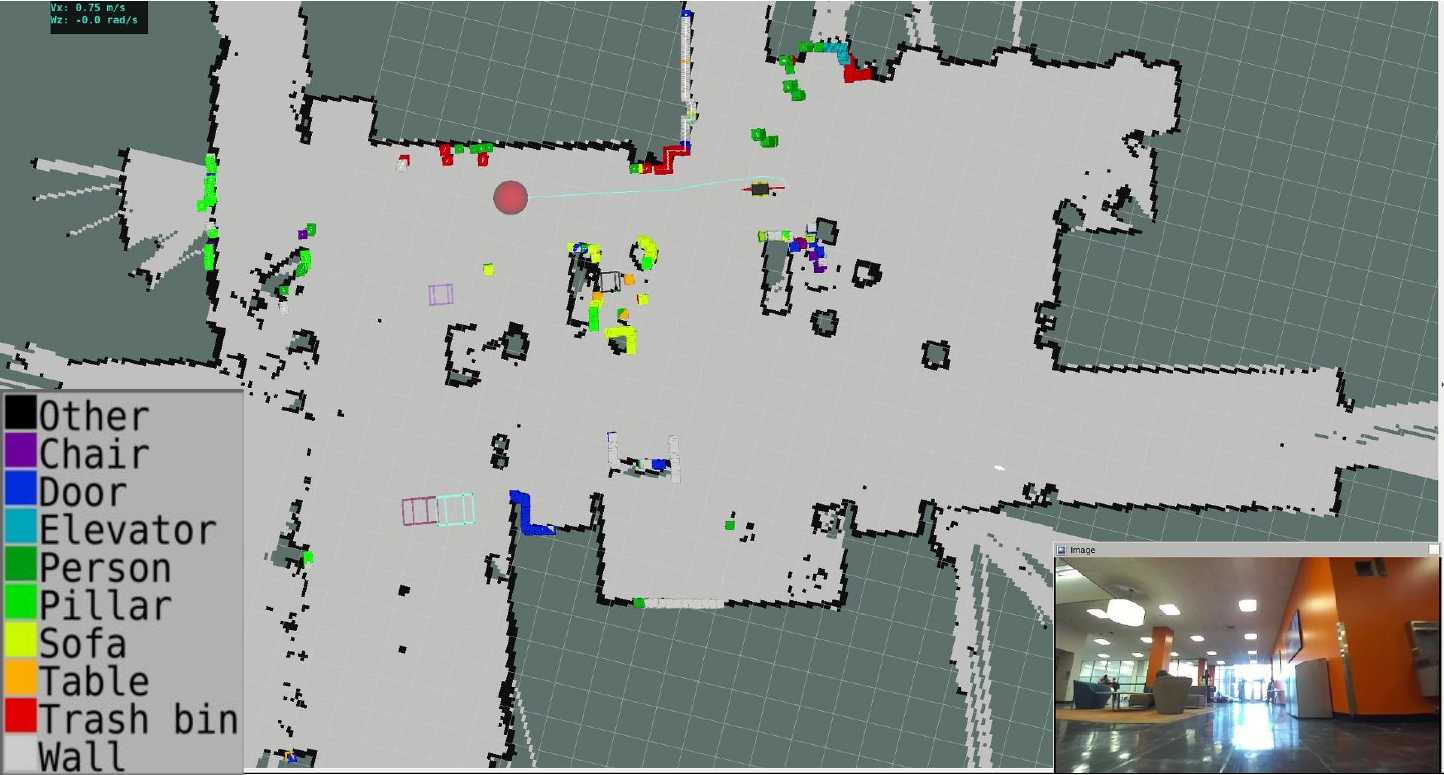}
            \label{fig:semantic_cnn_lobby2}
    }%
    % \vspace{0.001cm}
    
    \caption{Robot deployed with semantic CNN navigates through the lobby of Temple University’s engineering building.
            }
    \label{fig:semantic_cnn_showcase1}
\end{figure*}

\begin{figure*}[t]
    \centering
    \subfloat[t]{
            \centering
            \includegraphics[width=0.45\textwidth]{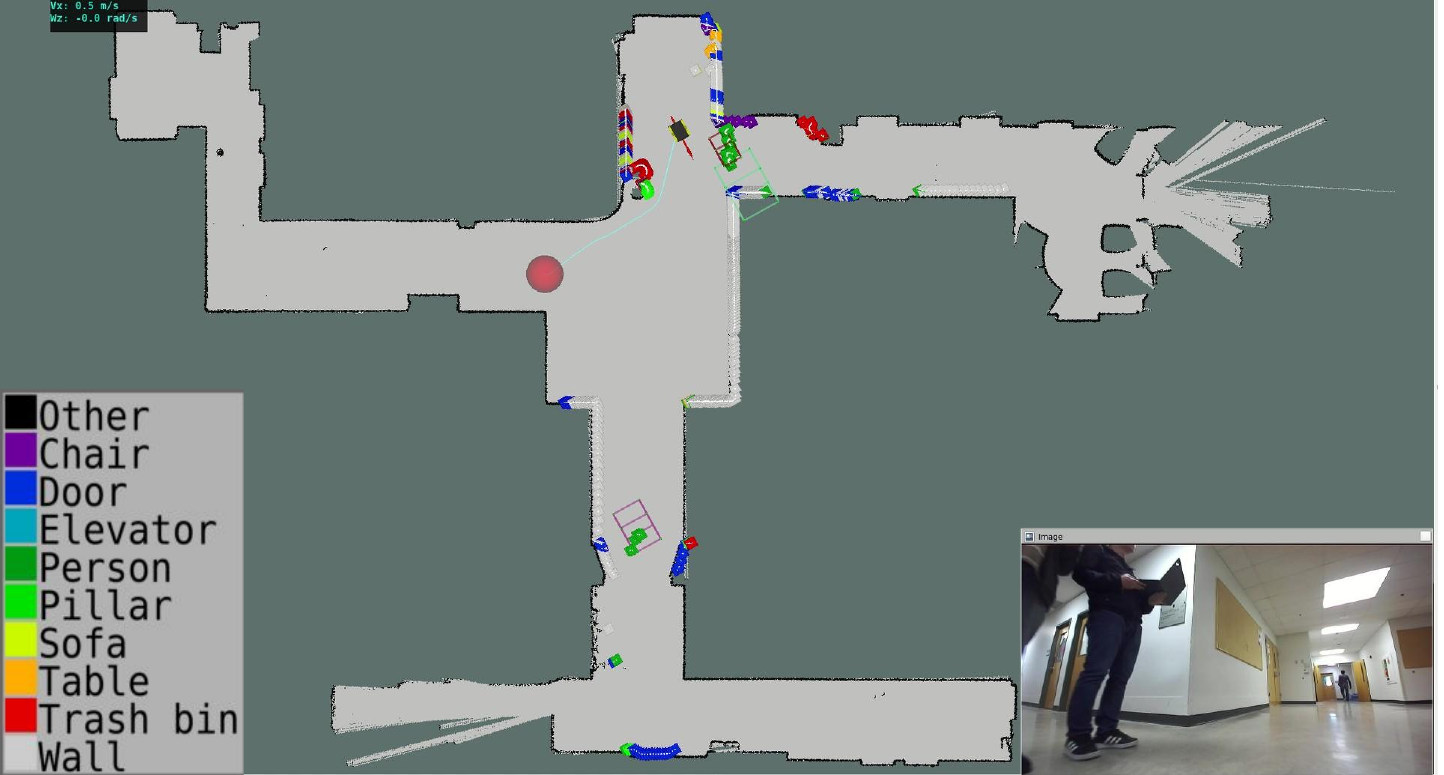}
            \label{fig:semantic_cnn_ength1}
    }%
    \subfloat[t+1]{
            \centering
            \includegraphics[width=0.45\textwidth]{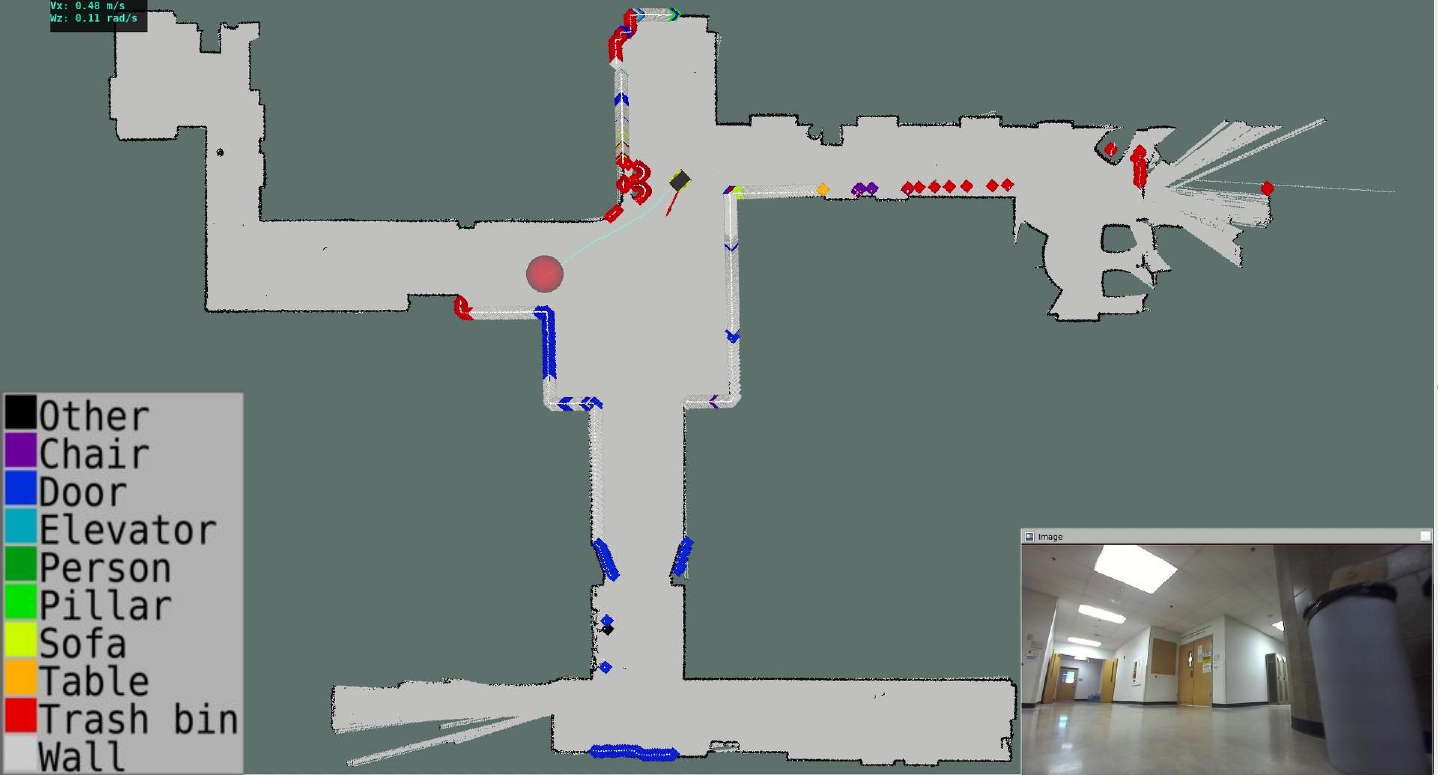}
            \label{fig:semantic_cnn_ength2}
    }%
    % \vspace{0.001cm}
    
    \caption{Robot deployed with semantic CNN navigates through the 4th-floor of Temple University’s engineering building.
            }
    \label{fig:semantic_cnn_showcase2}
\end{figure*}

\begin{figure*}[t]
    \centering
    \subfloat[t]{
            \centering
            \includegraphics[width=0.45\textwidth]{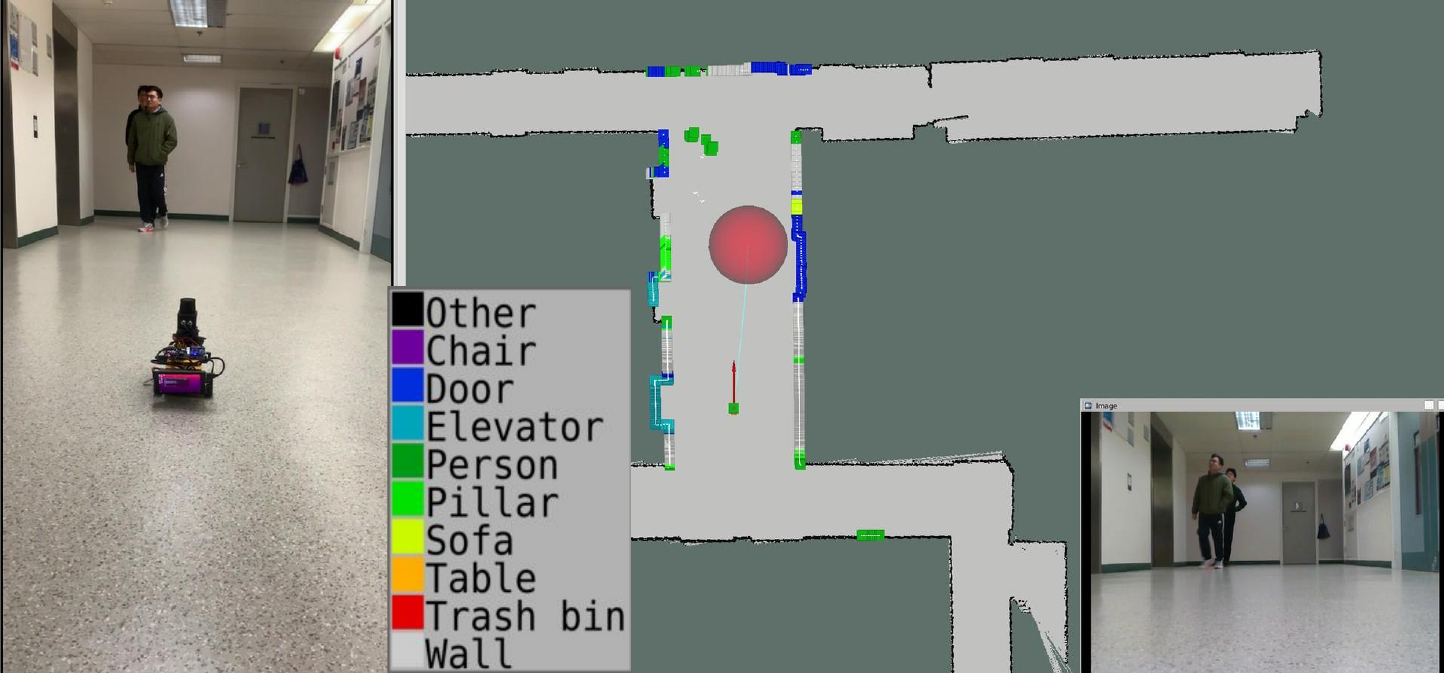}
            \label{fig:semantic_cnn_cyc1}
    }%
    \subfloat[t+1]{
            \centering
            \includegraphics[width=0.45\textwidth]{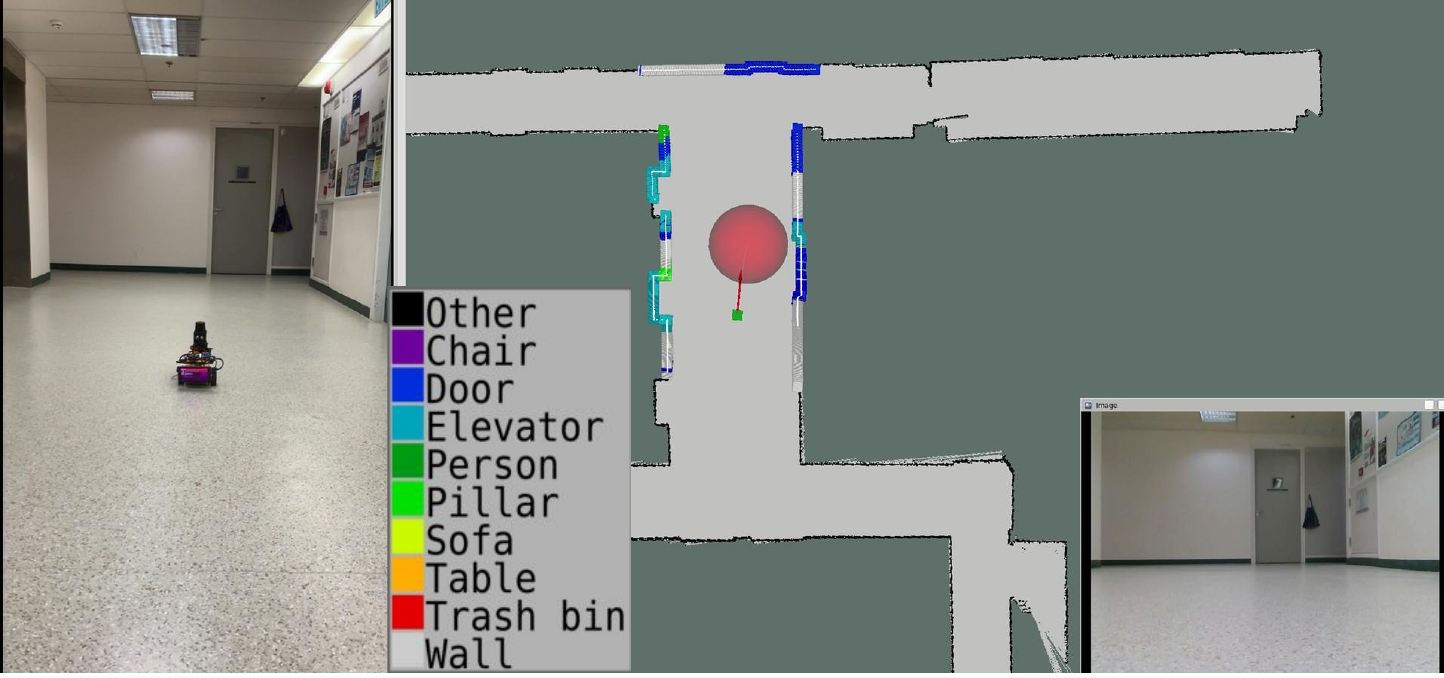}
            \label{fig:semantic_cnn_cyc2}
    }%
    % \vspace{0.001cm}
    
    \caption{Robot deployed with semantic CNN navigates through the 4th-floor of HKU’s Chow Yei Ching building.
            }
    \label{fig:semantic_cnn_showcase3}
\end{figure*}

\subsubsection{Qualitative Results}
\label{subsubsec:qualitative_results_navigation} 
We validate the real-world effectiveness of our S$^3$-Net semantic segmentation algorithm and Semantic CNN policy through physical robot experiments. 
As shown in \Cref{fig:semantic_cnn_showcase1,fig:semantic_cnn_showcase2} and the accompanying multimedia, a Jackal robot equipped with our system successfully perceives its environment and navigates around both static obstacles and moving pedestrians to reach predefined goals in Temple University's Engineering lobby and 4th-floor environments.

To test generalization, we directly deployed our Temple-trained models (using Hokuyo UTM-30LX-EW lidar data) without fine-tuning to a different robot platform (customized robot with WLR-716 lidar) at the University of Hong Kong. 
\Cref{fig:semantic_cnn_showcase3} demonstrates successful navigation in the Chow Yei Ching building, confirming strong cross-platform generalization across robot models, sensor types, and environmental conditions.

These results validate that our semantic 2D lidar workflow provides a practical, camera-free solution for enhancing semantic scene understanding in real-world robotic applications.

%%%%%%%%%%%%%%%%%%%%%%%%%%%%%%%%%%%%%%%%%%%%%%%%%%%%%%%%%%%%%%%%%%%%%%%%%%%%%%%%
\section{Conclusion}
\label{sec:conclusion}
This article presents a complete workflow for semantic scene understanding using only 2D lidar, demonstrating that fine-grained semantic perception significantly enhances mobile robotics algorithms. 
Our contributions are fourfold.

First, we introduce Semantic2D, the first 2D lidar semantic segmentation dataset for mobile robotics applications. 
This dataset provides point-wise annotations for nine indoor object categories (e.g., walls, tables, doors) and includes comprehensive data for various robotics tasks (object tracking, mapping, localization, and navigation), including poses, odometry, RGB/depth images, navigation goals, paths, and control commands. 
To enable efficient annotation, we develop SALSA, a semi-automatic labeling framework that combines manual map annotation with ICP-based scan alignment, significantly reducing manual effort. 
We validate SALSA on the OGM-Turtlebot2 and MIT Stata Center datasets, demonstrating its utility for creating high-quality 2D lidar annotations.

Second, we propose S$^3$-Net, an efficient stochastic semantic segmentation network based on a VAE that delivers robust performance on resource-constrained robots. 
Through ablation studies, we determine the optimal input representation and show that S$^3$-Net with range, intensity, and incident angle inputs outperforms both traditional geometry-based methods and other input configurations.

Third, we demonstrate practical applications in semantic mapping and navigation using only a single 2D lidar sensor. 
Our approach generates accurate semantic occupancy grid maps across different environments, while our semantically-aware navigation policies (Semantic Pfeiffer and Semantic CNN) outperform their non-semantic counterparts in both simulated and real-world experiments, including cross-platform deployment on different robots and sensors.

Finally, we open-source our dataset and algorithms to encourage further research in 2D lidar-based semantic understanding for mobile robotics.

%%%%%%%%%%%%%%%%%%%%%%%%%%%%%%%%%%%%%%%%%%%%%%%%%%%%%%%%%%%%%%%%%%%%%%%%%%%%%%%%
\begin{acks}
\label{sec:acknowledgment}
The authors would like to thank Alkesh Srivastava for his help in teleoperating the Jackal robot and labeling six environment maps, and Kevin Formento for his help in teleoperating the Jackal robot and collecting data in three environments. 
\end{acks}

\begin{funding}
This work was funded by Temple University and the University of Hong Kong.
\end{funding}

\begin{dci}
The authors declare that there is no conflict of interest. 
\end{dci}

%%%%%%%%%%%%%%%%%%%%%%%%%%%%%%%%%%%%%%%%%%%%%%%%%%%%%%%%%%%%%%%%%%%%%%%%%%%%%%%%
\bibliographystyle{bib/SageH}
\bibliography{bib/self,bib/refs}

\end{document}